\documentclass{article}

  \usepackage[eandd, final]{neurips_2026}

\usepackage[utf8]{inputenc} % allow utf-8 input
\usepackage[T1]{fontenc}    % use 8-bit T1 fonts
\usepackage{hyperref}       % hyperlinks
\usepackage{url}            % simple URL typesetting
\usepackage{booktabs}       % professional-quality tables
\usepackage{amsfonts}       % blackboard math symbols
\usepackage{nicefrac}       % compact symbols for 1/2, etc.
\usepackage{microtype}      % microtypography
\usepackage{xcolor}         % colors
\usepackage{graphicx}
\usepackage{subcaption}
\usepackage{siunitx}
\usepackage{natbib}
\usepackage{threeparttable}
\usepackage{amsmath}
\usepackage{amssymb}
\usepackage{chngcntr} % counter per section in appendix
\usepackage{pifont} % for checkmarks
\usepackage{array}
\usepackage{tikz}
\usetikzlibrary{arrows.meta,positioning}

\title{CompleteRXN: Toward Completing Open Chemical Reaction Databases}

\author{%
  Gabriel Vogel$^{1}$ \quad
  Minouk Noordsij$^{1}$ \quad
  Evgeny Pidko$^{2}$ \quad
  Jana M. Weber$^{1}$ \\
  $^{1}$Department of Intelligent Systems, Delft Bioinformatics Lab, Delft University of Technology \\
  $^{2}$Department of Chemical Engineering, Delft University of Technology \\
  Delft, The Netherlands \\
  \texttt{g.vogel@tudelft.nl} \quad
  \texttt{j.m.weber@tudelft.nl}
}

\begin{document}

\maketitle

\begin{abstract}
Chemical reaction datasets such as USPTO suffer from substantial incompleteness, frequently missing byproducts, co-reactants, and stoichiometric coefficients. This limits their applicability and reliability in downstream applications. Here, we introduce CompleteRXN, a large-scale supervised benchmark for reaction completion under realistic missing-data conditions. We construct a dataset of aligned incomplete and atom-balanced reactions by mapping USPTO records to curated mechanistic reactions. We evaluate representative baselines, including a novel encoder-decoder reaction completion model with constrained decoding, the Constrained Reaction Balancer (CRB), and a recent algorithmic method, SynRBL. On our CompleteRXN benchmark, the CRB achieves high performance across splits of increasing difficulty, reaching 99.28\% equivalence accuracy on the random split and 91.71\% on the extreme out-of-distribution split. SynRBL produces many balanced and chemically plausible completions, but with lower accuracy on the benchmark test splits. Across all methods, performance degrades with increasing incompleteness. We observe a substantial drop when evaluating on reactions outside the benchmark (full uncurated USPTO), highlighting the gap between benchmark performance and practical robustness and motivating future work. Data, code and models can be found at: \url{https://github.com/Intelligent-molecular-systems/CompleteRxn-Benchmarking}.

\end{abstract}

\section{Introduction}
\label{sec:introduction}

Modern chemistry is increasingly driven by data-centric technologies. Large-scale chemical reaction datasets, combined with advances in machine learning, have enabled substantial progress in complex chemical tasks, such as reaction prediction, retrosynthesis, and molecular design~\cite{du2024GenAIDrugDiscoveryReview, coley2017MLforRS, segler2017FPandRS, coley2017reactionPrediction, schwaller2019molecularTransformerFP}. Reaction data is also becoming important beyond such core tasks, for instance in automated process modeling and sustainability assessment, where reaction pathways are linked to process and environmental impact models~\cite{weber2021chemicaldataintelligence, ioannou2021process}. Recent work shows that retrosynthesis-derived reaction networks can be used to generate life cycle inventory data at scale~\cite{chen2026LCACrystal}. 

Large-scale reaction datasets inherit systematic biases and quality limitations from preceding data mining steps. The widely used USPTO dataset~\cite{lowe2012extraction}, for example, is derived from patent text and has a substantial fraction of chemically incomplete reactions: important molecules such as byproducts are vastly omitted and stoichiometric factors are neglected, resulting in atom- and charge-unbalanced reactions. This poses a serious issue for large-scale automated process modeling and sustainability assessment, which rely on chemically consistent records for mass- and energy balances~\cite{blanco2024machine}. It further remains an open question to what extent fully atom- and charge-balanced reaction data impacts the performance of the plethora of machine learning models trained on this data, e.g. for reaction prediction and retrosynthesis.

Prior work has addressed reaction data quality through preprocessing, filtering, and standardization~\cite{schwaller2018USPTOSTEREO,jin2017USPTOMIT}, as well as through methods for reaction rebalancing and completion~\cite{phan2024reaction, van2024completing, zhang2024hybridcompletingreactions}. In parallel, a recent effort has introduced a benchmark suite for reaction modeling, including the task of reaction rebalancing~\cite{phan2026synrxn}. However, evaluation of existing approaches either relies on synthetic corruption of balanced reactions~\cite{van2024completing, zhang2024hybridcompletingreactions} or on small-scale manual validation of testsets~\cite{phan2024reaction, phan2026synrxn}.

In this work, we introduce CompleteRXN, a large-scale supervised benchmark for chemical reaction completion under realistic missing-data conditions. We construct a dataset of aligned reaction pairs by mapping incomplete USPTO records to counterparts derived from an atom-balanced mechanistic dataset~\cite{joung2025electron}, yielding realistic input-output pairs for systematic evaluation. We formulate reaction completion as a structured missing-data problem, where incomplete reaction representations must be mapped to chemically valid, atom-balanced outputs. Our work complements the recent benchmark effort for reaction completion in SynRXN~\cite{phan2026synrxn}, by providing a larger set of input–output pairs with template-curated, mechanistically derived targets, rather than relying on sample-level manual validation of test sets.

We evaluate both algorithmic and machine learning approaches on the task of completing incomplete reaction equations across random, mechanism-based, and out-of-distribution splits, using an evaluation metric that accounts for chemically equivalent completions. Our results show strong completion performance on the CompleteRXN benchmark under random splits while revealing systematic degradation under distribution shift and increasing reaction incompleteness. Furthermore, as the template-aligned data represents a relatively clean subset of USPTO data, we additionally evaluate models on the remaining USPTO reactions, which better reflect real-world complexity. On this task, model performance drops significantly, motivating the development of complementary manually curated benchmarks targeting noisy and challenging reactions.

A summary of our main contributions is:
\begin{itemize}
    \item A large-scale dataset of aligned incomplete and atom-balanced reactions derived from USPTO data, enabling supervised learning for reaction completion
    \item A benchmark and evaluation protocol, including challenging out-of-distribution splits that test generalization across reaction mechanisms and increasing levels of incompleteness 
    \item A constrained decoding strategy for sequence-to-sequence models that enforces atom-balance during generation (Constrained Reaction Balancer, CRB)
    \item A systematic analysis of algorithmic and machine learning model performance

\end{itemize}

\section{Related Work}
\label{sec:related_work}
Related work on reaction completion spans algorithmic, data-driven, and hybrid approaches.
SynRBL~\cite{phan2024reaction} formulates the completion problem as a combination of chemical rules and graph-based alignment, adding common small molecules and reconstructing missing fragments via subgraph matching. Van Wijngaarden et al.~\cite{van2024completing} combine rule-based completion with sequence-to-sequence models, while Zhang et al.~\cite{zhang2024hybridcompletingreactions} follow a hybrid strategy that integrates heuristic rebalancing with masked language modeling to iteratively infer missing species. A slightly different formulation of reaction completion was introduced by Zipoli et al.~\cite{zipoli2024completion}, who train a single transformer jointly on forward prediction, retrosynthesis, and completion tasks. In their setting, completion is defined as predicting any missing molecules within a reaction SMILES, disregarding atom balances.

From a data perspective, these approaches differ primarily in how training data, test data, and ground truth are constructed (Table~\ref{tab:data_comparison}). Van Wijngaarden et al.~\cite{van2024completing} and Zhang et al.~\cite{zhang2024hybridcompletingreactions} first construct balanced parent reactions using rule-based or heuristic completion procedures. These completed reactions are subsequently partialized to generate synthetic incomplete-complete pairs for model training and evaluation. To quantify how this differs from empirical incompleteness, we apply the corruption procedures of prior work~\cite{zhang2024hybridcompletingreactions,van2024completing} to the same balanced targets used in CompleteRXN. Synthetic corruption is not uniformly easier, but produces systematically different missing-data distributions: omissions are more symmetric across reaction sides and tend to contain larger and more carbon-rich fragments, whereas empirical USPTO omissions are strongly product-side biased (Appendix~\ref{app:synthetic_corruption_comparison}).

In contrast, a major advance, SynRXN~\cite{phan2026synrxn} enables evaluating reaction completion on realistic incomplete reactions found in the USPTO. However, the approach constructs ground truth using model-generated completions (via SynRBL) with manual validation of test sets, introducing dependence on a specific model. As summarized in Table~\ref{tab:data_comparison}, existing approaches therefore trade off realism, scalability, and ground-truth independence. In this work, we provide a large-scale benchmark with realistic incomplete reactions and atom-balanced targets derived from a mechanistic dataset~\cite{joung2025electron}. The targets are based on expert-curated templates that are systematically applied to template-matched USPTO records, yielding scalable and model-independent ground truth.

\newcommand{\cmark}{\textcolor{green!50!black}{\ding{51}}}
\newcommand{\xmark}{\textcolor{red!60!black}{\ding{55}}}
\begin{table}[t]
\centering
\small
\caption{Comparison of input reaction data, completed reaction data, and evaluation across different reaction completion approaches.}
\label{tab:data_comparison}
\begin{tabular}{p{3.4cm} >{\centering\arraybackslash}p{2.0cm} >{\centering\arraybackslash}p{2.0cm} >{\centering\arraybackslash}p{2.1cm} >{\centering\arraybackslash}p{2.1cm}}
\toprule
 & \textbf{Zipoli et al.}~\cite{zipoli2024completion} 
 & \textbf{Zhang / van~W.}~\cite{zhang2024hybridcompletingreactions, van2024completing} 
 & \textbf{SynRXN}~\cite{phan2026synrxn} 
 & \textbf{CompleteRXN (ours)} \\
\midrule

\multicolumn{5}{l}{\textbf{Input data}} \\
Incomplete-input origin & \xmark (synthetic) & \xmark (synthetic) & \cmark(empirical) n=48,167 & \cmark(empirical) n=206,423 \\

\midrule
\multicolumn{5}{l}{\textbf{Complete ground truth}} \\
Human validation & \xmark & \xmark & \cmark\ (examples) & \cmark\ (templates) \\
Atom-balanced reactions & \xmark & \cmark & \cmark & \cmark \\
Ground-truth origin & N.A. & Rule-based & SynRBL  + expert valid. & Template-curated (experts) \\
Constr. model-independent & \cmark & \xmark & \xmark & \cmark \\

\midrule
\multicolumn{5}{l}{\textbf{Evaluation}} \\
Exact-match & \cmark & \cmark & \cmark & \cmark \\
Equivalence-match & \xmark & \xmark & \xmark & \cmark \\

\bottomrule
\end{tabular}
\end{table}

\section{Incompleteness and missing patterns in USPTO database}
\label{sec:data-analysis}
The USPTO chemical reaction dataset~\cite{lowe2012extraction} is largely atom-unbalanced, with only $\approx$4.8\% of reactions being atom-balanced. The most common issue is missing byproducts (i.e., missing molecules on the product side), likely because they are often not explicitly reported in patents and thus not captured by text extraction pipelines. Other sources of incompleteness include missing co-reactants, incorrect stoichiometry, or mislabeled molecules (e.g., reactants vs. reagents or solvents). In some cases, reactions also contain molecules that do not belong to a reaction or multiple alternative (not labeled as such) molecules when patents describe several possible reactants or products. Over the full dataset, on average, reactions are missing 8.9 atoms (of which on average 2.5 atoms are carbons), indicating a large and non-trivial space of possible completions.

Importantly, reaction incompleteness is not random but follows structured patterns across different levels of reaction description. \emph{Reaction classes} group reactions by overall transformation type, \emph{reaction templates} capture specific structural changes between reactants and products, and \emph{mechanisms} describe the underlying chemical principles. Across reaction classes, missing information varies systematically: some classes predominantly lack small molecules, while others are missing larger fragments, often including carbon-containing parts (Appendix Figure~\ref{fig:class_incompleteness_patterns}). At the template level, these patterns become highly consistent. For a given template, reactions typically share the same atom-balance pattern. In the USPTO 1k TPL subset~\cite{schwaller2021mapping}, 782 templates exhibit a single atom-balance pattern across all reactions, and most templates are either consistently unbalanced (96.1\%) or fully balanced (3.0\%), with only 0.9\% showing mixed behavior (Appendix Figure~\ref{fig:template_completeness}).

\section{The CompleteRXN Dataset}
\label{sec:dataset}
We construct a dataset for benchmarking chemical reaction completion models with atom-balanced single-line reaction representations. The dataset combines two complementary sources: \textbf{incomplete reactions}, given by raw USPTO records that serve as realistic and noisy inputs, and \textbf{curated reactions}, which are atom-balanced records derived using mechanistic reaction data from FlowER~\cite{joung2025electron} and serve as high-quality targets. FlowER, a tool for reaction mechanism prediction, provides atom-balanced mechanistic pathways for USPTO reactions, constructed by applying expert-curated elementary-step templates to the corresponding reaction records.

\paragraph{Data format}
Molecules are represented using the text-based Simplified Molecular Input Line Entry System (SMILES). Reactions are encoded as Reaction SMILES, where reactants, reagents, and products are separated by arrow tokens, and individual molecules by dots (e.g. \texttt{CCO.CC(=O)O > O=S(=O)(O)O > CC(=O)OCC.O}).
For consistency, we move reagents to the reactant side:
\texttt{reactant1.reactant2.reagent1.reagent2 > > product1}. 
This simplifies the mapping below and the evaluation of completion models (Section~\ref{sec:benchmark}), while mitigating annotation noise from misassigned reagents in the raw USPTO records. As a result, completion models must implicitly infer the role of each input molecule (e.g., reactant vs. reagent). Molecules that act as unchanged reagents must be reproduced on the product side to obtain a balanced reaction, which we refer to as reagent restoration.

\paragraph{Mapping and processing procedure}
To enable supervised learning, we construct an explicit mapping between USPTO reactions and their corresponding curated counterparts, because the original template-curation pipeline~\cite{joung2025electron} does not preserve links to the source data (USPTO) and relies on closed-source software (NameRxn~\cite{NextMove_NameRxn}). Our mapping procedure consists of the following steps. 

\begin{enumerate}
    \item Merging multi-step mechanistic reactions from the FlowER dataset~\cite{joung2025electron} into a single reaction (\texttt{reactants of the first step $\gg$ products of the final step}).
    \item Mapping: Successful if one specific USPTO reaction SMILES (\texttt{reactants.reagents $\gg$ products}) is fully contained in one specific FlowER reaction.
    \item Reintroducing stereochemistry (3D arrangement of atoms) from USPTO records, as FlowER does not include stereochemical information.
\end{enumerate}

As a result, we aligned \textbf{206,423} incomplete (USPTO) to atom-balanced (FlowER) reaction equations (see Appendix~\ref{app:mapping_details} for details). These aligned pairs define the reaction completion task in Section~\ref{sec:benchmark}, and are illustrated in Figure~\ref{fig:reaction_completion_example}. While the FlowER targets originate from expert-curated reaction templates, we independently assessed resulting targets at the record level. For 53 randomly sampled matched reactions, 51~(96.23\%) were chemically compatible with at least one of three independent completions by chemistry annotators (Appendix~\ref{app:target_audit}, with examples in Appendix~\ref{app:target_audit_cases}).

To characterize benchmark incompleteness beyond simple bookkeeping, we stratify CompleteRXN by missing-species type. Overall, 56.39\% of reactions contain at least one non-helper carbon-bearing omission, showing that the benchmark is not dominated by common helper-species restoration. A detailed breakdown is provided in Appendix~\ref{app:missing_species}.

\section{The CompleteRXN Benchmark}
\label{sec:benchmark}
\begin{figure}[t]
\definecolor{myblue}{RGB}{0,114,178}
\definecolor{mygreen}{RGB}{0,158,115}
    \centering
    \begin{tikzpicture}[
        font=\small,
        highlightByproduct/.style={
            rounded corners=2pt,
            line width=1.1pt,
            mygreen
        },
        annByproduct/.style={
            font=\scriptsize,
            text=mygreen,
            align=center
        },
        highlightReagent/.style={
            rounded corners=2pt,
            line width=1.1pt,
            myblue
        },
        annReagent/.style={
            font=\scriptsize,
            text=myblue,
            align=center
        }
    ]

        % -------------------------
        % Input label and image
        % Keep original width to preserve molecule size
        % -------------------------
        \node[anchor=west, text=gray] at (0,0.25)
            {Input: incomplete USPTO reaction};

        \node[anchor=north west, inner sep=0] (input) at (0,0) {
            \includegraphics[width=0.72\linewidth]{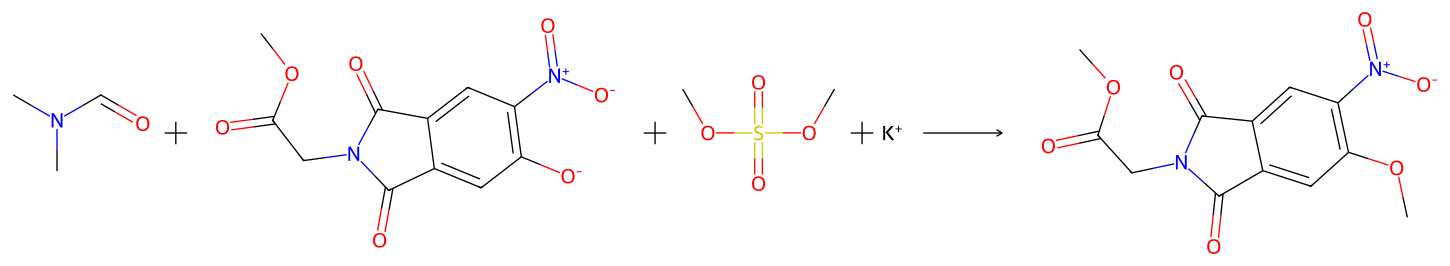}
        };

        % -------------------------
        % Output label and image
        % Reduced vertical spacing
        % Keep original width to preserve molecule size
        % -------------------------
        \node[anchor=west, text=gray] at (0,-1.90)
            {Output: curated, atom-balanced reaction};

        \node[anchor=north west, inner sep=0] (output) at (0,-2.10) {
            \includegraphics[width=0.952\linewidth]{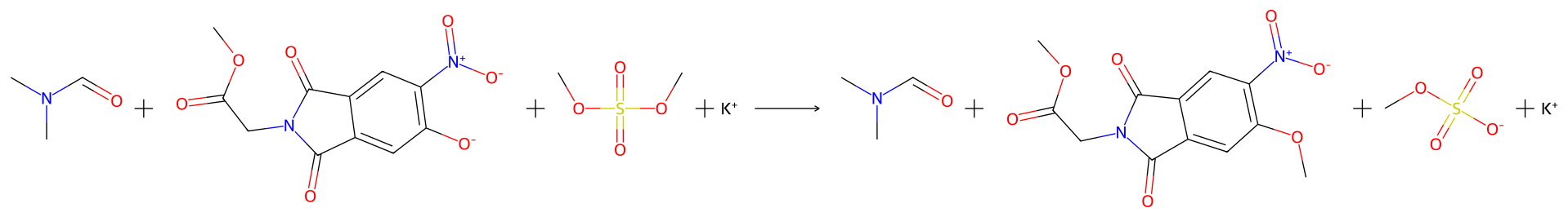}
        };

        % -------------------------
        % Highlight added / restored species
        % Coordinates may need small final tuning
        % -------------------------

        % byproduct
        \draw[highlightReagent]
            (7.08,-2.5) rectangle (8.15,-3.45);

        \node[annReagent, anchor=north] at (7.6,-3.45)
            {restored \\reagent};

        % 4) reagent ion
        \draw[highlightByproduct]
           (11.7,-2.5) rectangle (12.85,-3.45);

        \node[annByproduct, anchor=south] at (12.25,-2.5)
            {byproduct};

        % 4) reagent ion
        \draw[highlightReagent]
           (13.03,-2.5) rectangle (13.27,-3.45);

        \node[annReagent, anchor=north] at (13.2,-3.45)
            {restored \\ion};

        % -------------------------
        % Prediction / completion arrow
        % -------------------------
        \draw[
            ->,
            line width=0.6pt,
            black
        ]
            (10.5,-0.5)
            .. controls (11.8,-1.0) and (11.8,-1.5) ..
            (11.2,-1.9);
        
        \node[
            font=\scriptsize,
            align=center,
            anchor=west
        ] at (11.6,-1.25)
            {Completion\\Task};

    \end{tikzpicture}

    \caption{
    Reaction completion task. An incomplete reaction from USPTO (top) is paired with its atom-balanced counterpart derived from the FlowER~\cite{joung2025electron} dataset (bottom). The model needs to predict missing species to transform the incomplete input into a chemically valid, atom-balanced reaction.
    }
    \label{fig:reaction_completion_example}
\end{figure}
We propose the following benchmarking framework for evaluating reaction completion models.

\paragraph{Task definition and representation}
We formulate reaction completion as a structured missing-data problem, mapping an incomplete reaction to a chemically valid, atom-balanced reaction, as illustrated in Figure~\ref{fig:reaction_completion_example}. We use the reaction SMILES representation (Section~\ref{sec:dataset}) for both input and output. Formally:

\begin{itemize}
    \item \textbf{Input:} Incomplete reaction (missing reactants, reagents, or byproducts)
    \item \textbf{Output:} Completed, atom-balanced reaction, either as a full equation or the missing molecules (Section~\ref{sec:models})
\end{itemize}

\begin{figure}[htbp]
    \centering
    \begin{subfigure}[b]{0.49\linewidth}
        \centering
        \includegraphics[width=\linewidth]{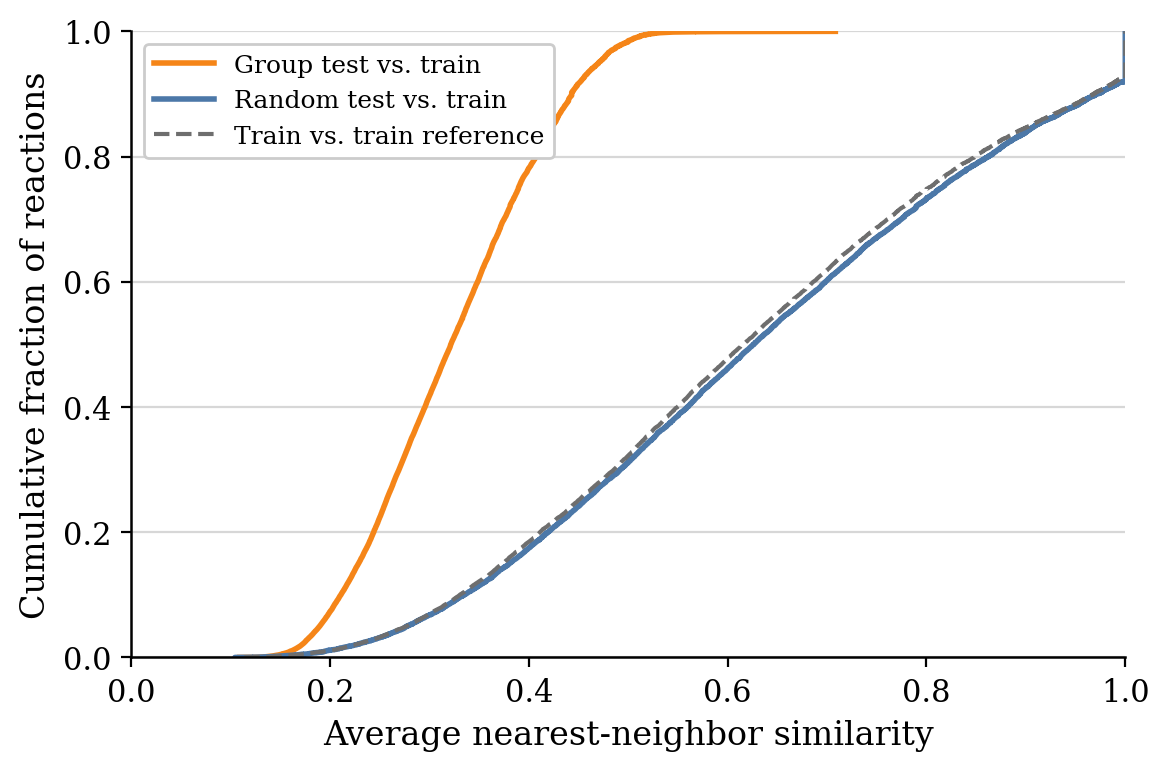}
        \caption{Group-based split (ii)}
        \label{fig:group_split_cdf}
    \end{subfigure}
    \hfill
    \begin{subfigure}[b]{0.49\linewidth}
        \centering
        \includegraphics[width=\linewidth]{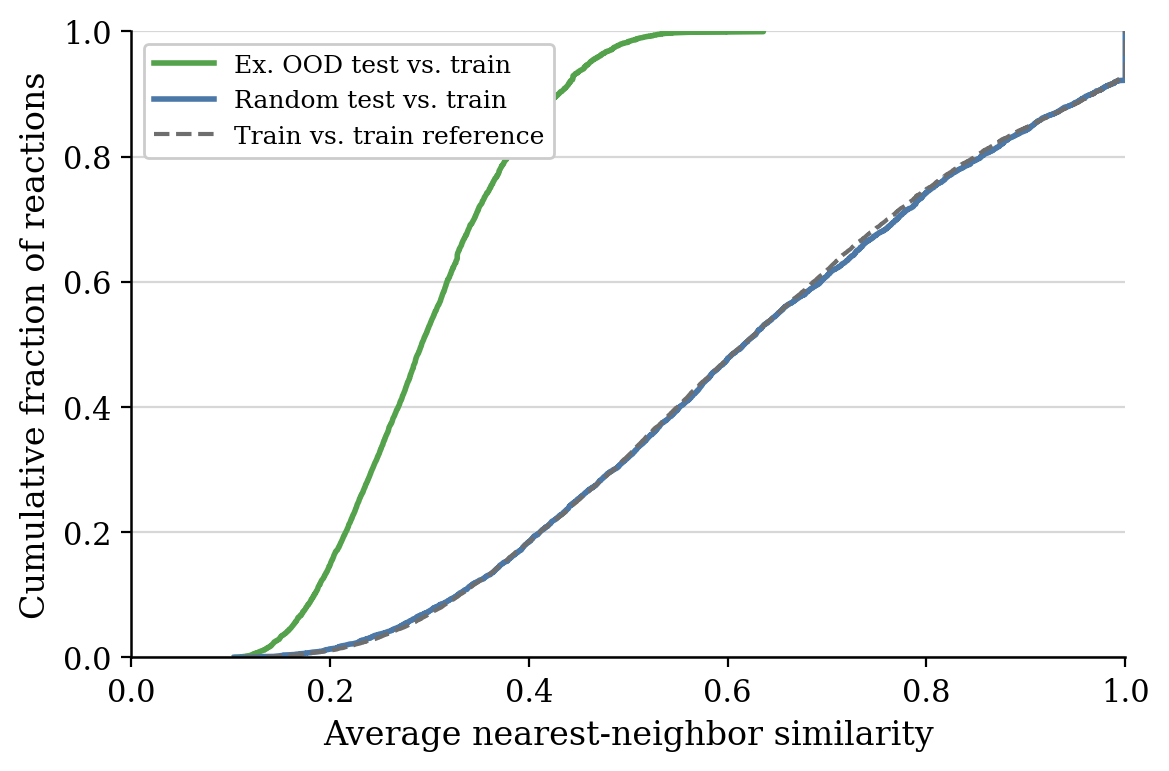}
        \caption{Extreme OOD split (iii)}
        \label{fig:extreme_ood_split_cdf}
    \end{subfigure}
    \caption{
    Distribution shift induced by the proposed data splits. We plot cumulative distributions of nearest-neighbor similarity in DRFP space. For each reaction, similarity is computed as the average Tanimoto similarity to its top-5 nearest neighbors in the comparison set. Compared with random splitting, both a) the group-based and b) extreme OOD splits shift leftward, indicating lower similarity between test reactions and the training set. The random split train-train curve is shown as a reference. }
    \label{fig:split_cdf_distribution_shift}
\end{figure}

\paragraph{Data splits.}
We evaluate models under three settings: (i) random splits, (ii) mechanism-aware group splits, and (iii) extreme out-of-distribution (OOD) splits. Random splits can overestimate performance because reactions with the same local transformation but different substituents (template-level redundancy) may appear in both training and test sets. To mitigate this, we construct group splits by representing reactions with DRFP fingerprints~\cite{probst2022DRFP}, grouping structurally similar reactions by fingerprint similarity, and assigning entire groups to train or test sets. This reduces leakage and tests generalization across reaction mechanisms. For the extreme OOD split, we further increase difficulty by selecting test groups that are both distant from the training data in fingerprint space and highly incomplete, measured by missing atom and missing carbon counts. This increases the average number of missing carbon atoms in the test set from 0.95 to 3.94. Figure~\ref{fig:split_cdf_distribution_shift} shows the resulting distribution shifts: compared with random splitting, the group-based and extreme OOD test sets are shifted toward lower (left) nearest-neighbor similarity to the training set, indicating reduced train-test similarity. Full construction details and more analysis are provided in Appendix~\ref{app:data_splits}.

\paragraph{Evaluation}

Evaluating reaction completion is challenging because multiple valid completions and equivalent reaction SMILES representations may exist. We therefore developed two metrics, \textbf{exact-match accuracy} and \textbf{equivalence accuracy}. Exact-match accuracy requires the predicted and target reactions to match exactly after canonicalizing molecule SMILES, respecting stoichiometry but ignoring molecule order within each reaction side. Exact match is a useful strict string-level metric, yet it is sensitive to chemical nuances in notation style.

Equivalence accuracy is our primary metric. It counts a prediction as correct if it is chemically equivalent to the target considering common representation ambiguities in reaction datasets. In particular, it tolerates alternative ionic representations, such as molecular \texttt{[Na][Cl]} vs. dissociated \texttt{[Na$^+$].[Cl$^-$]} salts, proton (\texttt{H$^+$}) redistribution within one side of a reaction equation, and a small number of alternative notations for frequent small molecules and reagents. Further details on the development of this metric are provided in Appendix~\ref{app:equiv_acc_metric}. 
RB and CRB are evaluated against the reference with stereochemistry. SynRBL (v1.0.5) removes stereochemistry during input standardization in its default pipeline, which we use here. We therefore compare SynRBL with the reference without stereochemistry. This more lenient comparison affects the 20.6\% of benchmark reactions whose reference contains stereochemistry. In only 1.0\% of reactions does the stereochemistry occur in the missing molecules themselves, otherwise, it occurs only in molecules already present in the input.

The RB and CRB models (Section~\ref{sec:models}) are fine-tuned from a pretrained Molecular Transformer and inherit its fixed token vocabulary. We therefore evaluate all methods on the same in-vocabulary test reactions, excluding examples with out-of-vocabulary tokens in the input or target. Across five repetitions, this excludes 7.71\%, 6.76\%, and 17.42\% of test observations for the random, group-based, and extreme-OOD splits, respectively. Unless stated otherwise, all CompleteRXN benchmark results reported in the main text and appendices use these in-vocabulary test reactions. Full test-set results without vocabulary filtering are reported in Appendix~\ref{app:vocab_filtering}.

\paragraph{Data and code availability} The dataset with all splits is available on \href{https://huggingface.co/datasets/Intelligent-molecular-systems/completeRXN-benchmark}{Hugging Face}, and the code for loading the benchmark, scoring predictions and reproducing our results on \href{https://github.com/Intelligent-molecular-systems/CompleteRxn-Benchmarking}{GitHub}.

\section{Baseline models}
\label{sec:models}
\paragraph{Machine learning models}
We train a Molecular Transformer model from a pretrained checkpoint (for forward prediction)~\cite{schwaller2019molecularTransformerFP} and transfer-learn reaction completion, as introduced in~\cite{van2024completing}. This does not introduce direct target leakage, because the completion models predict molecules that are absent from the USPTO records used for pretraining. The model hyperparameters are adopted from the Molecular Transformer, but we finetune the models for 50000 steps, where each step corresponds to one optimizer update over four dynamically sized 4,096-token minibatches. This corresponds to approximately 40 passes over the training set.

We use two different model variations, which vary not in training but inference: Reaction Balancer (RB)~\cite{van2024completing} and a novel Constrained Reaction Balancer (CRB). For the CRB, we compute a mask during beam-search decoding that is based on the atom-balance of the incomplete reaction. This enforces that the beam search decoding only finishes once the reaction is balanced (or the maximum tokens are reached). Further, this prohibits the prediction of atoms on the product side that are not on the completed reactant side. A detailed model description is provided in Appendix~\ref{app:CRB_methods}.

\paragraph{SynRBL}
We further evaluate SynRBL~\cite{phan2024reaction}, a recent algorithmic approach to infer missing molecules in unbalanced reactions. It combines two complementary strategies: (i) a rule-based approach for carbon-balanced reactions, which uses a library of chemical rules to add small missing molecules (e.g., $\mathrm{H}_2\mathrm{O}$, HCl, ions) based on atom count differences, and (ii) a separate approach for carbon-unbalanced reactions, which aligns reactant and product molecular graphs via maximum common subgraph (MCS) matching to identify unmatched fragments corresponding to missing compounds, and reconstructs them using heuristic merge and expand rules.

SynRBL is designed to propose chemically plausible rebalanced reactions rather than to reproduce a single reference completion. In the original SynRBL work, accuracy is defined as the fraction of proposed rebalanced reactions that are chemically correct, with correctness determined by manual expert inspection rather than by comparison to a fixed ground-truth reaction. This evaluation setup reflects the ambiguity of reaction completion: multiple chemically valid completions may exist for the same incomplete input. In our benchmark, however, predictions are compared against a specific curated target. SynRBL may therefore produce chemically plausible alternatives that do not exactly match this target. For this reason, exact-match accuracy is likely to underestimate SynRBL's performance. We primarily compare SynRBL using our equivalence-based metric, while noting that this metric may still miss some valid alternatives. As an additional method-specific diagnostic, we also report the fraction of SynRBL predictions assigned high confidence by the method itself.

\section{Results and discussions}
\label{sec:experiments}
\paragraph{ML models outperform the algorithmic baseline, yet also degrade under distribution shifts.}

The constrained reaction balancer (CRB) shows the best performance across all evaluation settings. We report both \textit{exact-match accuracy} and more importantly \textit{equivalence accuracy}. Across all split settings, the constrained model (CRB) consistently outperforms the unconstrained baseline (RB), achieving higher accuracy on the random, group-based, and extreme OOD splits (Table~\ref{tab:main_results}). For example, equivalence accuracy improves from 98.00\% to 99.28\% on the random split and from 83.34\% to 91.71\% on the extreme OOD split. These improvements are significant across all three split settings under paired per-reaction tests with multiple testing correction ($p_{\mathrm{adj}}<10^{-99}$; Appendix~\ref{app:CRBvsRB}). These results show that constrained beam search effectively guides decoding toward chemically correct solutions, both in-distribution and under distribution shift. We further stratify the random-split evaluation by missing-species type (Appendix~\ref{app:missing_species}). CRB remains highly accurate on reactions with non-helper carbon-bearing omissions, reaching 99.10\% equivalence accuracy on this subset. As an additional evaluation on an independently constructed benchmark, we apply CRB to SynRXN Complex~\cite{phan2026synrxn} (see Appendix~\ref{app:synrxn}).

The algorithmic baseline SynRBL generally underperforms the machine learning models on our benchmark dataset. 
Under the equivalence accuracy, SynRBL achieves substantially lower accuracy than the ML models across all splits (Table~\ref{tab:main_results}),with larger gaps on the extreme OOD split (35.06\% vs.\ 91.71\%). To account for the metric limitation (see Section~\ref{sec:models}), we also report the fraction of high-confidence predictions (self-reported confidence $>$ 50\%). Still, SynRBL balances a smaller fraction of reactions than CRB correctly predicts, demonstrating a performance gap. Further analysis is provided in Appendix~\ref{app:SynRBL_analysis}. 

\begin{figure}[hb]
    \centering
    \includegraphics[width=0.99\linewidth]{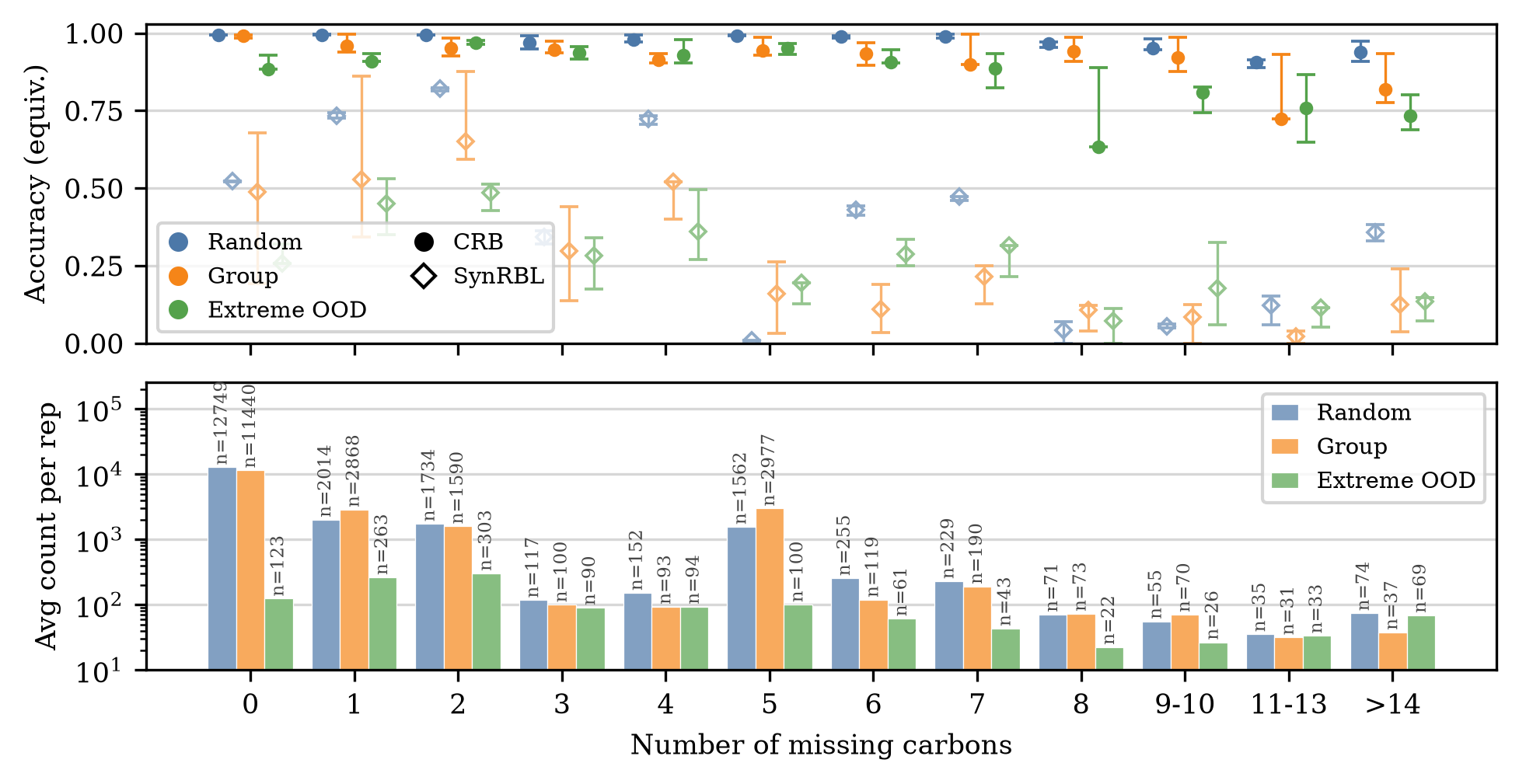}
    \caption{
    Equivalence accuracy over reaction incompleteness across random, group-based, and extreme OOD splits. 
    Top: Accuracy decreases with increasing difficulty; CRB consistently outperforms SynRBL. Error bars show the interquartile range (Q25--Q75).
    Bottom: Average number of test samples per bin (log scale), highlighting data scarcity in high-missing regimes.
    }
    \label{fig:accuracy_vs_difficulty}
\end{figure}

For all models, we observe a performance decrease as the evaluation setting becomes more challenging. Accuracy drops from random to group-based to extreme OOD splits and further declines with increasing numbers of missing carbon atoms (Figure~\ref{fig:accuracy_vs_difficulty}). This effect is amplified by the data distribution: highly incomplete reactions are underrepresented (Figure~\ref{fig:accuracy_vs_difficulty}, bottom), so the most difficult cases combine high intrinsic complexity with limited training coverage.

The impact of these factors differs between learned and algorithmic approaches. For the ML models, performance drops are driven by both increasing incompleteness and distribution shift. Yet, CRB consistently degrades less than RB across all settings, indicating that constrained beam search improves robustness under increasing task difficulty (see Figure~\ref{fig:appendix_CRB_vs_RB}). For SynRBL, the performance also decreases with increasing incompleteness and is lower on the OOD split. A key observation is the higher variability on the group-based split. This suggests that SynRBL performance depends strongly on the specific reaction mechanisms present in each fold, with some being well handled and others less so.
\begin{table}[ht]
\centering
\small
\caption{Extended evaluation of all models across splits. Results are evaluated on the in-vocabulary test reactions defined in Section~\ref{sec:benchmark} with equivalence accuracy as primary metric. For SynRBL, the fraction of predictions with confidence $>$ 50\% reflects internally confident solutions and is not directly comparable to accuracy.\shortstack{$^\dagger$}The column Balanced (\%) notes the top-1 balanced percentage for the RB/CRB and the solved percentage for SynRBL (without confidence threshold). All results are averaged over five folds (per split type) and $\pm$ in the result tables denotes standard deviation. SynRBL is evaluated without stereochemistry (see Section~\ref{sec:benchmark}).}
\label{tab:main_results}
\begin{tabular}{l l c c c c c}
\hline
Split & Model 
& \shortstack{Top-1 Acc\\(exact) (\%)} 
& \shortstack{Top-1 Acc\\(equiv.) (\%)} 
& \shortstack{Conf. $>$50\%\\(\%)} 
& \shortstack{Inv.\\SMILES (\%)} 
& \shortstack{Balanced$^\dagger$\\(\%)} \\
\hline

R
& RB & 97.19 $\pm$ 0.11 & 98.00 $\pm$ 0.11 & - & 0.14 $\pm$ 0.04 & 98.17 $\pm$ 0.11 \\
& CRB & 98.22 $\pm$ 0.06 & \textbf{99.28 $\pm$ 0.06} & - & 0.30 $\pm$ 0.05 & 99.57 $\pm$ 0.05 \\
& SynRBL & 5.82 $\pm$ 0.10 & 52.39 $\pm$ 0.15 & 73.78 $\pm$ 0.25 & - & 89.99 $\pm$ 0.29 \\

\hline

G
& RB & 92.76 $\pm$ 5.10 & 94.31 $\pm$ 4.23 & - & 0.33 $\pm$ 0.49 & 95.55 $\pm$ 2.55 \\
& CRB & 94.97 $\pm$ 4.09 & \textbf{97.32 $\pm$ 2.63} & - & 0.61 $\pm$ 0.41 & 99.06 $\pm$ 0.58 \\
& SynRBL & 2.95 $\pm$ 1.99 & 44.29 $\pm$ 28.02 & 79.72 $\pm$ 8.63 & - & 91.34 $\pm$ 4.71 \\

\hline

OOD
& RB & 80.98 $\pm$ 5.66 & 83.34 $\pm$ 5.05 & - & 1.06 $\pm$ 0.91 & 84.01 $\pm$ 4.94 \\
& CRB & 87.40 $\pm$ 4.14 & \textbf{91.71 $\pm$ 3.21} & - & 4.45 $\pm$ 1.80 & 93.49 $\pm$ 2.36 \\
& SynRBL & 6.66 $\pm$ 5.24 & 35.06 $\pm$ 17.10 & 59.14 $\pm$ 12.04 & - & 78.07 $\pm$ 4.82 \\

\hline
\end{tabular}
\end{table}
\paragraph{Errors are concentrated in a small subset of templates.}
Focusing on the random split to isolate template-level effects, we observe that errors are not uniformly distributed across templates. As introduced in Section~\ref{sec:data-analysis}, a subset of reactions in the USPTO dataset has associated template labels~\cite{schwaller2021mapping}, e.g., in the random testsets, 57.53\%~$\pm$~0.54\% of test reactions are labelled. Within this subset, the CRB solves a large fraction of templates perfectly: 513 out of 635 template hashes (80.79\%) achieve an accuracy of 1.0. At the same time, most errors are concentrated in a small number of templates: 50\% of all errors originate from just 28 templates (4.41\%), and 90\% from 99 templates (15.59\%). A similar concentration is observed for SynRBL, but at a lower overall performance level. Only 62 out of 635 templates (9.76\%) are solved perfectly, and errors are distributed across a larger set of templates: 50\% of errors come from 33 templates (5.20\%), while 90\% are found in 240 templates (37.80\%).

This analysis is limited to reactions with available template labels and therefore does not fully explain all observed errors. Nevertheless, together with the observation that missing atom patterns are highly consistent within templates (Section~\ref{sec:data-analysis}), these results suggest that reaction completion difficulty is largely determined at the template level. This implies that improving performance on a relatively small number of challenging templates could lead to substantial gains when scaling reaction completion.

\begin{figure}[ht]
    \centering

    % legend
    \includegraphics[width=0.6\linewidth]{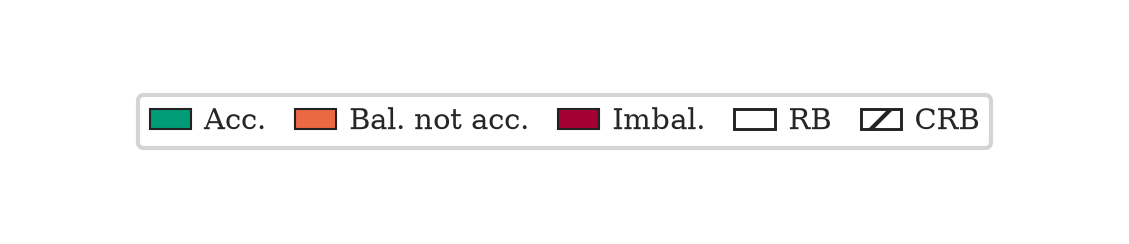}

    % three panels
    \begin{subfigure}[t]{0.372\linewidth}
        \centering
        \includegraphics[width=\linewidth]{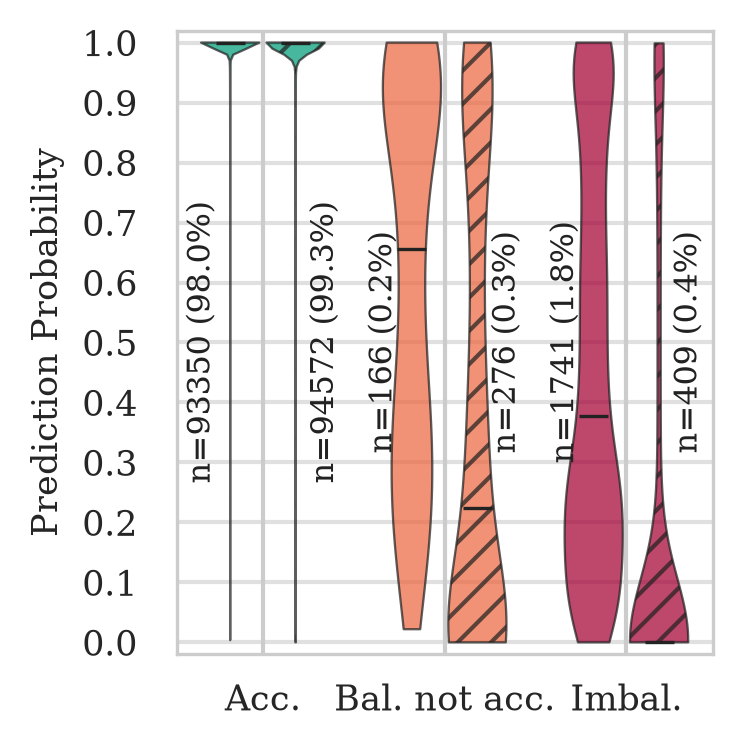}
        \vspace{-1.5em}
        \caption{Random ($\mathrm{n}=95{,}257$)}
    \end{subfigure}
    \hfill
    \begin{subfigure}[t]{0.3\linewidth}
        \centering
        \includegraphics[width=\linewidth]{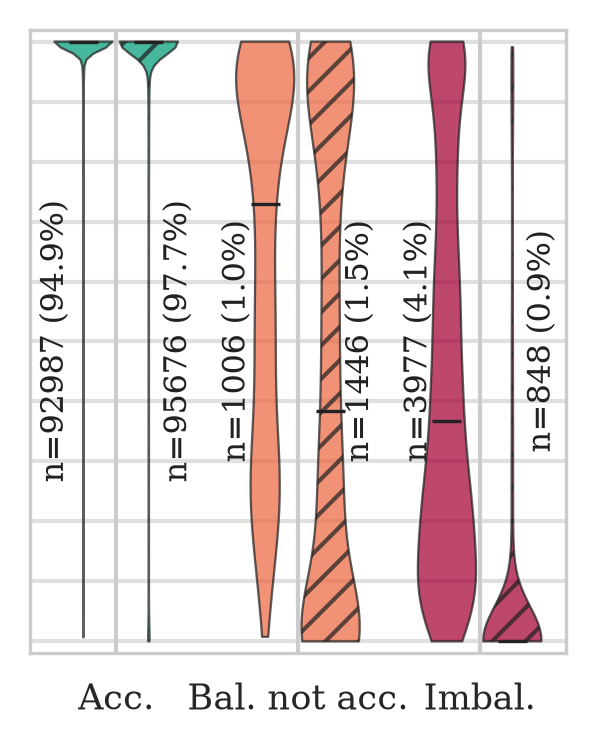}
        \vspace{-1.5em}
        \caption{Group ($\mathrm{n}=97{,}970$)}
    \end{subfigure}
    \hfill
    \begin{subfigure}[t]{0.3\linewidth}
        \centering
        \includegraphics[width=\linewidth]{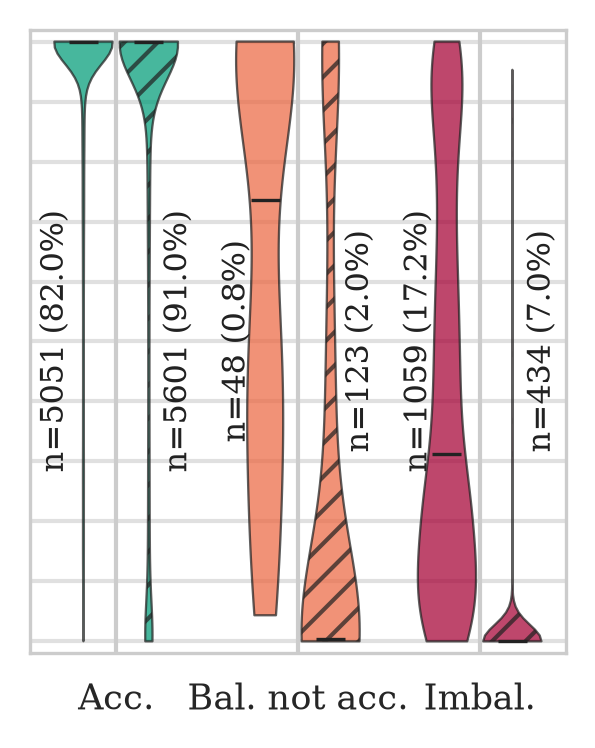}
        \vspace{-1.5em}
        \caption{Extreme OOD ($\mathrm{n}=6{,}158$)}
    \end{subfigure}

    \caption{Prediction probability distributions of accurate, balanced but inaccurate and unbalanced predictions for different test splits. 
    Accurate (equivalence acc.) predictions concentrate at high confidence, while unbalanced predictions occur mainly at low confidence. 
    Compared with RB, CRB reduces unbalanced outputs and shifts errors toward balanced predictions, including balanced but incorrect cases. 
    The effect is strongest on the harder splits.}
    \label{fig:confidence_predictions}
\end{figure}

\paragraph{ML confidence is informative but not sufficient for correctness.}
We use the sequence prediction probability during decoding as a confidence estimate. Figure~\ref{fig:confidence_predictions} shows top-1 prediction probabilities for accurate, balanced but incorrect, and unbalanced predictions. For both RB and CRB, accurate predictions concentrate at high confidence, while unbalanced predictions occur mostly at low confidence, with a clearer separation for CRB. Constrained decoding (CRB) strongly reduces the number of unbalanced outputs and shifts the error mass toward chemically balanced predictions, mostly toward accurate (green) but also toward balanced but incorrect (orange) predictions. This effect is most pronounced on the group and extreme OOD splits, where RB produces many unbalanced outputs but CRB removes a large fraction of them, consistent with Table~\ref{tab:main_results}. 

However, the remaining CRB errors are more often balanced but incorrect and therefore harder to detect than unbalanced outputs. Compared to RB, their confidence distribution shifts slightly toward lower values (see median lines in Figure~\ref{fig:confidence_predictions}), indicating that the constraint reduces some overconfident errors. However, a substantial fraction of these predictions still receives high confidence, showing that balanced but incorrect completions remain difficult to distinguish from correct ones based on prediction probability alone. 

\paragraph{Cross-method agreement shows high precision.}
Agreement between CRB and SynRBL provides a substantially stronger reliability signal than individual model confidence. On the benchmark test sets, the two methods agree on (i) 52.26\%, (ii) 46.21\%, and (iii) 37.84\% of predictions, depending on the split. Within these agreement subsets, 99.94\% on the random split (i), 99.82\% on the group split (ii), and 99.53\% on the extreme OOD split (iii) also match the benchmark reference under our equivalence metric. The few agreements that do not match the benchmark reference also illustrate remaining ambiguity in the equivalence evaluation (Appendix~\ref{app:agreement}). Overall, this indicates that cross-method agreement defines a smaller but highly reliable subset of completions. At the same time, agreement is a conservative criterion: the benchmark evaluation uses a single reference completion per reaction and may not account for alternative valid completions. Consequently, agreement prioritizes precision over coverage, potentially excluding chemically plausible but non-matching solutions. This motivates its use as a high-precision selection strategy when applying models to data without ground truth.

%Overall, confidence is useful for identifying invalid predictions, but not sufficient for certifying correctness. This limitation becomes more pronounced once balance is enforced, and additional signals are needed in practice, especially in settings without ground truth (see next paragraph).  

\paragraph{Benchmark--real-world performance gap.}
While the CompleteRXN benchmark enables controlled evaluation under realistic and structured missing-data patterns, it represents a curated, template-aligned subset of the USPTO dataset. In contrast, the full USPTO data contains rarer reaction types and additional noise, including incorrect or additional molecules, often requiring both completion and correction. Consequently, benchmark performance is expected to overestimate real-world robustness.

This gap becomes evident when applying both methods to a large-scale uncurated dataset. SynRBL produces balanced reactions for 75.45\% (665{,}774/882{,}357) of inputs, whereas CRB yields balanced outputs for only 49.31\% (435{,}124/882{,}357). This reflects the increased difficulty of generating chemically consistent completions under realistic noise, as well as the reliance of CRB on clean, template-aligned patterns (see also Appendix~\ref{app:uncurated_results}).

To obtain a reliable subset of predictions in the absence of ground truth, we use cross-method agreement, motivated by its high precision in the benchmark. Both methods produce balanced outputs for 384,580 reactions, of which 268,839 (69.90\%) agree under our equivalence metric, corresponding to 30.47\% of the full remaining USPTO dataset. Since no validated ground truth reference is available for these reactions, agreement should be interpreted as a conservative estimate of completion success and as a proxy for reliability, rather than as an accuracy estimate. We discuss additional filtering based on model-specific confidence scores in Appendix~\ref{app:uncurated_results}.

\section{Toward realistic reaction completion}\label{sec:limitations}

High equivalent accuracies in our benchmark setting suggest that the models can enable meaningful completion of large parts of the USPTO dataset. Nonetheless, several limitations persist, which are discussed below together with our ongoing research efforts to address them.

\paragraph{Limitations of benchmark data and evaluation strategies.}
The main limitation of our benchmark data is that it does not fully capture the complexity of real-world reaction data. While the dataset provides high-quality incomplete–complete reaction pairs, it is derived from common template-matched and curated reactions and therefore underrepresents rare, noisy, or unconventional reaction types. As a result, the benchmark primarily captures completion under structured and template-consistent missing patterns, representing a more controlled setting than the full USPTO dataset.

In addition, the dataset excludes important sources of real-world noise, such as incorrectly extracted or inserted molecules. Consequently, the evaluated models address completion, but not correction, even though a combination of both will be required in practice. Strong benchmark performance should therefore not be interpreted as equivalent robustness on raw USPTO data: in practice, performance drops substantially under realistic noise (Sec.~\ref{sec:experiments}, Suppl.~\ref{app:uncurated_results}), highlighting the gap between controlled evaluation and real-world applicability. Moreover, model behavior diverges in this setting. CRB achieves higher accuracy on the benchmark, but it degrades more strongly on uncurated data. SynRBL produces a higher fraction of balanced reactions on the uncurated data. This reflects complementary strengths of machine learning and rule-based approaches: ML-based completion is particularly effective when the target chemistry and incompleteness patterns are represented in the training data, whereas rule-based methods can remain useful for unfamiliar transformations when their rules apply. In applications where reliability is prioritized, agreement between CRB and SynRBL can provide a conservative selection criterion, while disagreements can be flagged for further inspection.

A second limitation concerns the ambiguity of evaluation. Even with the equivalence-based metric introduced here, valid alternative completions may not always be counted as correct, particularly when multiple plausible outcomes exist. This affects all methods and is especially relevant when comparing machine learning models, which tend to learn dataset-specific completion patterns, to non-data-driven approaches such as SynRBL, which may propose alternative chemically valid solutions. In the absence of ground truth, cross-method agreement provides a high-precision signal, but applies only to a limited subset of reactions, indicating that reliable large-scale validation remains challenging.

Finally, the use of expert-curated templates introduces additional biases. In particular, multiple procedural steps (e.g., basic hydrolysis followed by acidic workup) are often merged into a single reaction, requiring models to implicitly reconstruct missing temporal structure.

%A third limitation is that the most difficult cases remain challenging for all methods. Performance drops for reactions with high levels of incompleteness and under distribution shift. While constrained decoding improves robustness, highly incomplete reactions and challenging reaction mechanisms are still difficult for both learned and algorithmic approaches. In addition, we do not evaluate LLM-based agentic methods, which may be especially promising for such hard cases because they can combine external knowledge sources such as patent lookup, web search, and tool use. At present, however, due to their computational cost, LLM-based agentic approaches are not evaluated in this work. In practice, simpler and more efficient models are preferable when they achieve high accuracy and confidence, especially for large-scale reaction completion.

\paragraph{Outlook: Web-based manual curation}
These limitations point to the need for more realistic and challenging benchmark data. In particular, many reactions in the full USPTO require not only completion but also correction of errors. To address this, as a final part of this benchmark, we are currently collecting manual curations of particularly challenging reactions by chemistry annotators (see Appendix~\ref{app:expert_curation}) through a self-developed web-based framework. Once a reasonable fraction of curated reactions is achieved, this dataset will (i) provide more realistic data scarcity scenarios, (ii) focus on most challenging underrepresented reactions, (iii) go beyond completion toward joint completion and correction, and (iv) including the possibility of multiple valid outcomes.

%We are already applying this framework in an ongoing expert curation effort involving multiple contributors with organic chemistry expertise. This shows that the approach is practical at small scale and can be used to iteratively build more challenging and realistic benchmark datasets. We expect that similar workflows could also be adopted by other groups to curate domain-specific reaction datasets and to extend existing resources beyond simple completion toward correction and refinement. Such datasets, together with hybrid and agentic approaches, may provide an important next step toward reliable large-scale reaction completion.

\section{Conclusion}
We introduce a dataset and benchmark for completing incomplete chemical reaction equations and use it to compare two learned and one algorithmic approach. The constrained encoder–decoder model achieves strong performance and consistently outperforms its unconstrained version and an algorithmic baseline, although all methods degrade as reactions become more incomplete and more dissimilar from the training data. We further show that most benchmark reactions can be solved reliably, while errors are concentrated in a small set of challenging reaction types. We believe that continuous improvements on benchmark data quality as in the proposed manual curation efforts are essential for advancing reaction completion approaches. Our benchmark and analysis provide the foundation for developing advanced completion strategies through fair, reproducible, and diverse test settings.

Resulting large-scale atom- and charge-balanced reaction datasets will impact AI-driven chemistry, including applications such as synthesis planning and reaction prediction, and it will be the enabler for large mass-balanced reaction analysis and optimization, urgently needed for sustainability assessment. 

\begin{ack}
This work was supported by the Mainstreaming Open Science Fund of the Open Science Community Delft (OSCD). The fund financed the development of the web-based reaction curation framework and the hiring of chemistry annotators. We thank the chemistry annotators for the discussions and their careful curation work.
\end{ack}

\bibliographystyle{abbrv}
\bibliography{literature}

@article{phan2024reaction,
  title={Reaction rebalancing: a novel approach to curating reaction databases},
  author={Phan, Tieu-Long and Weinbauer, Klaus and G{\"a}rtner, Thomas and Merkle, Daniel and Andersen, Jakob L and Fagerberg, Rolf and Stadler, Peter F},
  journal={Journal of Cheminformatics},
  volume={16},
  number={1},
  pages={82},
  year={2024},
  publisher={Springer}
}

@article{joung2025electron,
  title={Electron flow matching for generative reaction mechanism prediction},
  author={Joung, Joonyoung F and Fong, Mun Hong and Casetti, Nicholas and Liles, Jordan P and Dassanayake, Ne S and Coley, Connor W},
  journal={Nature},
  volume={645},
  number={8079},
  pages={115--123},
  year={2025},
  publisher={Nature Publishing Group UK London}
}

@incollection{van2024completing,
  title={Completing partial reaction equations with rule and language model-based methods},
  author={van Wijngaarden, Matthijs and Vogel, Gabriel and Weber, Jana Marie},
  booktitle={Computer Aided Chemical Engineering},
  volume={53},
  pages={3139--3144},
  year={2024},
  publisher={Elsevier}
}

@article{schwaller2019molecularTransformerFP,
  title={Molecular transformer: a model for uncertainty-calibrated chemical reaction prediction},
  author={Schwaller, Philippe and Laino, Teodoro and Gaudin, Th{\'e}ophile and Bolgar, Peter and Hunter, Christopher A and Bekas, Costas and Lee, Alpha A},
  journal={ACS central science},
  volume={5},
  number={9},
  pages={1572--1583},
  year={2019},
  publisher={ACS Publications}
}

@article{coley2017reactionPrediction,
  title={Prediction of organic reaction outcomes using machine learning},
  author={Coley, Connor W and Barzilay, Regina and Jaakkola, Tommi S and Green, William H and Jensen, Klavs F},
  journal={ACS central science},
  volume={3},
  number={5},
  pages={434--443},
  year={2017},
  publisher={ACS Publications}
}

@article{coley2017MLforRS,
  title={Computer-assisted retrosynthesis based on molecular similarity},
  author={Coley, Connor W and Rogers, Luke and Green, William H and Jensen, Klavs F},
  journal={ACS central science},
  volume={3},
  number={12},
  pages={1237--1245},
  year={2017},
  publisher={ACS Publications}
}

@article{segler2017FPandRS,
  title={Neural-symbolic machine learning for retrosynthesis and reaction prediction},
  author={Segler, Marwin HS and Waller, Mark P},
  journal={Chemistry--A European Journal},
  volume={23},
  number={25},
  pages={5966--5971},
  year={2017},
  publisher={Wiley Online Library}
}

@article{du2024GenAIDrugDiscoveryReview,
  title={Machine learning-aided generative molecular design},
  author={Du, Yuanqi and Jamasb, Arian R and Guo, Jeff and Fu, Tianfan and Harris, Charles and Wang, Yingheng and Duan, Chenru and Li{\`o}, Pietro and Schwaller, Philippe and Blundell, Tom L},
  journal={Nature Machine Intelligence},
  volume={6},
  number={6},
  pages={589--604},
  year={2024},
  publisher={Nature Publishing Group UK London}
}

@article{chen2026LCACrystal,
  title={Life cycle assessment for all organic chemicals},
  author={Chen, Shaohan and Langhorst, Tim and N{\"o}hl, Julian and Oberschelp, Christopher and Pillich, Martin and Schilling, Johannes and Bardow, Andr{\'e}},
  journal={arXiv preprint arXiv:2603.15686},
  year={2026}
}

@article{weber2021chemicaldataintelligence,
  title={Chemical data intelligence for sustainable chemistry},
  author={Weber, Jana M and Guo, Zhen and Zhang, Chonghuan and Schweidtmann, Artur M and Lapkin, Alexei A},
  journal={Chemical Society Reviews},
  volume={50},
  number={21},
  pages={12013--12036},
  year={2021},
  publisher={Royal Society of Chemistry}
}

@article{ioannou2021process,
  title={Process modelling and life cycle assessment coupled with experimental work to shape the future sustainable production of chemicals and fuels},
  author={Ioannou, Iasonas and D'Angelo, Sebastiano Carlo and Gal{\'a}n-Mart{\'\i}n, {\'A}ngel and Pozo, Carlos and P{\'e}rez-Ram{\'\i}rez, Javier and Guill{\'e}n-Gos{\'a}lbez, Gonzalo},
  journal={Reaction Chemistry \& Engineering},
  volume={6},
  number={7},
  pages={1179--1194},
  year={2021},
  publisher={Royal Society of Chemistry}
}

@article{blanco2024machine,
  title={Machine learning to support prospective life cycle assessment of emerging chemical technologies},
  author={Blanco, CF and Pauliks, N and Donati, F and Engberg, N and Weber, J},
  journal={Current Opinion in Green and Sustainable Chemistry},
  volume={50},
  pages={100979},
  year={2024},
  publisher={Elsevier}
}

@article{zhang2024hybridcompletingreactions,
  title={Completing and balancing database excerpted chemical reactions with a hybrid mechanistic-machine learning approach},
  author={Zhang, Chonghuan and Arun, Adarsh and Lapkin, Alexei A},
  journal={ACS omega},
  volume={9},
  number={16},
  pages={18385--18399},
  year={2024},
  publisher={ACS Publications}
}

@phdthesis{lowe2012extraction,
  title={Extraction of chemical structures and reactions from the literature},
  author={Lowe, Daniel Mark},
  year={2012}
}

@article{probst2022DRFP,
  title={Reaction classification and yield prediction using the differential reaction fingerprint DRFP},
  author={Probst, Daniel and Schwaller, Philippe and Reymond, Jean-Louis},
  journal={Digital discovery},
  volume={1},
  number={2},
  pages={91--97},
  year={2022},
  publisher={Royal Society of Chemistry}
}

@misc{NextMove_NameRxn,
  author       = {{NextMove Software}},
  title        = {{NameRxn}},
  year         = {2026},
  url          = {https://www.nextmovesoftware.com/namerxn.html},
  note         = {Accessed: 2026-04-05}
}

@article{jin2017USPTOMIT,
  title={Predicting organic reaction outcomes with weisfeiler-lehman network},
  author={Jin, Wengong and Coley, Connor and Barzilay, Regina and Jaakkola, Tommi},
  journal={Advances in neural information processing systems},
  volume={30},
  year={2017}
}

@article{schwaller2018USPTOSTEREO,
  title={“Found in Translation”: predicting outcomes of complex organic chemistry reactions using neural sequence-to-sequence models},
  author={Schwaller, Philippe and Gaudin, Theophile and Lanyi, David and Bekas, Costas and Laino, Teodoro},
  journal={Chemical science},
  volume={9},
  number={28},
  pages={6091--6098},
  year={2018},
  publisher={Royal Society of Chemistry}
}

@article{schwaller2021mapping,
  title={Mapping the space of chemical reactions using attention-based neural networks},
  author={Schwaller, Philippe and Probst, Daniel and Vaucher, Alain C and Nair, Vishnu H and Kreutter, David and Laino, Teodoro and Reymond, Jean-Louis},
  journal={Nature Machine Intelligence},
  volume={3},
  number={2},
  pages={144--152},
  year={2021},
  publisher={Nature Publishing Group}
}

@article{schneider2016s,
  title={What’s what: The (nearly) definitive guide to reaction role assignment},
  author={Schneider, Nadine and Stiefl, Nikolaus and Landrum, Gregory A},
  journal={Journal of chemical information and modeling},
  volume={56},
  number={12},
  pages={2336--2346},
  year={2016},
  publisher={ACS Publications}
}

@article{schneider2015development,
  title={Development of a novel fingerprint for chemical reactions and its application to large-scale reaction classification and similarity},
  author={Schneider, Nadine and Lowe, Daniel M and Sayle, Roger A and Landrum, Gregory A},
  journal={Journal of chemical information and modeling},
  volume={55},
  number={1},
  pages={39--53},
  year={2015},
  publisher={ACS Publications}
}

@article{phan2026synrxn,
  title={SynRXN: An Open Benchmark and Curated Dataset for Computational Reaction Modeling},
  author={Phan, Tieu-Long and Nguyen Song, Nhu-Ngoc and Stadler, Peter F},
  journal={Scientific Data},
  volume={13},
  number={1},
  pages={625},
  year={2026},
  publisher={Nature Publishing Group UK London}
}

@article{zipoli2024completion,
  title={Completion of partial chemical equations},
  author={Zipoli, Federico and Ayadi, Zeineb and Schwaller, Philippe and Laino, Teodoro and Vaucher, Alain C},
  journal={Machine Learning: Science and Technology},
  volume={5},
  number={2},
  pages={025071},
  year={2024},
  publisher={IOP Publishing}
}
%\section*{References}
%\medskip
%{
%\small
%[1] Lalala
%}

%%%%%%%%%%%%%%%%%%%%%%%%%%%%%%%%%%%%%%%%%%%%%%%%%%%%%%%%%%%%
\newpage
\appendix
\counterwithin{figure}{section}
\counterwithin{table}{section}

\renewcommand{\thefigure}{\thesection\arabic{figure}}
\renewcommand{\thetable}{\thesection\arabic{table}}
\section{Additional dataset analysis and construction details}

The USPTO reaction dataset is highly incomplete: across commonly used subsets, fewer than 10\% of reactions are atom- and charge-balanced. Missing information predominantly consists of small molecules (e.g., H$_2$O, HCl), but larger carbon-containing fragments also occur, especially in less filtered subsets.

\begin{figure}[htbp]
    \centering
    \begin{subfigure}[t]{0.49\linewidth}
        \centering
        \includegraphics[width=\linewidth]{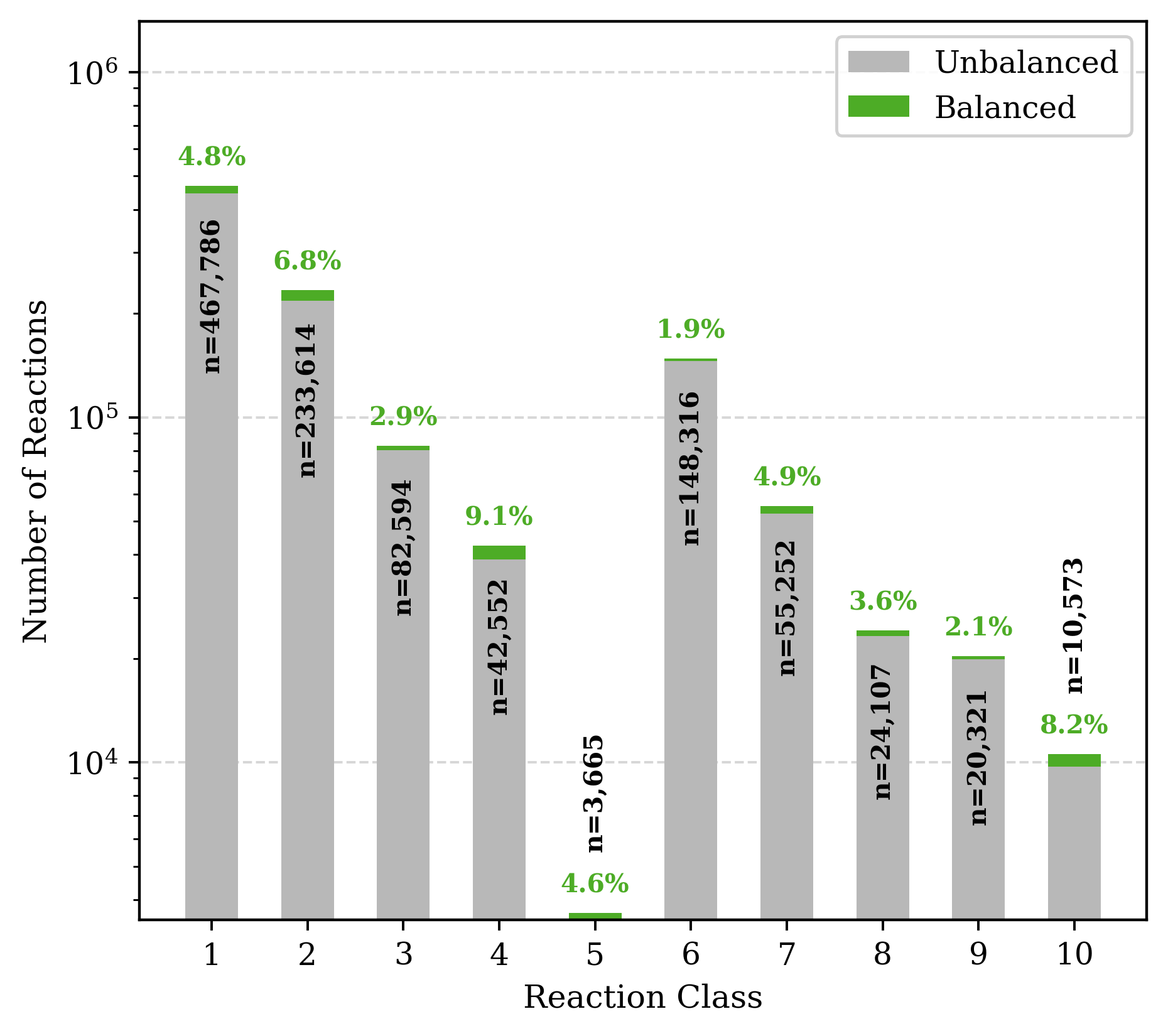}
        \caption{Balanced vs.\ unbalanced reactions per class}
    \end{subfigure}
    \hfill
    \begin{subfigure}[t]{0.49\linewidth}
        \centering
        \includegraphics[width=\linewidth]{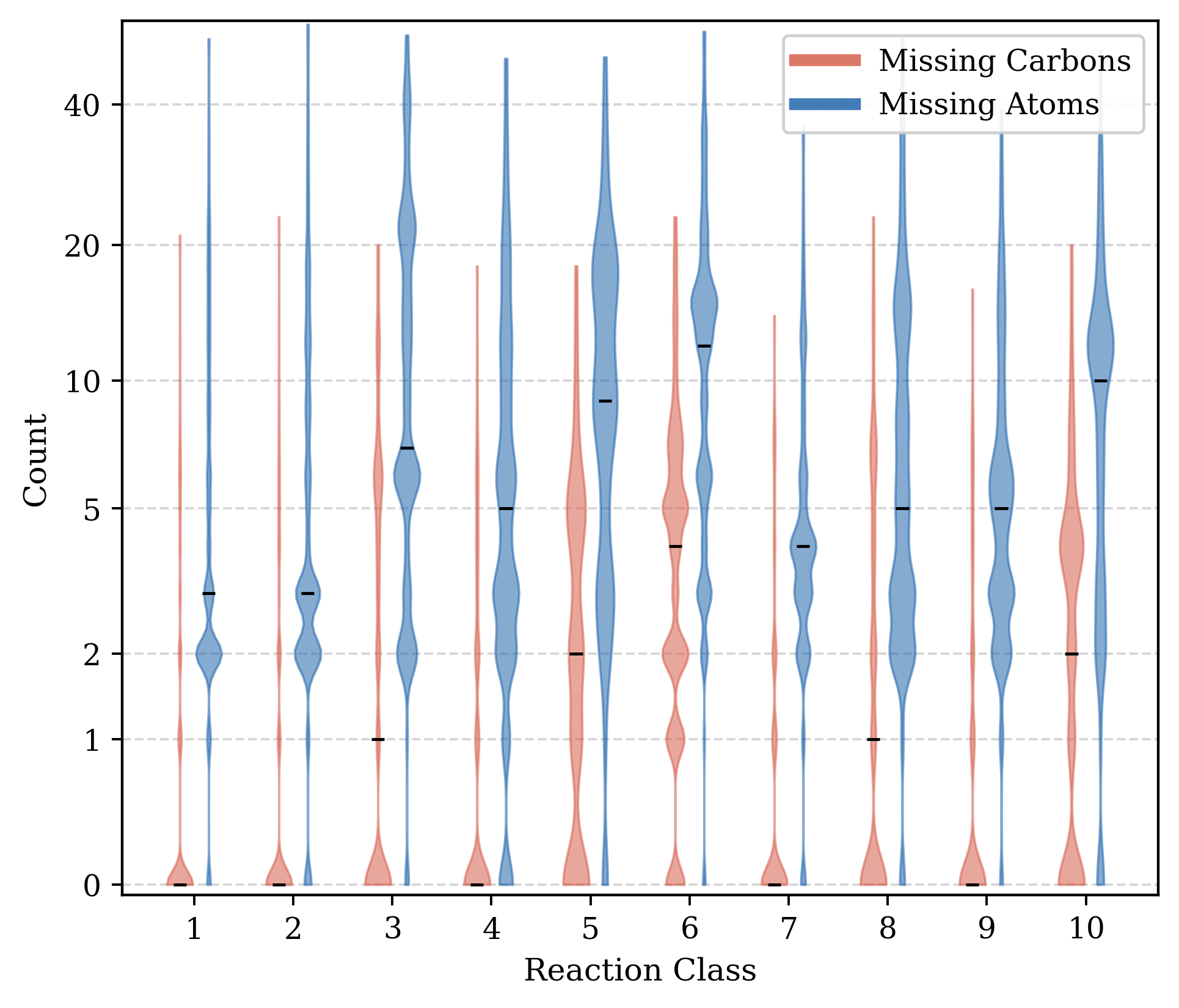}
        \caption{Missing atoms and carbons per class}
    \end{subfigure}
    
    \caption{Incompleteness patterns across the 10 main reaction classes in the USPTO dataset (1: heteroatom alkylation and arylation, 2: acylation and related processes, 3: C–C bond formation, 4: heterocycle formation, 5: protections, 6: deprotections, 7: reductions, 8: oxidations, 9: functional group interconversion (FGI), 10: functional group addition (FGA)). 
    (a) Reaction counts per class, split into atom-balanced and unbalanced reactions. Only a small fraction of reactions are balanced across all classes, with variation in class size.
    (b) Distribution of missing atoms and missing carbon atoms per class (log-scaled). While many reactions are missing only a few atoms, several classes show broader distributions and higher counts of missing carbon atoms, indicating increased structural complexity and variability in completion difficulty.}
    
    \label{fig:class_incompleteness_patterns}
\end{figure}

To analyze incompleteness across reaction classes, we assign reactions to 10 main reaction classes as introduced in \cite{schneider2015development}. Because these labels are not readily available we train a classifier on DRFP features of the USPTO 50k subsets~\cite{schneider2015development, schneider2016s}, and predict the class labels for the remaining part of the USPTO, as introduced in the DRFP paper~\cite{probst2022DRFP}. Figure~\ref{fig:class_incompleteness_patterns} shows that incompleteness varies systematically across reaction classes. While most reactions are missing only a few atoms, several classes exhibit broader distributions and larger numbers of missing carbon atoms, indicating increased structural complexity.

\begin{figure}[htbp]
    \centering
    \begin{subfigure}[t]{0.48\linewidth}
        \centering
        \includegraphics[width=\linewidth]{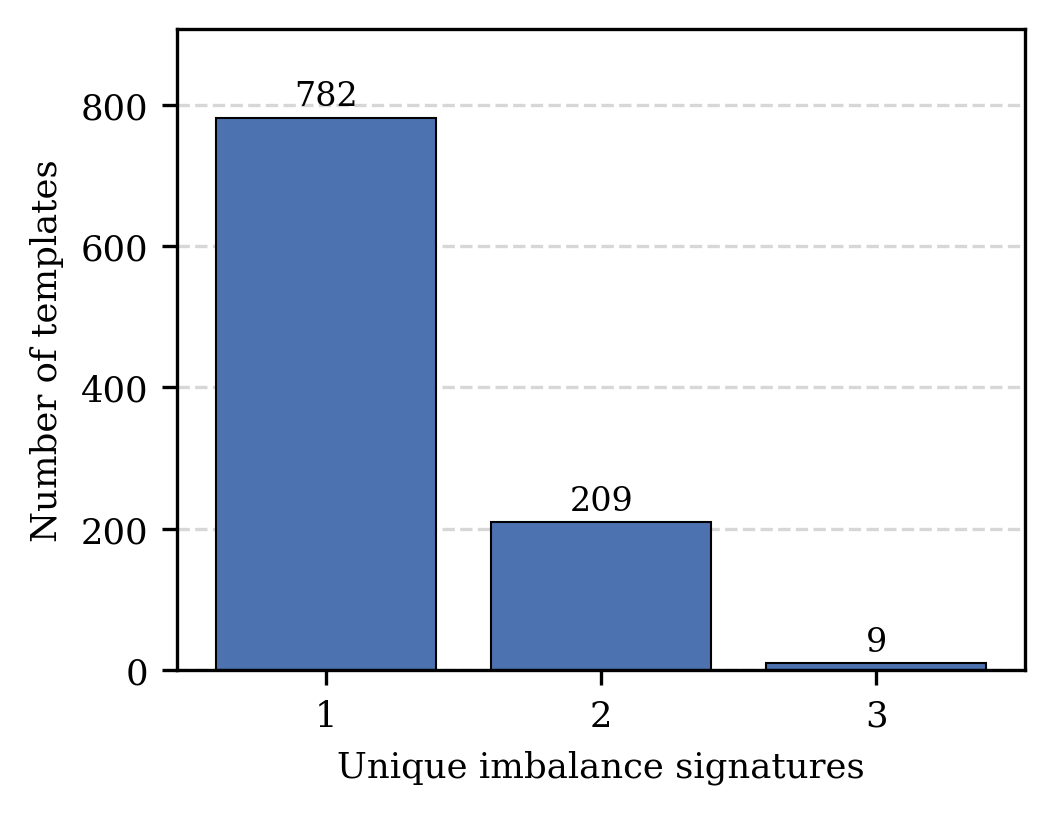}
        \caption{Unique atom-balance signatures per template}
    \end{subfigure}
    \hfill
    \begin{subfigure}[t]{0.48\linewidth}
        \centering
        \includegraphics[width=\linewidth]{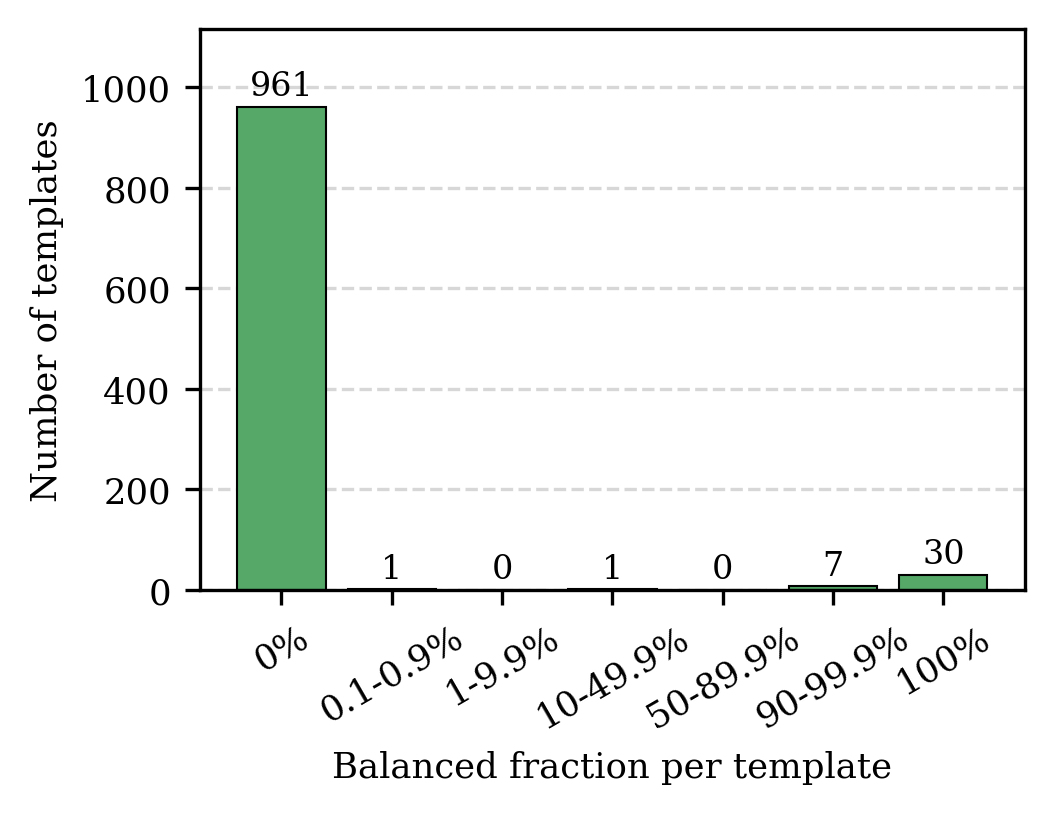}
        \caption{Fraction of balanced reactions per template}
    \end{subfigure}

    \vspace{0.5em}

    \begin{subfigure}[t]{0.48\linewidth}
        \centering
        \includegraphics[width=\linewidth]{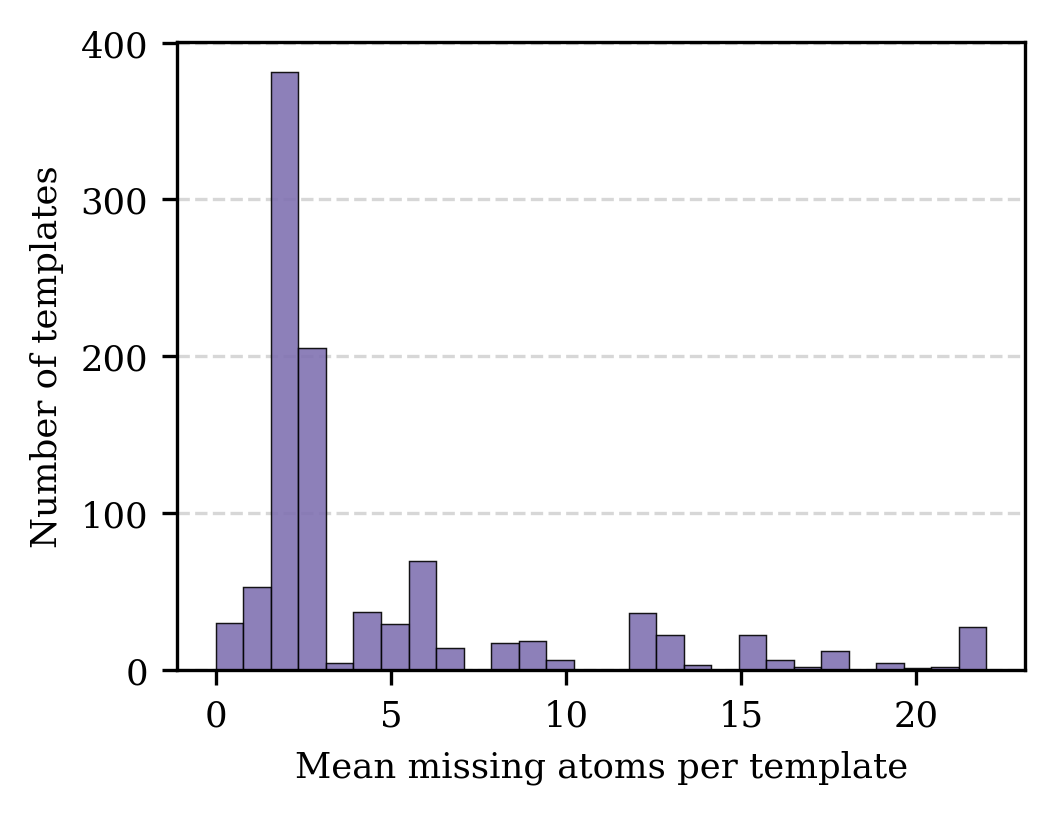}
        \caption{Mean missing atoms per template}
    \end{subfigure}
    \hfill
    \begin{subfigure}[t]{0.48\linewidth}
        \centering
        \includegraphics[width=\linewidth]{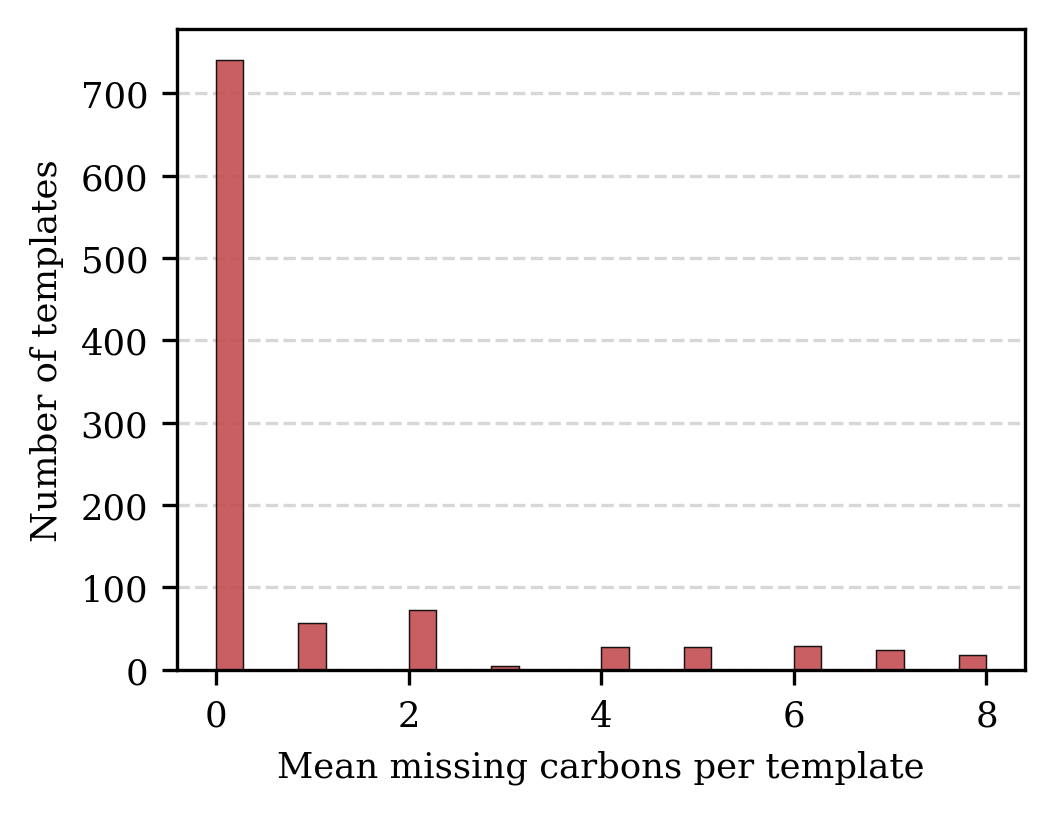}
        \caption{Mean missing carbons per template}
    \end{subfigure}

    \caption{Analysis of atom balance at the level of reaction templates in the USPTO 1k TPL subset~\cite{schwaller2021mapping}. 
(a) Number of unique atom-balance signatures per template. Most templates exhibit a single consistent atom-balance pattern across reactions within the same template. 
(b) Fraction of atom-balanced reactions per template. Templates are almost always either fully unbalanced or fully balanced, with very few mixed cases. 
(c) Distribution of the mean number of missing atoms per template, showing variation across templates. 
(d) Distribution of the mean number of missing carbon atoms per template. Most templates have no missing carbon atoms, while a smaller subset shows larger structural gaps, corresponding to more challenging completion cases.}
    \label{fig:template_completeness}
\end{figure}

At the template level (Figure~\ref{fig:template_completeness}), atom-balance patterns are highly consistent: most templates exhibit a single atom-balance signature and are either almost always balanced or almost always unbalanced. This suggests that reaction completion is largely governed by template-level structure.

\subsection{Mapping between USPTO subsets and FlowER}\label{app:mapping_details}
We construct mappings between commonly used USPTO subsets to identify overlapping reactions and ensure consistent analysis. The resulting mapping is released as metadata in the CompleteRXN dataset, see \url{https://huggingface.co/datasets/Intelligent-molecular-systems/completeRXN-benchmark} and \url{https://github.com/Intelligent-molecular-systems/CompleteRxn-Benchmarking}.

For the CompleteRXN construction, 240,192 of 1,088,780 USPTO records (22.06\%) matched at least one FlowER record. Of these, 206,617 mapped exactly one FlowER record to one USPTO record, 24,561 mapped one USPTO record to multiple FlowER records, and 9,014 mapped multiple USPTO records to one FlowER record. For the CompleteRXN benchmark, we retained only the strict one-to-one mappings. Final validity checks after stereochemistry backmapping remove a further 194 reactions: 147 whose stereochemistry could not be transferred unambiguously from the USPTO record, and 47 whose input, including reagents and stereochemistry, is not fully contained in the matched FlowER target. Overall, this yields 206,423 benchmark pairs.

%Camera ready addition
\subsection{Independent record-level audit of benchmark targets}
\label{app:target_audit}
The FlowER-derived targets originate from expert-curated reaction templates that are systematically applied to matching USPTO records. This provides expert curation at the template level, but does not constitute independent validation of every resulting reaction pair. We therefore additionally performed a record-level audit with chemistry annotators: seven chemistry students hired for this task (three graduate and four undergraduate students), all with completed training in organic chemistry. During our ongoing manual curation effort (Appendix~\ref{app:expert_curation}), one randomly selected matched USPTO reaction was embedded as a control in each batch of ten reactions, without access to the corresponding CompleteRXN target. Because individual completions can contain errors, we rely on several independent completions per reaction. Of the 103 embedded controls, 53 received exactly three independent completions, and we restrict the audit to these.

Across the 53 reactions (159 individual completions), the benchmark target matched at least one annotator completion under our equivalence metric (Appendix~\ref{app:equiv_acc_metric}) for 50/53 reactions (94.34\%) and the majority of annotator completions for 48/53 (90.57\%). Individually, 141 of the 159 completions (88.68\%) matched the benchmark target. Agreement is thus higher per reaction than per completion, as intended by the three independent completions.

We manually inspected the three reactions for which no annotator completion matched automatically. In one case, two annotators wrote molecular hydrogen in a SMILES notation that the metric does not normalize, and their completions were otherwise chemically equivalent to the benchmark target. The remaining two showed substantive differences in byproduct accounting. Combining automatic equivalence with this manual adjudication, the benchmark target was chemically compatible with at least one independent annotator completion in 51/53 reactions (96.23\%). Table~\ref{tab:target_audit_agreement} summarizes the agreement per control reaction, and Appendix~\ref{app:target_audit_cases} shows one control per agreement level together with all three controls without a match.

%Camera ready addition
\subsection{Empirical vs. synthetic incompleteness}\label{app:synthetic_corruption_comparison}
To compare empirical and synthetic missingness directly, we hold the same 206,423 balanced CompleteRXN targets fixed and construct incomplete inputs following prior work. Following Zhang et al.~\cite{zhang2024hybridcompletingreactions}, we hide each reactant and product molecule once, yielding 980,591 inputs with exactly one missing molecule. Following van Wijngaarden et al.~\cite{van2024completing}, we generate one incomplete input per target by randomly removing complete molecules up to 50\% of the reaction atoms. CompleteRXN instead uses the corresponding incomplete USPTO record, after identified reagents are moved to the reactant side.

\begin{table}[htbp]
\centering
\caption{Comparison of empirical CompleteRXN incompleteness with synthetic corruption schemes applied to the same balanced targets.}
\label{tab:synthetic_comparison}
\small
\setlength{\tabcolsep}{5pt}

\begin{tabular*}{\linewidth}{@{\extracolsep{\fill}}lrrrr@{}}
\toprule
\multicolumn{5}{l}{\textit{(a) Location of missing molecules}} \\
Input construction & Reactant only & Product only & Both & None / reagent \\
\midrule
CompleteRXN empirical & 0.06\% & 69.07\% & 20.57\% & 10.31\% \\
Zhang et al.~\cite{zhang2024hybridcompletingreactions} & 50.76\% & 49.24\% & 0.00\% & 0.00\% \\
van Wijngaarden et al.~\cite{van2024completing} & 30.75\% & 33.18\% & 36.07\% & 0.00\% \\
\midrule

\multicolumn{5}{l}{\textit{(b) Amount of missing information}} \\
Input construction & Mean missing mol. & $>1$ missing & Median missing atoms & \\
\midrule
CompleteRXN empirical & 1.75 & 50.43\% & 6 & \\
Zhang et al.~\cite{zhang2024hybridcompletingreactions} & 1.00 & 0.00\% & 20 & \\
van Wijngaarden et al.~\cite{van2024completing} & 1.78 & 57.39\% & 33 & \\
\midrule

\multicolumn{5}{l}{\textit{(c) Composition of missing species}} \\
Input construction & Helper & C-free & C-bearing & \\
\midrule
CompleteRXN empirical & 10.39\% & 27.68\% & 61.92\% & \\
Zhang et al.~\cite{zhang2024hybridcompletingreactions} & 5.36\% & 16.55\% & 78.09\% & \\
van Wijngaarden et al.~\cite{van2024completing} & 6.45\% & 22.82\% & 70.74\% & \\
\bottomrule
\end{tabular*}
\end{table}

The synthetic corruption schemes therefore produce systematically different missing-data distributions from the empirical USPTO omissions. Synthetic omissions are more symmetric across reaction sides, more frequently carbon-bearing, and involve larger missing fragments, whereas CompleteRXN is strongly biased toward product-side omissions and misses slightly more helper and carbon-free species. Importantly, these results do not imply that synthetic completion tasks are uniformly easier, but they characterize a different reaction-completion distribution.

The approaches also differ in how the balanced targets are constructed. In prior work~\cite{zhang2024hybridcompletingreactions,van2024completing}, balanced parent reactions are constructed in part using rule-based completion. This restricts the data to reactions that the rules can complete, and the targets follow the conventions of the balancing method. In CompleteRXN, 37.5\% of the reactions have an atom deficit that cannot be closed by a single common helper molecule, and in only 11.8\% of the reactions does the reference add exactly the atom deficit of the USPTO record. In the other cases, the reference also restores reagents, uses ionic forms or adds larger fragments. CompleteRXN instead derives its balanced targets from FlowER reactions based on expert-curated mechanistic templates and aligns these independently with the corresponding USPTO records. This distinction is separate from the input-corruption comparison above.

\section{Additional Method Details}
\subsection{Experimental details}\label{app:experiment_details}
We evaluate each split type using five independent fold realizations. 
Reported values in all results tables are means across these folds, $\pm$ denotes standard deviation, and figure error bars show the interquartile range (Q25--Q75).

\paragraph{Computational ressources} Runtimes were not systematically tracked, but we give the following estimates: Data processing and split generation were run on a single CPU core with 16 GB RAM. Dataset construction (FlowER–USPTO mapping, reagent identification, stereochemistry transfer, benchmark column generation) and split generation (DRFP fingerprinting, KNN-graph construction) each completed within a few hours.

SynRBL inference was run on a single CPU core (4 GB RAM) per job. Each chunk of the benchmark required up to 4 hours; the full benchmark was parallelised across chunks on a SLURM cluster.

CRB and RB training was performed on a single GPU (different kinds). Based on a rough throughput of ~10k steps per hour observed during development, fine-tuning from a pretrained checkpoint (50k steps) took approximately 5 hours per repetition. With 5 repetitions across 3 split types (random, group OOD, extreme OOD) for two model variants (RB and CRB), the total GPU time for all training runs was on the order of 50 GPU hours. CRB and RB inference was run on a single GPU (different kinds) using beam search with beam size 5. Each inference job (one split × one repetition) was allocated 3 hours; actual runtimes were not recorded but were generally well within this limit (CRB takes longer than RB). Scoring (metric computation on the beam outputs) ran on a single CPU core and completed within minutes per job.

\subsection{Data splits}\label{app:data_splits}
We evaluate models under three split scenarios: (i) \textit{random} splits, (ii) \textit{mechanism group-based} splits, and (iii) \textit{extreme out-of-distribution (OOD)} splits. Split (ii) tests generalization across reaction mechanisms while preventing highly similar reactions from appearing across train and test sets. Split (iii) additionally focuses on testing performance on reactions with a high number of missing atoms and carbons. All split types are repeated with 5 different folds. 

For splits (ii) and (iii), reactions are represented using the Differential Reaction Fingerprint (DRFP)~\cite{probst2022DRFP}, which captures the local chemistry at the reaction center while being largely invariant to peripheral scaffold variation. This aligns well with our task, as reactions sharing the same transformation often exhibit similar balance-completion patterns (Supplementary Fig.~\ref{fig:template_completeness}, top-left).

\begin{figure}[htbp]
    \centering
    \includegraphics[width=0.98\linewidth]{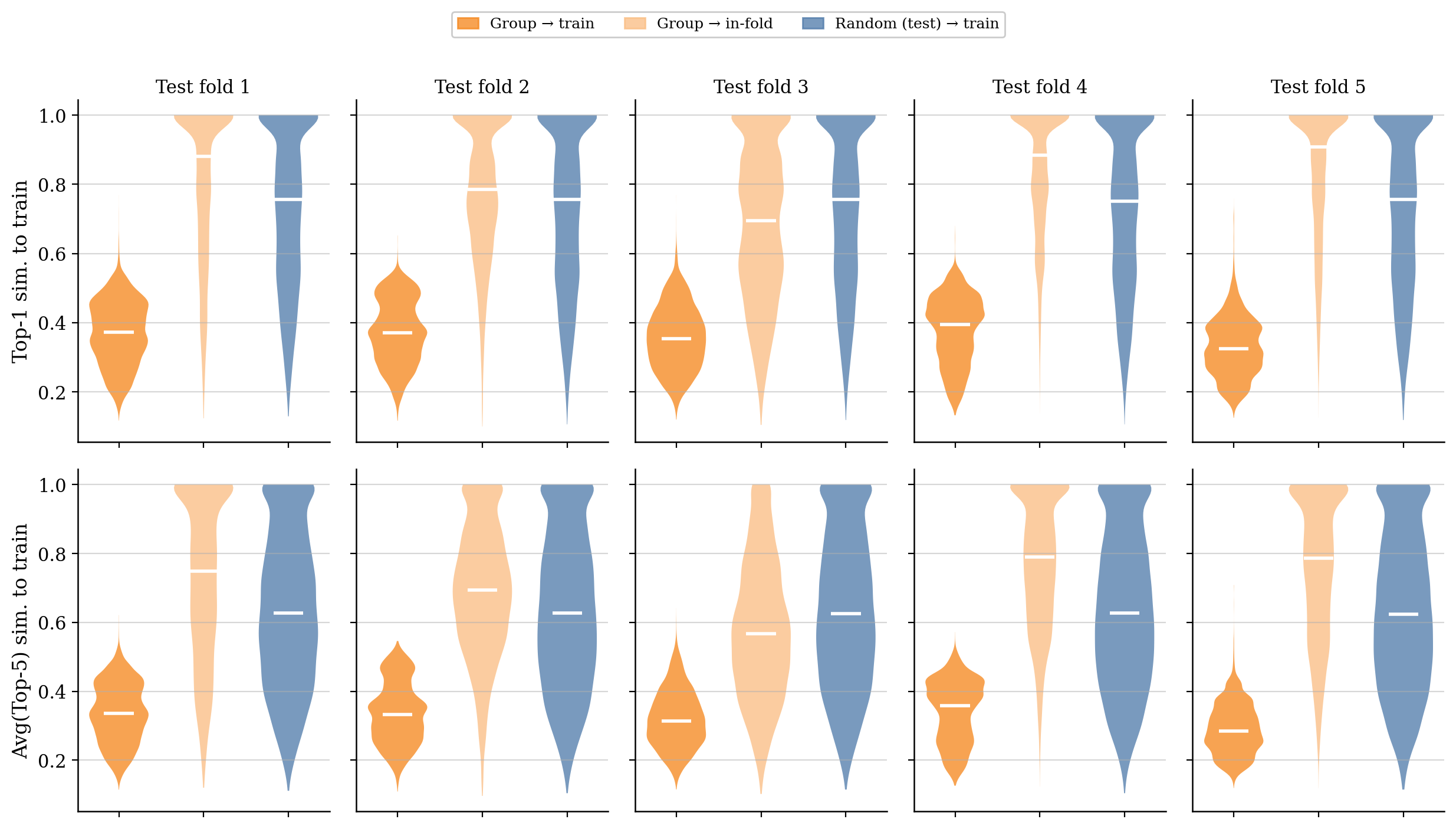}
    \caption{Group split: Tanimoto similarities for each test sample to the most similar neighbor (top) and five most similar (bottom) train reactions. There is a clear distribution difference in group-based vs. random split. High similarities within the test fold (especially folds 1,2, and 5) illustrate that within a test fold the reaction mechanisms are more similar than to the training folds.}
    \label{fig:violin_group_split}
\end{figure}

\begin{figure}[htbp]
    \centering
    \includegraphics[width=0.98\linewidth]{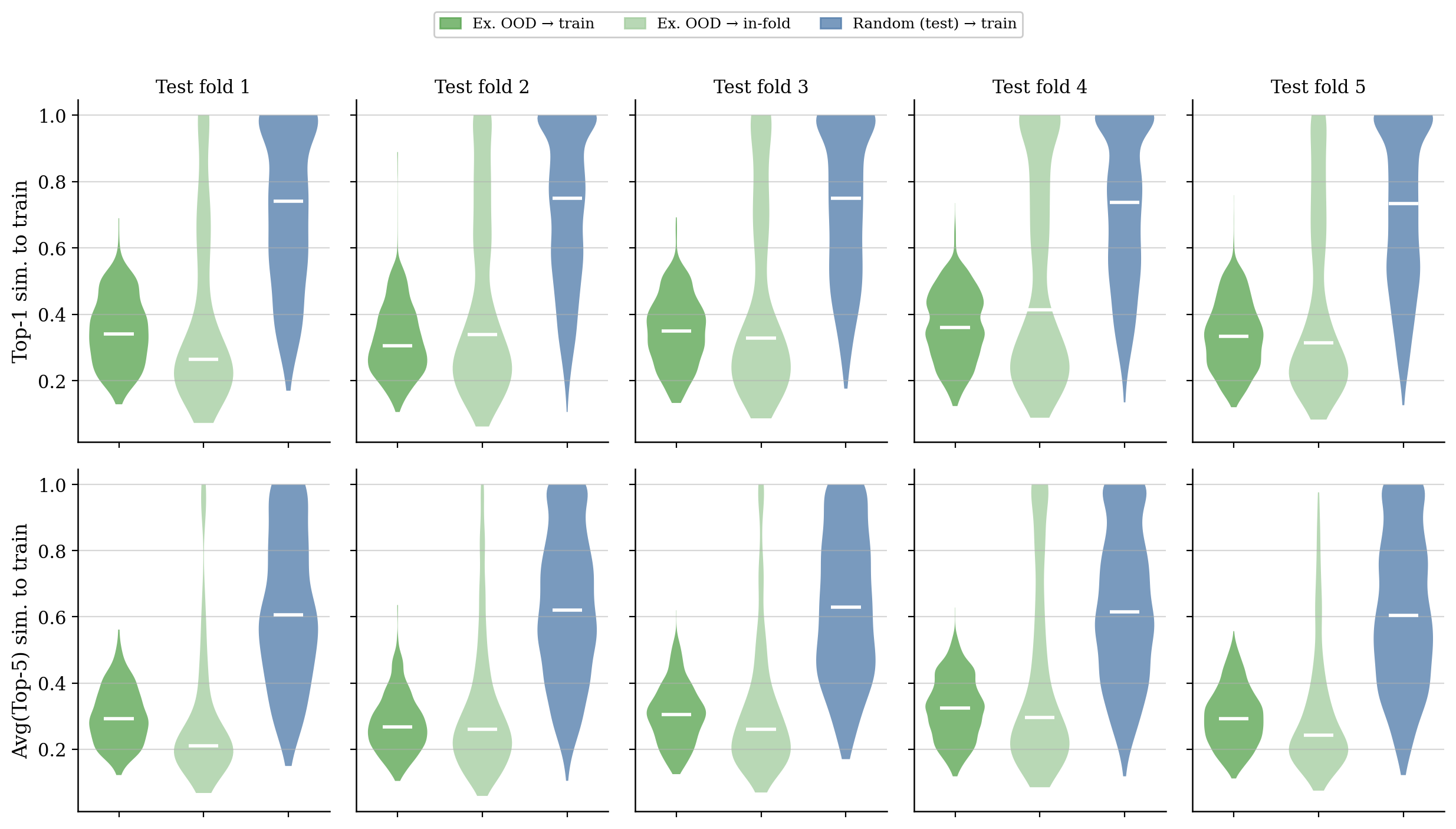}
    \caption{Extreme OOD split: Tanimoto similarities for each test sample to the most similar neighbor (top) and five most similar (bottom) train reactions. There is a clear distribution difference in extreme-OOD vs. random split. High similarities within the test fold (especially folds 1,2, and 5) illustrate that within a test fold the reaction mechanisms are significantly more similar than to the training folds.}
    \label{fig:violin_extreme_OOD}
\end{figure}

We first form \emph{leakage groups} of highly similar reactions. For efficiency, we construct a $k$-nearest neighbor graph ($k=100$) in a reduced DRFP space and compute exact Tanimoto similarity only for candidate neighbor pairs. A relatively large neighborhood size reduces the risk of missing highly similar reactions under approximate neighbor search. Edges with Tanimoto similarity above 0.55 are retained, and connected reactions define leakage groups. This threshold captures strong transformation-level similarity while avoiding overly large groups that would collapse multiple distinct transformation types. While these values are not uniquely optimal, they yield stable group structures and, importantly, produce substantial distribution shifts between train and test compared to random splitting (Fig.~\ref{fig:split_cdf_distribution_shift} and Suppl. Fig.~\ref{fig:violin_group_split} and~\ref{fig:violin_extreme_OOD}), resulting in a more challenging evaluation setting.

For the (ii) \textit{mechanism group-based} split, we compute one centroid per leakage group in the DRFP space and cluster all group centroids into 10 clusters of $\approx$10\% of the whole dataset using k-means. These clusters enable train/validation/test splits comparable to the random setting while preserving mechanistic separation.

For the \textit{extreme OOD} split, we again start from the leakage groups but restrict test groups to reactions that are both isolated in reaction space and highly incomplete. Isolation is defined by distance to the nearest group centroid, while incompleteness is measured by the number of missing atoms and missing carbon atoms. Selecting the most incomplete groups using a top-20\% criterion increases the average number of missing atoms from 4.5 (train) to 13.09 (test) and missing carbon atoms from 0.95 to 3.94. Because these criteria are stricter, test sets are smaller (400–2400 samples; see Section~\ref{sec:experiments}). To maintain comparable training sizes, we subsample the remaining data to 80\% of the full dataset, matching the train-set size used for splits (i) and (ii). 


\paragraph{Sensitivity analysis of split construction}
We first test the sensitivity of the extreme-OOD construction to the DRFP similarity threshold and neighborhood size $k$ using separate reconstructions of the splits. With $k=100$, DRFP thresholds of 0.50, 0.55, 0.60, 0.65, and 0.70 produce 6,594, 7,549, 9,003, 9,837, and 10,393 reactions, respectively, with mean fold median nearest-training similarities of 0.325, 0.338, 0.353, 0.362, and 0.360. For comparison, the corresponding random-split control has a median similarity of 0.778. At a threshold of 0.55, varying $k$ between 50, 100, and 200 gives 7,884, 7,549, and 7,594 reactions with median similarities of 0.349, 0.338, and 0.337, respectively. Thus, these choices affect the exact composition of the split more than the magnitude of the train-test similarity shift.

We additionally test whether the reported model ranking depends on the incompleteness threshold used to select the most incomplete reactions. For this analysis, we retain the exact manuscript folds and apply stricter thresholds to the same held-out groups. The original top-20\% (p80) setting contains 7,457 reactions before vocabulary filtering and 6,158 evaluated reactions afterward (Appendix~\ref{app:vocab_filtering}). As shown in Table~\ref{tab:ood_sensitivity}, performance is nearly unchanged at p85 (top 15\%), whereas the stricter p90 (top 10\%) subset is substantially more difficult and could be used in future works to compare models. The model ranking remains unchanged across all three settings. We therefore treat the selected DRFP threshold, $k$, and incompleteness threshold as practical rather than uniquely optimal choices.

\begin{table}[htbp]
\centering
\caption{Sensitivity of model performance in extreme-OOD split, with stricter incompleteness thresholds. p85 and p90 are stricter subsets of the original p80 extreme-OOD test sets. Equivalence accuracy is reported as mean $\pm$ sample SD over five folds.}
\label{tab:ood_sensitivity}
\small
\begin{tabular}{lrrrr}
\toprule
Cutoff & Evaluated $n$ & RB (\%) & CRB (\%) & SynRBL (\%) \\
\midrule
p80 (top 20\%): $\geq5$ atoms or $\geq1$ C &
7,457 / 6,158 & 83.34 $\pm$ 5.05 & 91.71 $\pm$ 3.21 & 35.06 $\pm$ 17.10 \\
p85 (top 15\%): $\geq8$ atoms or $\geq2$ C &
5,554 / 4,604 & 83.68 $\pm$ 4.96 & 91.51 $\pm$ 3.15 & 35.24 $\pm$ 17.49 \\
p90 (top 10\%): $\geq13$ atoms or $\geq4$ C &
2,929 / 2,308 & 75.14 $\pm$ 3.91 & 85.31 $\pm$ 3.54 & 26.86 $\pm$ 9.26 \\
\bottomrule
\end{tabular}
\end{table}

\subsection{Constrained Reaction Balancer}\label{app:CRB_methods}

The constrained reaction balancer (CRB) modifies beam search by enforcing atom-balance constraints during decoding. At each decoding step, we apply a mask to the model logits to prevent tokens that would violate the current atom balance:
\[
p(y_t \mid y_{<t})
=
\operatorname{softmax}\left(z_t + m_t\right),
\]
where $z_t$ are the decoder logits and $m_t$ masks invalid tokens by assigning $-\infty$.

The mask is constructed dynamically based on the atom counts implied by the partial sequence $y_{<t}$. Decoding proceeds autoregressively: the model first generates the reactant side and then the product side. On the reactant side, the only constrained token is the arrow token. As long as atoms are missing on the reactant side (i.e., an atom type has a higher count on the product side of the input reaction), the arrow token is masked to prevent switching to the product side too early. All other tokens remain allowed, since additional reactant atoms can still be balanced later by generating corresponding product atoms.

After the arrow token is generated, the generated reactant side fixes the target atom counts for the product side. The constraint then becomes stricter: during product generation, tokens corresponding to already balanced atom types are masked, and the end-of-sequence token is only allowed once the reaction is atom- and charge-balanced. 
\[
m_t(v)=
\begin{cases}
-\infty & \text{if $v$ would violate atom balance},\\
-\infty & \text{if $v=\texttt{>}$ and reactants are incomplete},\\
-\infty & \text{if $v=\texttt{</s>}$ before balance is reached},\\
0 & \text{otherwise.}
\end{cases}
\]
Because each generated token changes the atom counts, the mask is updated at every decoding step for each beam.

For reactions containing out-of-vocabulary tokens, constrained decoding is disabled, since atom counts cannot be reliably computed. In the main benchmark we exclude these, but not in the analysis of the uncurated USPTO and the analysis in Appendix~\ref{app:vocab_filtering}. Overall, this enforces atom balance as a hard constraint, with the exception of hydrogen. This limitation arises because hydrogen atoms are mostly implicit in SMILES representations and depend on valence rules, making their exact counts difficult to track during token-level decoding. Despite this, the approach effectively enforces balance for all explicitly represented atom types while still allowing flexible generation of missing species.

\subsection{Evaluation metrics}\label{app:equiv_acc_metric}
We evaluate predicted reaction SMILES against ground-truth reactions using two complementary accuracy metrics.
\paragraph{Exact-match accuracy}
A prediction is counted as correct if the predicted and target reactions are identical after:
(i) parsing reactant and product sides,
(ii) canonicalizing each molecule SMILES with RDKit, and
(iii) comparing the resulting multisets of molecules on both sides.
This metric is invariant to molecule order within a reaction side, but preserves stoichiometry.

\paragraph{Equivalence accuracy}
Our primary metric is based on a more lenient equivalence match, which accounts for common representation ambiguities in chemical reaction datasets. Only valid predictions that are atom- and charge-balanced can count as correct. Starting from canonicalized reactions, we convert both the prediction and the target to one normal form using the chemistry-aware conventions in Table~\ref{tab:normalizations}. A prediction is counted as correct if both normal forms are identical. Transformations that change the chemistry of a reaction are deliberately not normalized. These are listed at the bottom of Table~\ref{tab:normalizations}.

\begin{table}[htbp]
\centering
\small
\caption{Conventions of the equivalence metric. Both the prediction and the target are converted to one normal form using these conventions.}
\label{tab:normalizations}
\begin{tabular}{p{0.28\linewidth} p{0.65\linewidth}}
\toprule
\textbf{Normalization} & \textbf{Intuitive description} \\
\midrule

Canonical
& Standardize molecule representations and ignore formatting differences. Molecules are canonicalized, so identical molecules are recognized even if written differently. Atom maps are ignored and isotopes are retained. \\

Acids and salts
& Treat common acids and salts consistently as ions. Hydrogen halides (HF, HCl, HBr, HI) are split into a proton and a halide. Common alkali-metal salts (halides, hydroxides, hydrides, methoxides, ethoxides, acetates and hydrogensulfates of Li, Na, K, Rb and Cs) are split into their ions. For example, NaCl and Na$^+$ + Cl$^-$ are considered equivalent. \\

Proton transfer
& Allow protons (H$^+$) to move between molecules on the same side of the reaction by normalizing the protonation state of N, O and S sites. This accounts for different choices of protonation state (acid--base form) that do not change the underlying chemistry. Bonds between heavy atoms are preserved. \\

Labile byproduct
& Chlorosulfurous acid (HO--S(=O)--Cl) and its conjugate base are written as SO$_2$, chloride and a proton. \\

Spectators
& Molecules that appear unchanged on both sides are cancelled by count before comparison. \\

\midrule
Not normalized
& Changes that are chemical reactions rather than alternative notations remain distinct: quenching of hydrides or carbanions (e.g.\ NaH + HCl vs.\ NaCl + H$_2$), carbonic-acid dehydration (H$_2$CO$_3$ vs.\ CO$_2$ + H$_2$O), reagent hydrolysis and activator consumption, radical recombination, and tautomer or constitutional-isomer conversions. \\

\bottomrule
\end{tabular}
\end{table}

\paragraph{Equivalence metric development and limitations: Manual audit}
To investigate limitations of the initial equivalence metric, one of the chemistry annotators (a graduate student, see Appendix~\ref{app:target_audit}) manually assessed a fixed-seed sample of 20 atom- and charge-balanced SynRBL predictions from the random split that this implementation classified as incorrect. The FlowER target and the SynRBL completion were shown with hidden identities and in randomized A/B order. The annotator judged 4/20 (20\%) chemically equivalent, 14/20 non-equivalent, and 2/20 uncertain. Among the 14 non-equivalent cases, the FlowER completion was preferred in eight, the SynRBL completion in three, and both were considered chemically plausible in three. As these are the judgements of a single annotator, several of them at low confidence (Table~\ref{tab:metric_audit_cases}), we report them as given and use them to identify representation differences rather than as a reference for chemical correctness.

This audit, together with further analysis of normalization behavior, informed revisions to the equivalence metric. All results reported in this paper use the revised implementation. It accepts all four completions judged chemically equivalent, while the remaining 16 cases remain metric-negative. Because the sample was selected using the initial metric and informed its revision, these findings are not an estimate of the revised metric's overall false-negative rate. The judgements also illustrate that distinct completions can both be chemically plausible. Table~\ref{tab:metric_audit_cases} lists all 20 annotator judgements, and Appendix~\ref{app:metric_audit_cases} shows one example per judgement. All audited examples, the annotations and both metric implementations are provided in the accompanying code repository.

\section{Additional results}
%Camera ready addition
\subsection{Out-of-vocabulary filtering}\label{app:vocab_filtering}
For the primary benchmark comparison, reactions containing out-of-vocabulary tokens in either the input or target are excluded as described in Section~5. In the full-test-set and uncurated-USPTO evaluations, no such filtering is applied. Out-of-vocabulary input tokens prevent reliable atom counting and therefore disable constrained decoding for CRB, while out-of-vocabulary target tokens cannot be generated by the RB or CRB due to the fixed Molecular Transformer vocabulary. Table~\ref{tab:vocabulary_filtering} lists the counts of excluded records from the main benchmark test splits, while Table~\ref{tab:main_results_with_OOV} lists the evaluation results of the models on the full testset. Compared to the main benchmark, as expected, the accuracy for CRB and RB drops notably, while for SynRBL it remains essentially unchanged in the random split, slightly increases in the group split and decreases in the OOD split. SynRBL's internal accuracy (Conf. >50\%) drops for all splits compared to the main benchmark. Even on the full test sets, CRB retains the highest equivalence accuracy, followed by RB and SynRBL.
\begin{table}[htbp]
\centering
\caption{Test-set coverage after out-of-vocabulary (OOV) filtering. In the main benchmark, we evaluate the models only on in-vocabulary reactions. A reaction is excluded if its input or target contains a token outside the pretrained Transformer vocabulary. Counts are pooled over five repetitions and therefore refer to test observations rather than unique reactions. The same reaction may appear in multiple repetitions.  Table~\ref{tab:main_results_with_OOV} reports results on the corresponding full test sets.}
\label{tab:vocabulary_filtering}
\begin{tabular}{lrrr}
\hline
Split & Full test sets & In-vocabulary (evaluated in Table~\ref{tab:main_results}) & Out-of-vocabulary (\%) \\
\hline
Random & 103,215 & 95,257 & 7,958 (7.71) \\
Group & 105,078 & 97,970 & 7,108 (6.76) \\
Extreme OOD & 7,457 & 6,158 & 1,299 (17.42) \\
\hline
\end{tabular}
\end{table}
\begin{table}[ht]
\centering
\small
\caption{Evaluation of all models across splits on the full test sets, including examples with out-of-vocabulary tokens.}

\label{tab:main_results_with_OOV}
\begin{tabular}{l l c c c c c}
\hline
Split & Model 
& \shortstack{Top-1 Acc\\(exact) (\%)} 
& \shortstack{Top-1 Acc\\(equiv.) (\%)} 
& \shortstack{Conf. $>$50\%\\(\%)} 
& \shortstack{Inv.\\SMILES (\%)} 
& \shortstack{Balanced\\(\%)} \\
\hline

R (incl. 
& RB & 90.13 $\pm$ 0.14 & 90.95 $\pm$ 0.13 & - & 2.08 $\pm$ 0.08 & 91.13 $\pm$ 0.13 \\
OOV)& CRB & 91.09 $\pm$ 0.11 & \textbf{93.77 $\pm$ 0.21} & - & 2.28 $\pm$ 0.22 & 94.18 $\pm$ 0.17 \\
& SynRBL & 5.45 $\pm$ 0.09 & 52.38 $\pm$ 0.15 & 72.26 $\pm$ 0.28 & - & 89.16 $\pm$ 0.24 \\

\hline

G (incl.
& RB & 86.83 $\pm$ 9.98 & 88.30 $\pm$ 9.07 & - & 1.42 $\pm$ 1.68 & 89.56 $\pm$ 9.02 \\
OOV)& CRB & 88.87 $\pm$ 9.54 & \textbf{92.69 $\pm$ 6.78} & - & 1.26 $\pm$ 0.98 & 95.00 $\pm$ 6.45 \\
& SynRBL & 2.68 $\pm$ 1.74 & 44.60 $\pm$ 27.62 & 78.88 $\pm$ 9.29 & - & 90.93 $\pm$ 4.58 \\

\hline

OOD (incl.
& RB & 70.46 $\pm$ 12.26 & 72.68 $\pm$ 11.96 & - & 7.64 $\pm$ 9.85 & 73.32 $\pm$ 11.98 \\
OOV)& CRB & 75.80 $\pm$ 11.77 & \textbf{82.20 $\pm$ 11.27} & - & 10.77 $\pm$ 10.13 & 84.01 $\pm$ 11.11 \\
& SynRBL & 5.57 $\pm$ 4.35 & 32.18 $\pm$ 16.34 & 55.65 $\pm$ 11.55 & - & 74.30 $\pm$ 5.54 \\

\hline

\end{tabular}
\end{table}

\subsection{Additional analysis on SynRBL performance}\label{app:SynRBL_analysis}
Table~\ref{tab:synRBL_results_extra} analyzes the impact of our preprocessing choice to move reagents to the reactant side. Since this can artificially increase the number of missing molecules/atoms, we compare two settings for SynRBL: without reagent identification (noRI), where reagents are merged into the reactant side (equivalent to the results in Table~\ref{tab:main_results}), and with reagent identification (RI), where reagents are kept in their original position between the reaction arrows (as provided in the USPTO dataset~\cite{lowe2012extraction} based on atom mapping).

When reagent identification is used (RI), equivalence accuracy increases by 2.3 to 2.9 percentage points and the fraction of balanced reactions also increases, indicating that SynRBL benefits from the additional structural information provided by explicitly separated reagents. However, the fraction of high-confidence predictions ($>50\%$) remains largely unchanged across settings, interestingly decreasing slightly for the random split and increasing for OOD and group split. This shows that preprocessing only has a minor effect on the algorithm. 

\begin{figure}[htbp]
    \centering
    \includegraphics[width=0.99\linewidth]{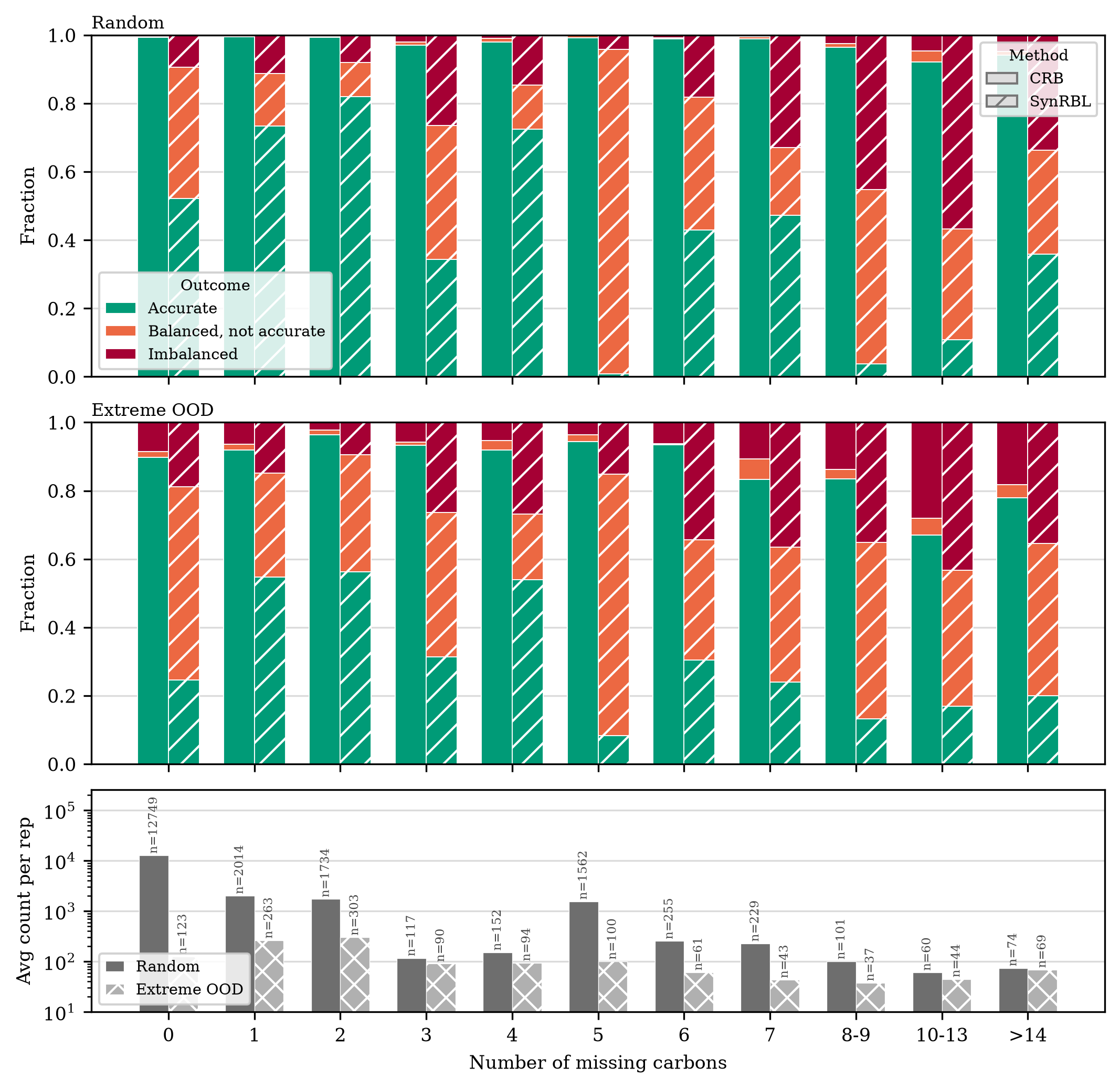}
    \caption{This figure reveals that the fraction of valid solutions according to SynRBL's internal accuracy (completed with confidence >50\%) would be higher (green + orange bar). We show this because there might be cases where our accuracy (equivalent) metric still does not catch certain valid solutions proposed by SynRBL.}
    \label{fig:placeholder}
\end{figure}

Figure~\ref{fig:placeholder} further analyzes error characteristics. SynRBL produces a substantial fraction of chemically balanced but non-matching predictions under our evaluation metric, reflecting the ambiguity of the completion task and the tendency of rule-based methods to propose alternative valid solutions. In contrast, CRB produces fewer such cases and a larger fraction of predictions that match the dataset targets. Even if all balanced but non-matching SynRBL outputs are treated as potentially valid, its effective performance remains significantly below that of CRB, confirming the advantage of the learned constrained decoding approach.
\begin{table}[htbp]
\centering
\small
\caption{Effect of including reagents (RI) in SynRBL input. While equivalence accuracy improves, the fraction of high-confidence predictions remains similar. This confirms that preprocessing does not explain the performance gap to ML models. All results are averaged over five folds (per split type) and $\pm$ in the result tables denotes standard deviation.}

\label{tab:synRBL_results_extra}
\begin{tabular}{l l c c c c}
\hline
Model & Split 
& \shortstack{Top-1 Acc\\(exact) (\%)} 
& \shortstack{Top-1 Acc\\(equiv.) (\%)} 
& \shortstack{Conf. $>$50\%\\(\%)} 
& \shortstack{Balanced\\(\%)} \\

\hline
SynRBL (noRI) & R   & 5.82 $\pm$ 0.10 & 52.39 $\pm$ 0.15 & 73.78 $\pm$ 0.25 & 89.99 $\pm$ 0.29 \\
SynRBL (noRI) & G   & 2.95 $\pm$ 1.99 & 44.29 $\pm$ 28.02 & 79.72 $\pm$ 8.63 & 91.34 $\pm$ 4.71 \\
SynRBL (noRI) & OOD & 6.66 $\pm$ 5.24 & 35.06 $\pm$ 17.10 & 59.14 $\pm$ 12.04 & 78.07 $\pm$ 4.82 \\
SynRBL (RI)   & R   & 1.15 $\pm$ 0.06 & 55.28 $\pm$ 0.22 & 71.38 $\pm$ 0.09 & 93.44 $\pm$ 0.11 \\
SynRBL (RI)   & G   & 0.70 $\pm$ 0.76 & 46.58 $\pm$ 28.95 & 81.59 $\pm$ 9.93 & 94.04 $\pm$ 2.91 \\
SynRBL (RI)   & OOD & 1.77 $\pm$ 1.87 & 37.37 $\pm$ 18.14 & 61.78 $\pm$ 19.19 & 79.47 $\pm$ 8.65 \\
\hline
\end{tabular}
\end{table}

%Camera ready addition:
\subsection{Missing-species stratification and completion results}
\label{app:missing_species}

To distinguish simple bookkeeping from more substantive completion cases, we stratify CompleteRXN by the type and location of missing species. Common helpers include species used in prior completion studies~\cite{zhang2024hybridcompletingreactions, van2024completing} such as water, HCl/HBr, ammonia, and common acids, bases, and solvents. Other missing species are separated into carbon-free (protons, counterions, other inorganic species) and carbon-bearing compounds. Lastly, we also record reagent restoration, where an identified reagent is restored on the product side, potentially in a modified protonation state. Table~\ref{tab:missing_species} summarizes the resulting benchmark composition and random-split top-1 equivalence accuracy.

\begin{table}[htbp]
\centering
\caption{Random-split top-1 equivalence accuracy by missing-species type.
Results are mean $\pm$ standard deviation over five repetitions.}
\label{tab:missing_species}
\small
\begin{tabular}{p{4.0cm}rrrr}
\toprule
Missing-species stratum & $n$ (\%) & RB (\%) & CRB (\%) & SynRBL (\%) \\
\midrule
Common helper only
& 7,676 (3.72) & 96.61 $\pm$ 0.71 & 98.87 $\pm$ 0.42 & 69.25 $\pm$ 0.63 \\
Carbon-free only
& 46,219 (22.39) & 98.41 $\pm$ 0.27 & 99.53 $\pm$ 0.07 & 75.50 $\pm$ 0.33 \\
Carbon-bearing only
& 75,521 (36.59) & 97.88 $\pm$ 0.20 & 99.14 $\pm$ 0.16 & 33.22 $\pm$ 0.16 \\
Helper + carbon-free
& 14,847 (7.19) & 98.60 $\pm$ 0.30 & 99.53 $\pm$ 0.09 & 8.51 $\pm$ 1.32 \\
Helper + carbon-bearing
& 8,136 (3.94) & 96.51 $\pm$ 0.57 & 98.91 $\pm$ 0.27 & 15.28 $\pm$ 1.77 \\
Carbon-free + carbon-bearing
& 29,732 (14.40) & 97.95 $\pm$ 0.29 & 99.15 $\pm$ 0.05 & 76.79 $\pm$ 1.37 \\
All three types
& 3,018 (1.46) & 95.36 $\pm$ 0.75 & 97.50 $\pm$ 1.00 & 5.06 $\pm$ 1.18 \\
\midrule
Any non-helper carbon-bearing omission
& 116,407 (56.39) & 97.76 $\pm$ 0.10 & 99.10 $\pm$ 0.08 & 42.41 $\pm$ 0.18 \\
Reagent restoration only
& 21,274 (10.31) & 98.57 $\pm$ 0.19 & 99.78 $\pm$ 0.03 & 89.43 $\pm$ 1.05 \\
\bottomrule
\end{tabular}
\end{table}

\paragraph{Complexity of missing molecules.}
More than half of the benchmark (56.39\%) contains at least one non-helper carbon-bearing omission. Within this subset, reactions have a median of two missing molecules, 31 missing atoms including implicit hydrogen, and eight missing carbon atoms. RB, CRB, and SynRBL achieve 97.76\%, 99.10\%, and 42.41\% equivalence accuracy, respectively.

\paragraph{Location of missing species.}
Missing molecules occur only on the product side in 69.07\% of reactions, on both sides in 20.57\%, and only on the reactant side in 0.06\%; 10.31\% require reagent restoration only. Overall, 56.17\% contain a product-side non-helper carbon-bearing omission, consistent with patent records emphasizing the main product while omitting additional products or byproducts.

\paragraph{SynRBL accuracy across missing species type.}
SynRBL's accuracy varies strongly between missing species types, which largely reflects how it completes two common reaction types. In the helper + carbon-free stratum, the majority of SynRBL errors (61\%) list an alkali-metal hydride base such as NaH unchanged next to the hydrogen halide released by the reaction. These two species cannot coexist, as they react to the metal halide and H$_2$, which is the completion given by the reference. The equivalence metric therefore does not treat these completions as equivalent. In the carbon-free + carbon-bearing stratum, 65\% of the reactions use a carbonate base. Here SynRBL lists the carbonate next to the hydrogen halide, whereas the reference gives bicarbonate and halide. Unlike hydride quenching, which forms H$_2$, these completions differ only by a proton transfer between molecules on the same side and are counted as equivalent.

Although carbon content alone does not establish mechanistic importance, these results show that CompleteRXN is not dominated by routine helper restoration and that CRB remains highly accurate on multi-molecule, carbon-bearing completion cases.

%camera ready addition
\subsection{Benchmark cross-method agreement analysis}
\label{app:agreement}

Agreement between CRB and SynRBL is evaluated after removing stereochemical annotations from the pairwise predictions, because SynRBL standardization removes stereochemistry whereas CRB retains it. This normalization is used only for the CRB-SynRBL agreement comparison, while individual benchmark accuracies remain evaluated against their respective references as described in Section~\ref{sec:benchmark}. Agreement is then defined when the two predictions match under our equivalence metric, independently of the benchmark reference. Table~\ref{tab:agreement} reports agreement coverage and the fraction of these agreements that also match the benchmark reference. 
Across the three evaluations, only 125 agreement observations (120 unique reactions) do not match the benchmark reference. Representative cases show recurring differences involving protonation and ionic speciation, as well as alternative small-molecule and reagent-derived byproduct assignments. These cases highlight remaining ambiguity in single-reference evaluation and in the equivalence metric, although genuinely distinct shared predictions also occur. Thus, cross-method agreement provides a high-precision reliability signal but does not guarantee chemical correctness. The complete set of benchmark-negative agreement cases and the code used to generate this analysis are provided in the accompanying repository.

\begin{table}[htbp]
\centering
\caption{Cross-method agreement between CRB and SynRBL. Coverage is the
fraction of test predictions on which both methods agree. Precision is the
fraction of these agreements that also match the benchmark reference under
the equivalence metric. Counts are aggregated over five repetitions per split type.}
\label{tab:agreement}
\small
\begin{tabular}{lrrrr}
\toprule
Split & Agreement $n$ & Coverage (\%) &
Reference-negative $n$ & Precision (\%) \\
\midrule
Random      & 49,782 & 52.26 & 32 & 99.94 \\
Group       & 45,275 & 46.21 & 82 & 99.82 \\
Extreme OOD & 2,330  & 37.84 & 11 & 99.53 \\
\bottomrule
\end{tabular}
\end{table}

\subsection{Impact of constrained decoding under increasing difficulty.}\label{app:CRBvsRB}
\begin{figure}[htbp]
    \centering
    \includegraphics[width=0.95\linewidth]{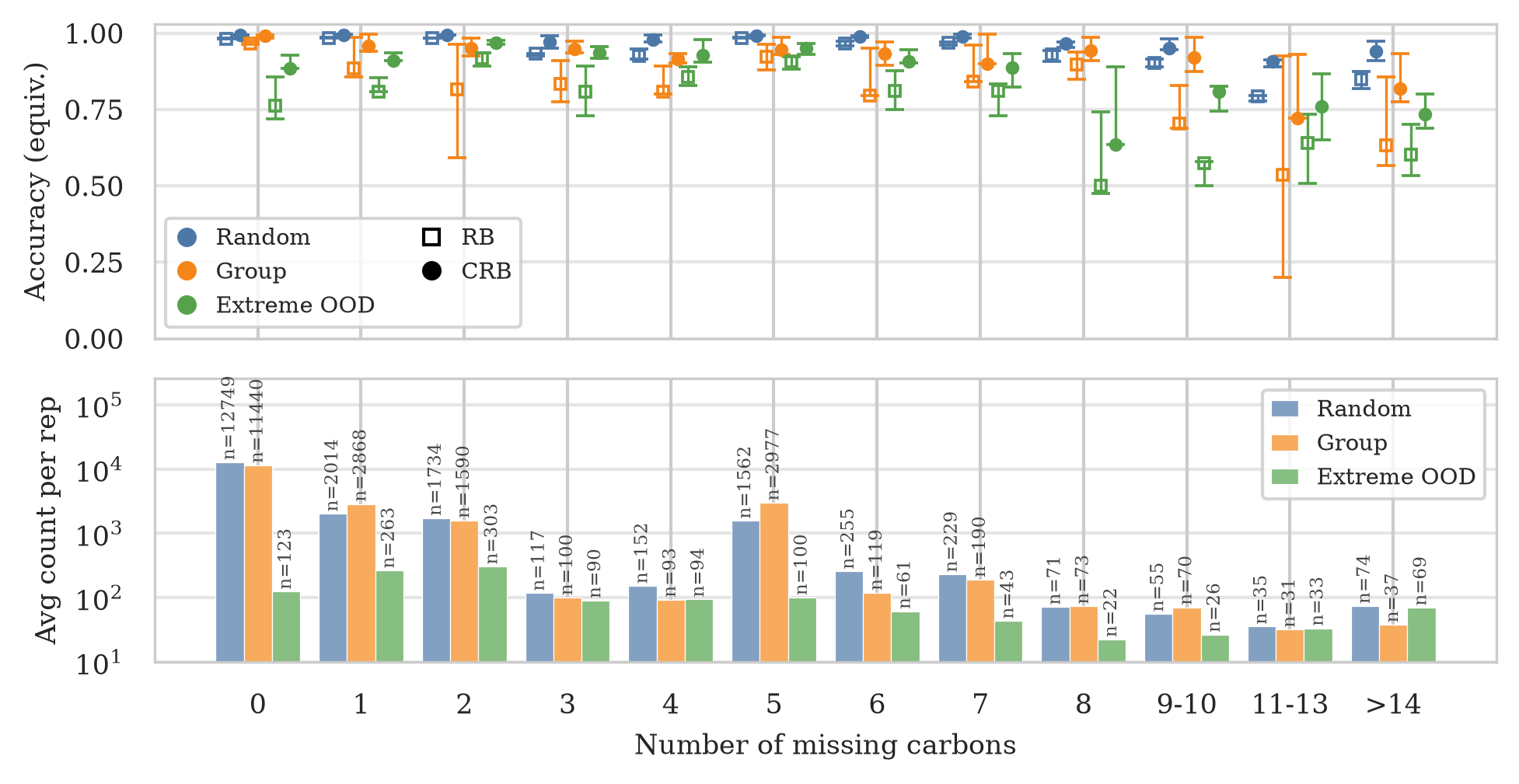}
    \caption{Equivalence accuracy of the Reaction Balancer without (RB) and with constrained decoding (CRB) as a function of reaction difficulty (number of missing carbons). CRB consistently outperforms RB across all regimes, with larger gains for more incomplete reactions.}
    \label{fig:appendix_CRB_vs_RB}
\end{figure}

Figure~\ref{fig:appendix_CRB_vs_RB} provides a more detailed view of the performance differences between the two encoder-decoder model variants as a function of task difficulty. Across all levels of missing carbons, constrained decoding (CRB) consistently outperforms the unconstrained baseline (RB). The gap becomes more pronounced for more difficult reactions, indicating that constrained beam search improves robustness by guiding decoding toward atom-balanced solutions, especially in highly unbalanced regimes.

We also test whether the improvement from RB to CRB is statistically significant using paired per-reaction tests with Holm correction. CRB improves equivalence accuracy by 1.28, 3.01, and 8.37 percentage points on the random, group-based, and extreme-OOD splits, respectively, with all three differences significant ($p_{\mathrm{adj}}<10^{-99}$). In cases where the two models differ in correctness, CRB is correct while RB is not for 1,222, 2,689, and 550 test reactions in the three settings, respectively. We observe no cases with the opposite outcome. The improvement from CRB also remains significant across all 36 missing-carbon bins shown in Figure~\ref{fig:appendix_CRB_vs_RB} ($p<0.005$).

\subsection{Additional details on performance on uncurated USPTO}
\label{app:uncurated_results}

To examine model behavior beyond the curated benchmark setting, we analyze predictions on the large-scale uncurated subset of the USPTO dataset. In this setting, no ground-truth completions are available, and evaluation therefore relies on proxies such as atom balance, model confidence, and cross-method agreement.

We first examine cross-method agreement and model-specific confidence estimates. Among the 384,580 reactions for which both methods produce balanced outputs, CRB and SynRBL agree on 268,839 (69.90\%) under our equivalence metric. Specifically, 30.40\% of SynRBL predictions exceed a confidence threshold of 0.50 (and 27.03\% exceed 0.90), while 17.49\% of CRB predictions exceed a prediction probability of 0.95. With the stricter confidence thresholds (SynRBL $\geq 0.90$, CRB $\geq 0.95$), both models are confident on 28,861 records, of which 18,987 (65.79\%) also agree. These agreeing, high-confidence records correspond to 4.94\% of all balanced CRB-SynRBL overlaps. Combining confidence and agreement therefore substantially reduces coverage. Since no validated reference is available for the uncurated USPTO data, these quantities should not be interpreted as accuracy or precision estimates.

\paragraph{CRB failure mode analysis}
Overall, 69.40\% of CRB outputs are parseable and 49.31\% are atom- and charge-balanced. Of the 177,232 parseable but imbalanced predictions, 170,093 (95.97\%) differ only in hydrogen balance, consistent with hydrogen not being explicitly constrained during decoding. Among invalid outputs, 264,174/270,001 (97.84\%) contain an unparseable predicted molecule. In addition, 90,153 predictions reach the 200-token decoding limit, of which 86,672 (96.14\%) are invalid, thus, truncation accounts for about one third of invalid outputs but is not their sole cause.

\begin{table}[htbp]
\centering
\caption{CRB outcomes on the 882,357 uncurated USPTO reactions. These reactions have no validated completion targets. Categories therefore describe syntactic validity and formal atom balance, not completion accuracy.}
\label{tab:uncurated_outcomes}
\small
\begin{tabular}{lrr}
\toprule
Outcome & $n$ & Fraction (\%) \\
\midrule
Balanced and parseable & 435,124 & 49.31 \\
Parseable; hydrogen-only imbalance & 170,093 & 19.28 \\
Parseable; other imbalance & 7,139 & 0.81 \\
Invalid predicted molecule & 264,174 & 29.94 \\
Malformed predicted reaction format & 5,477 & 0.62 \\
Unparseable molecule in USPTO input & 350 & 0.04 \\
\bottomrule
\end{tabular}
\end{table}

Furthermore, failures are not distributed uniformly across the uncurated dataset as shown in Table~\ref{tab:uncurated_failure_subsets}. They are substantially more frequent for records with larger or carbon-bearing deficits, transition-metal species, many components, and multiple products. These characteristics are correlated and indicate associations but not independent causes.

\begin{table}[htbp]
\centering
\caption{CRB outcomes for selected characteristics of the uncurated USPTO records. Characteristics are computed from the original USPTO records before identified reagents are moved to the reactant side. Subsets overlap and should not be interpreted as independent causal factors.}
\label{tab:uncurated_failure_subsets}
\small
\begin{tabular}{lrrr}
\toprule
Selected input subset & $n$ & Balanced (\%) & Invalid (\%) \\
\midrule
Zero missing atoms & 46,870 & 78.24 & 13.99 \\
More than 20 missing atoms & 116,877 & 21.33 & 61.19 \\
Carbon-free deficit & 464,099 & 60.54 & 19.74 \\
Carbon-bearing deficit & 418,258 & 36.86 & 42.65 \\
Contains transition metal & 132,290 & 12.49 & 56.90 \\
More than 10 left-side components & 35,749 & 15.25 & 64.67 \\
Multi-product record & 79,472 & 38.69 & 45.51 \\
\bottomrule
\end{tabular}
\end{table}

The statistics in Table~\ref{tab:uncurated_failure_subsets} are computed from the original USPTO records. Because identified reagents are subsequently moved to the reactant side before CRB inference, records that are initially balanced can become incomplete model inputs. Performance decreases as the number of moved reagent molecules increases (Table~\ref{tab:reagent_restoration}).

\begin{table}[htbp]
\centering
\caption{CRB outcomes for initially balanced reactions as a function of the number of identified reagent molecules moved to the reactant side before inference.}
\label{tab:reagent_restoration}
\small
\begin{tabular}{lrrr}
\toprule
Identified reagent molecules & $n$ & Balanced (\%) & Invalid (\%) \\
\midrule
0 & 4,855 & 98.93 & 0.29 \\
1--2 & 16,238 & 93.41 & 2.43 \\
3--5 & 15,499 & 77.12 & 12.08 \\
$>5$ & 9,859 & 48.13 & 39.68 \\
\bottomrule
\end{tabular}
\end{table}

Lastly, the uncurated records also differ systematically from the CompleteRXN benchmark data. Their median and 95th-percentile missing-atom counts are 5 and 34, compared with 3 and 16 in CompleteRXN, and 47.4\% contain a carbon-bearing deficit compared with 32.0\% in the benchmark. In a fixed random sample of 10,000 uncurated reactions, the median nearest-neighbor similarity to CompleteRXN is 0.300, while the 90th percentile reaches 0.718 (compared with 0.778 and 1.0 within CompleteRXN). Thus, the uncurated set partly overlaps with CompleteRXN but also contains a substantial less-similar tail. Consistent with this, CRB produces balanced outputs for 68.50\% of sampled reactions with nearest-CompleteRXN similarity $\geq 0.55$, compared with 39.05\% for reactions below 0.20.

\subsection{Additional evaluation on SynRXN Complex}
\label{app:synrxn}

We additionally evaluate CRB on SynRXN~\cite{phan2026synrxn}, which provides an independently developed benchmark for reaction balancing. The main distinction between CompleteRXN and SynRXN is how the balanced reference reactions are constructed. In CompleteRXN, incomplete USPTO records are matched to FlowER reactions derived from expert-curated mechanistic templates. In SynRXN, balanced reference reactions are generated with SynRBL, filtered by model confidence ($>90\%$), and reportedly checked manually afterwards. CompleteRXN is also designed for both training and evaluation: its larger size and random, mechanism-grouped, and extreme-OOD splits allow us to study model training and generalization, whereas SynRXN focuses primarily on evaluation.

SynRXN combines two datasets: one derived from the USPTO and one USPTO independent dataset (\textit{SynRXN Complex}). To avoid potential overlap with the USPTO-derived SynRXN set, we evaluate CRB on \textit{SynRXN Complex}. Of the 1,746 records, 36 reference reactions are not atom- and charge-balanced and 348 inputs are already balanced. Restricting the evaluation to the 1,363 records with non-empty, balanced references, CRB produces balanced outputs for 857/1,363 (62.88\%) and outputs that are both balanced and reference-equivalent for 747/1,363 (54.81\%). Most balance failures arise from invalid generated SMILES or residual hydrogen imbalance, consistent with the failure modes observed on the uncurated USPTO data.

A mismatch with the SynRXN reference does not necessarily imply that the CRB completion is chemically incorrect. We manually inspected 12 disagreements, sampled under the initial implementation of the equivalence metric. In eight cases, both completions appeared chemically plausible, in two cases, we considered the CRB completion more plausible, and in two cases, we considered the SynRXN completion more plausible. Under the revised metric, five of the eight plausible alternatives are counted as equivalent to the reference, while the remaining seven cases still differ. For example, in one intramolecular cyclization, CRB produces methane, whereas the SynRXN reference adds an oxygen species and produces methanol. Both equations are balanced, but methanol formation from methoxy-group cleavage is chemically more plausible, and we therefore preferred the SynRXN completion. Conversely, in one HATU-mediated amidation, the SynRXN reference represents the acid and amine forming the amide and water while carrying HATU through unchanged. CRB instead consumes HATU and produces tetramethylurea and HOAt-derived species, which better reflects the expected stoichiometric role of the coupling reagent.

Overall, the SynRXN evaluation reveals mostly known CRB limitations in out-of-distribution test sets and hydrogen handling. It also illustrates that detailed reagent and byproduct accounting is not always unique: model-assisted references may encode one chemically acceptable level of accounting, while CRB, trained on a different type of reference data, may produce another. CompleteRXN therefore complements SynRXN through substantially larger training and evaluation sets, controlled generalization splits, and targets constructed independently of the balancing methods evaluated here. The SynRXN evaluation outputs, manually inspected disagreements, and analysis code are provided in the accompanying repository.

\section{Manual curation details}
\label{app:expert_curation}

While the CompleteRXN benchmark enables controlled evaluation under structured and realistic missing-data patterns, it does not fully capture the diversity and noise of the full USPTO as shown in the last results paragraph of Section~\ref{sec:experiments}. In particular, rare reaction types, large structural gaps, and incorrectly reported reactions remain underrepresented.

To address this, we propose a complementary manually curated dataset constructed via targeted selection of challenging reactions. The selection strategy is guided by three principles:

\begin{itemize}
    \item \textbf{Diversity:} coverage across reaction classes and templates
    \item \textbf{Complexity:} preference for highly unbalanced reactions 
    \item \textbf{Coverage:} inclusion of underrepresented and out-of-distribution reactions
\end{itemize}

Reactions are curated by chemistry annotators using a tool-supported workflow with multiple independent annotations per reaction to ensure quality and consistency. Importantly, this setting goes beyond pure completion: annotators are allowed to modify incorrect or duplicated/alternative molecules, enabling the joint task of reaction completion and correction.

In the future, this dataset is intended as final part completing this benchmark to evaluate completion methods under realistic conditions, where both missing information and incorrect entries must be resolved. A detailed description of the selection and prioritization procedure will be provided alongside the release of the curated dataset.

\section{Manual audits}
% Generated by analysis/expert_audits/build_appendix.py; do not edit by hand.
\providecommand{\expertauditbarrier}{\clearpage}
\makeatletter\@ifundefined{expertauditpx}{\newlength{\expertauditpx}}{}\makeatother
\setlength{\expertauditpx}{\dimexpr(\textwidth-2.4cm-2mm)/2654\relax}

\subsection{Reactions from the audit of benchmark targets}
\label{app:target_audit_cases}

\begin{figure}[htbp]
\centering\small
\begin{tabular}{@{}p{2.4cm}@{\hspace{2mm}}l@{}}
USPTO input & \raisebox{-.5\height}{\includegraphics[width=911\expertauditpx]{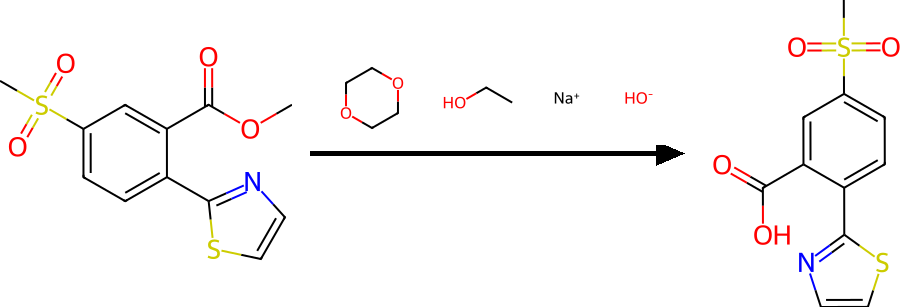}} \\[2mm]
FlowER target & \raisebox{-.5\height}{\includegraphics[width=1182\expertauditpx]{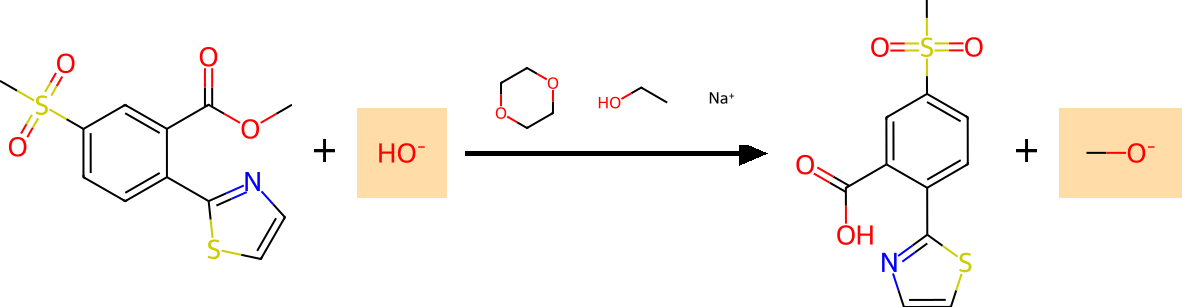}} \\[2mm]
Annotator 4\newline(match) & \raisebox{-.5\height}{\includegraphics[width=1420\expertauditpx]{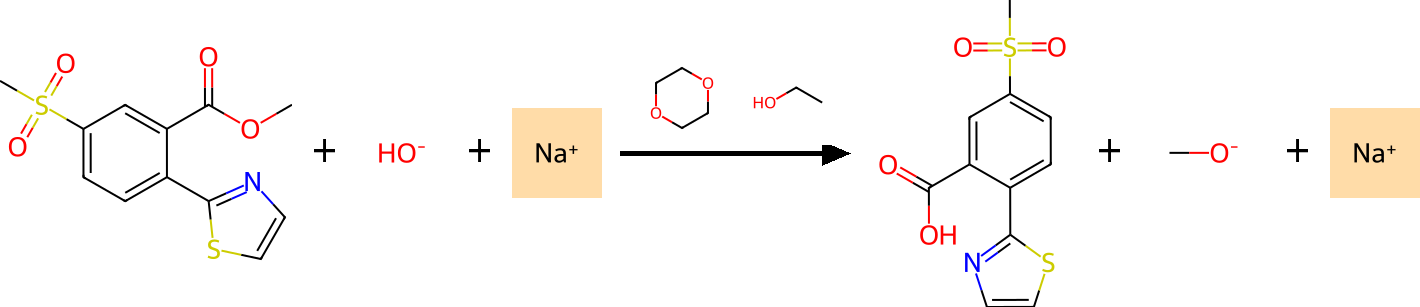}} \\[2mm]
Annotator 5\newline(match) & \raisebox{-.5\height}{\includegraphics[width=1254\expertauditpx]{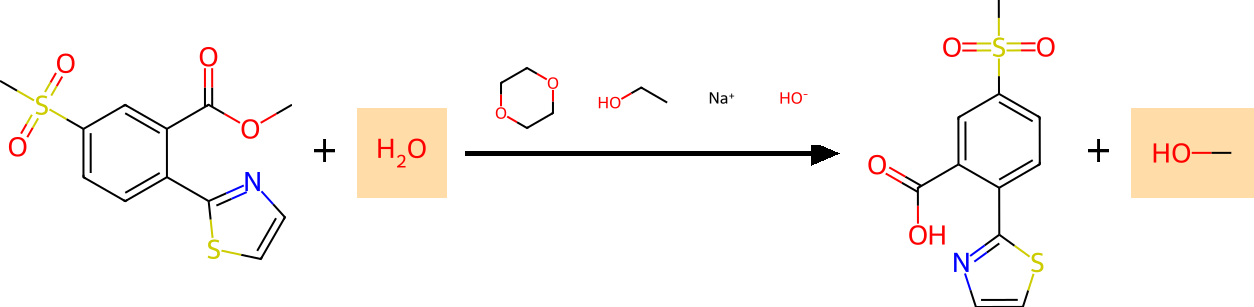}} \\[2mm]
Annotator 6\newline(match) & \raisebox{-.5\height}{\includegraphics[width=1420\expertauditpx]{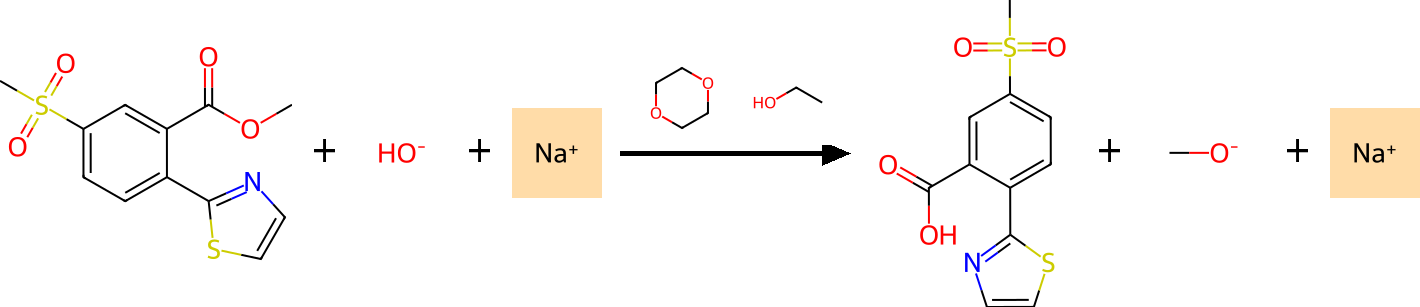}} \\
\end{tabular}
\caption{Target audit, control 17 (\texttt{USPTO\_US07462617B2\_476457}). Three of the three independent chemistry-annotator completions match the benchmark target under the equivalence metric. Shaded molecules in the target are missing from the same side of at least one compared completion. Shaded molecules in a completion do not occur on the same side of the target. Agents are drawn above the arrow and are not compared. Selection: the control with the lowest ID among the 43 controls with this level of agreement.}
\label{fig:target_audit_control_17}
\end{figure}

\begin{figure}[htbp]
\centering\small
\begin{tabular}{@{}p{2.4cm}@{\hspace{2mm}}l@{}}
USPTO input & \raisebox{-.5\height}{\includegraphics[width=1249\expertauditpx]{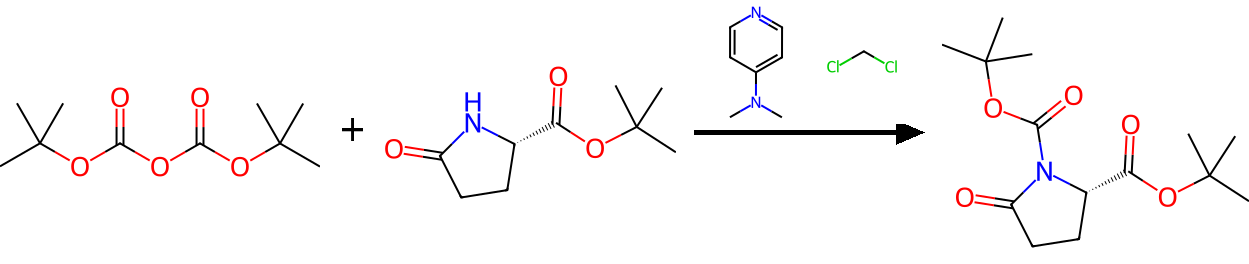}} \\[2mm]
FlowER target & \raisebox{-.5\height}{\includegraphics[width=1682\expertauditpx]{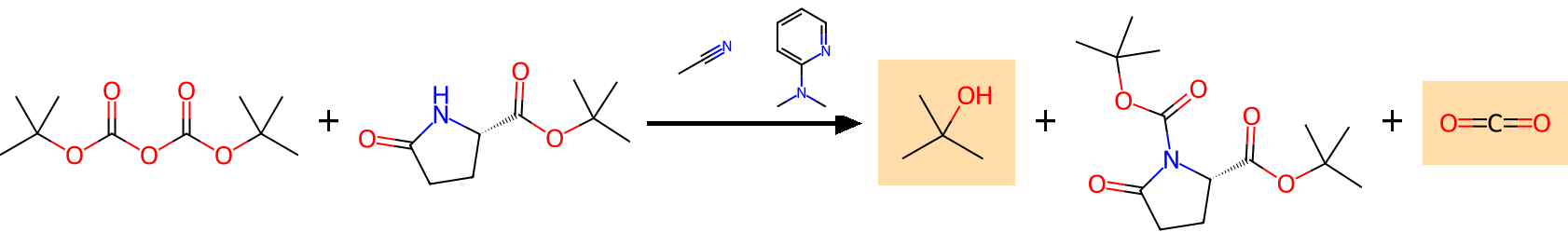}} \\[2mm]
Annotator 4\newline(match) & \raisebox{-.5\height}{\includegraphics[width=1682\expertauditpx]{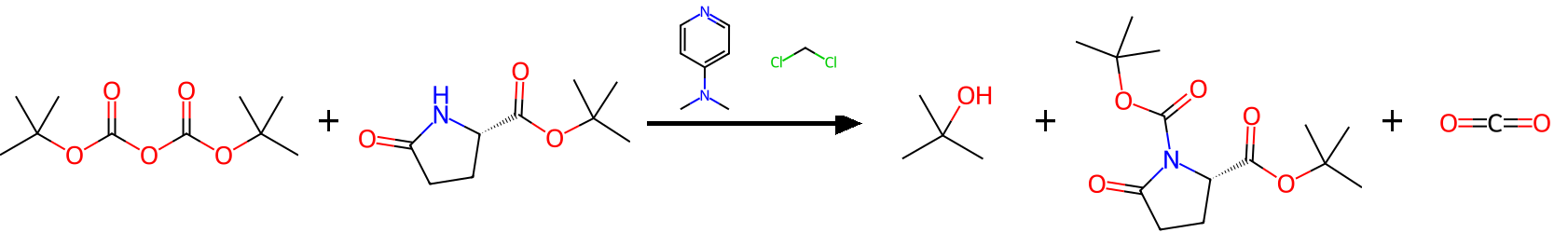}} \\[2mm]
Annotator 5\newline(no match) & \raisebox{-.5\height}{\includegraphics[width=1519\expertauditpx]{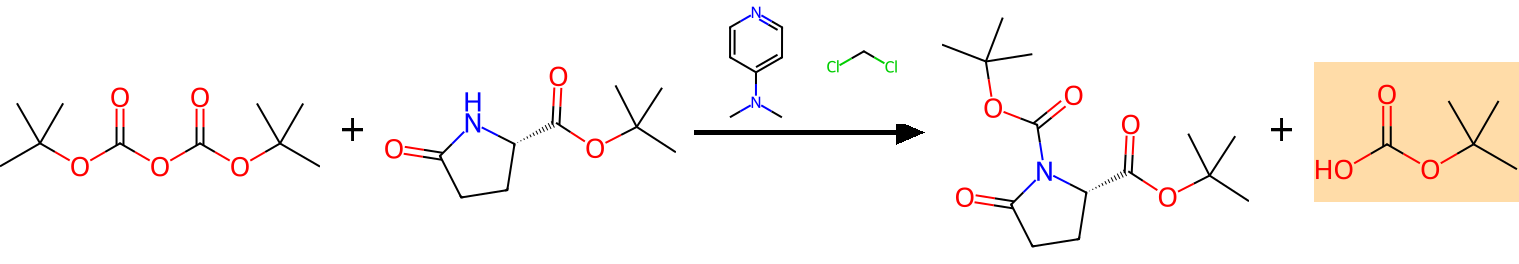}} \\[2mm]
Annotator 6\newline(match) & \raisebox{-.5\height}{\includegraphics[width=1652\expertauditpx]{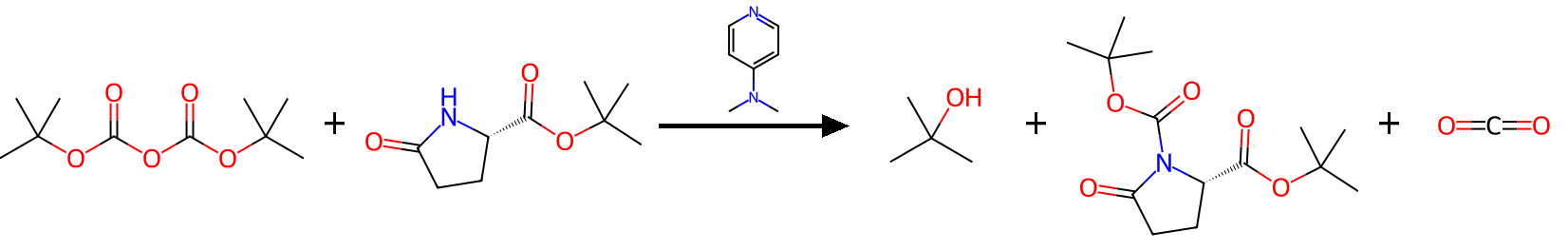}} \\
\end{tabular}
\caption{Target audit, control 2 (\texttt{USPTO\_US07094800B2\_395559}). Two of the three independent chemistry-annotator completions match the benchmark target under the equivalence metric. Shaded molecules in the target are missing from the same side of at least one compared completion. Shaded molecules in a completion do not occur on the same side of the target. Agents are drawn above the arrow and are not compared. Selection: the control with the lowest ID among the 5 controls with this level of agreement.}
\label{fig:target_audit_control_2}
\end{figure}

\begin{figure}[htbp]
\centering\small
\begin{tabular}{@{}p{2.4cm}@{\hspace{2mm}}l@{}}
USPTO input & \raisebox{-.5\height}{\includegraphics[width=964\expertauditpx]{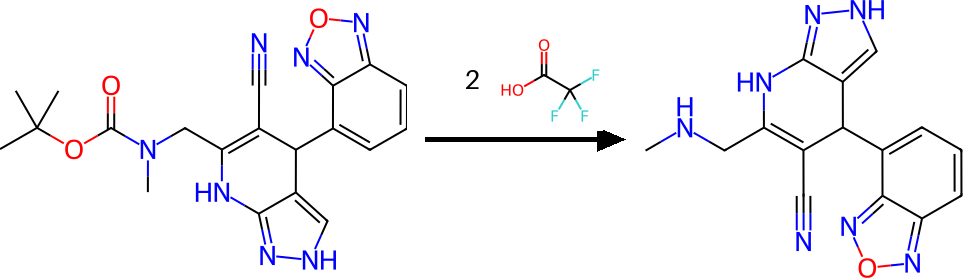}} \\[2mm]
FlowER target & \raisebox{-.5\height}{\includegraphics[width=1397\expertauditpx]{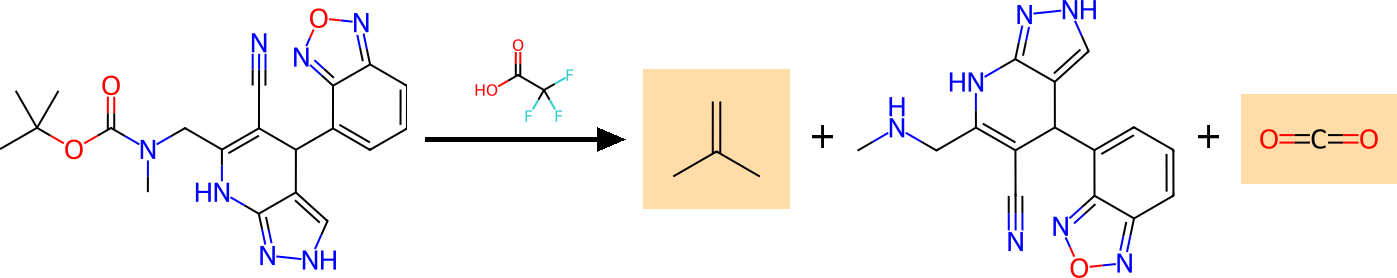}} \\[2mm]
Annotator 2\newline(no match) & \raisebox{-.5\height}{\includegraphics[width=1879\expertauditpx]{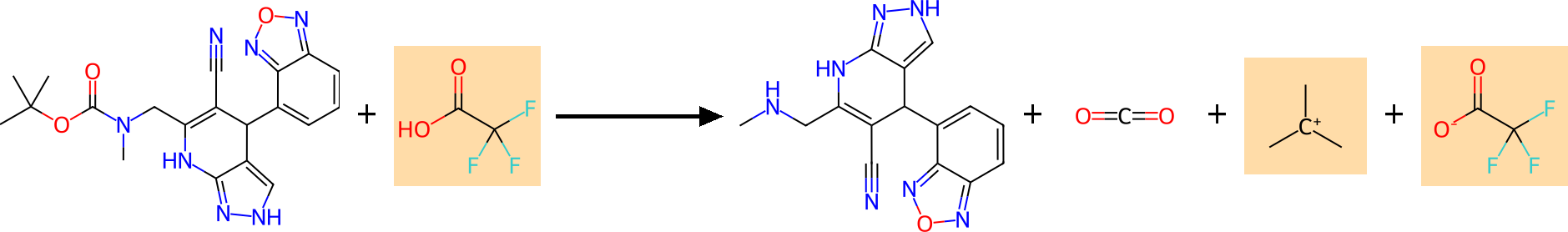}} \\[2mm]
Annotator 3\newline(no match) & \raisebox{-.5\height}{\includegraphics[width=1439\expertauditpx]{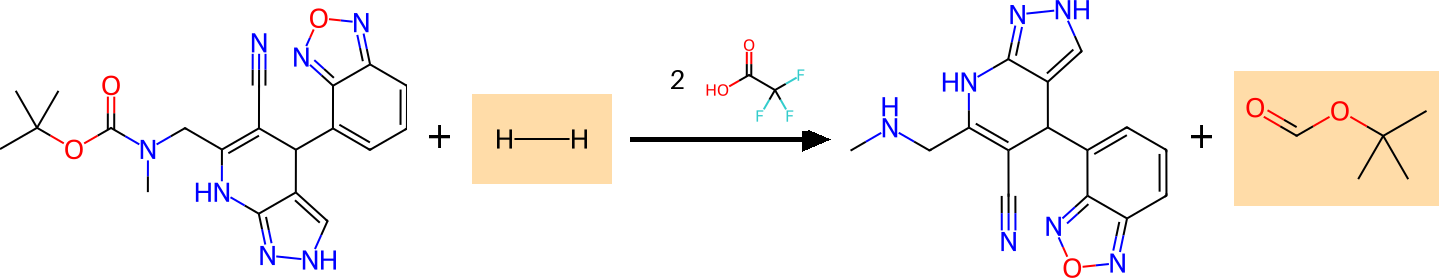}} \\[2mm]
Annotator 4\newline(match) & \raisebox{-.5\height}{\includegraphics[width=1397\expertauditpx]{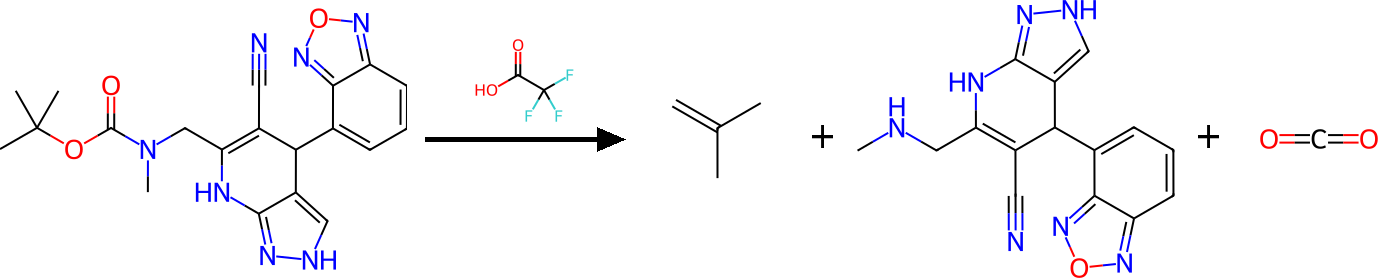}} \\
\end{tabular}
\caption{Target audit, control 177 (\texttt{USPTO\_US06977262B2\_1047922}). One of the three independent chemistry-annotator completions matches the benchmark target under the equivalence metric. Shaded molecules in the target are missing from the same side of at least one compared completion. Shaded molecules in a completion do not occur on the same side of the target. Agents are drawn above the arrow and are not compared. Selection: the control with the lowest ID among the 2 controls with this level of agreement.}
\label{fig:target_audit_control_177}
\end{figure}

\begin{figure}[htbp]
\centering\small
\begin{tabular}{@{}p{2.4cm}@{\hspace{2mm}}l@{}}
USPTO input & \raisebox{-.5\height}{\includegraphics[width=1801\expertauditpx]{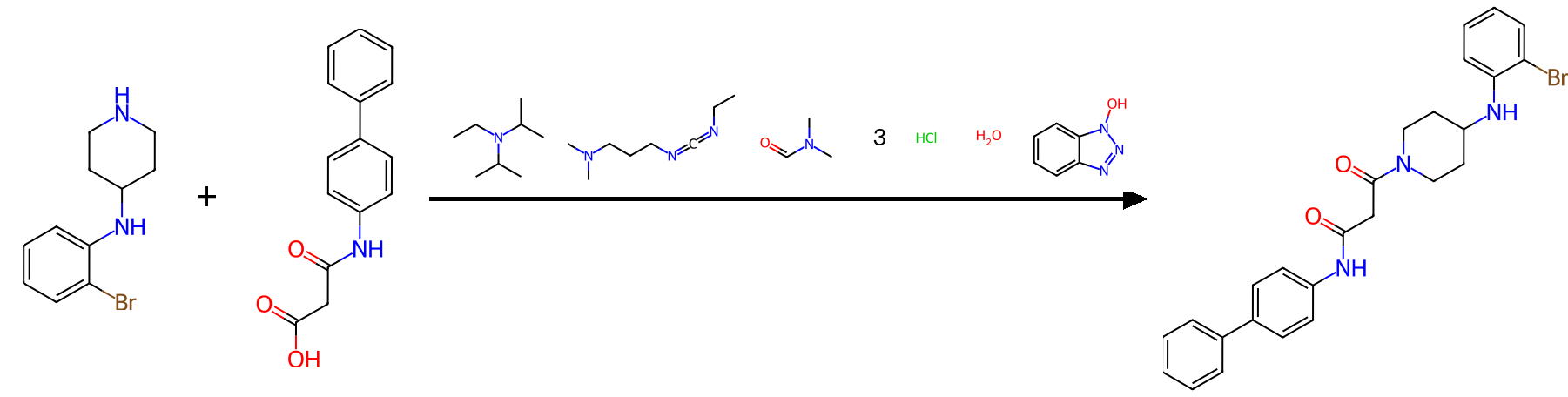}} \\[2mm]
FlowER target & \raisebox{-.5\height}{\includegraphics[width=2390\expertauditpx]{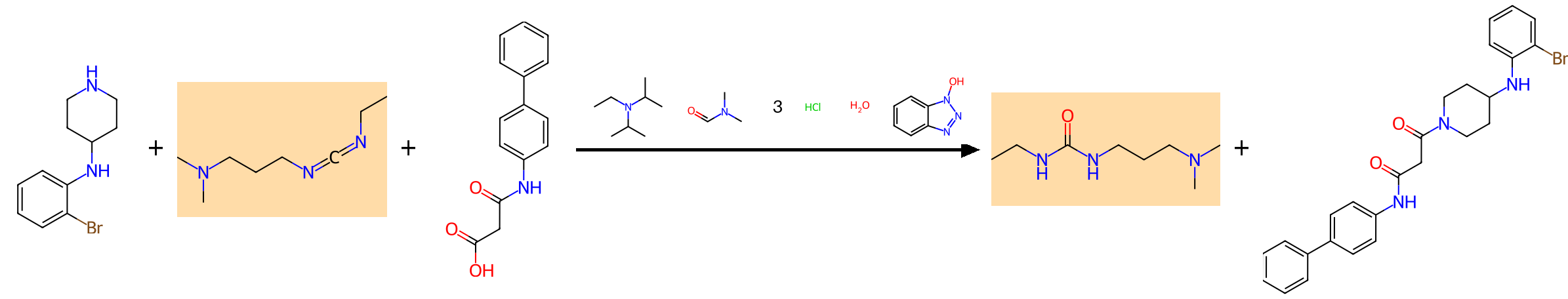}} \\[2mm]
Annotator 1\newline(no match) & \raisebox{-.5\height}{\includegraphics[width=1884\expertauditpx]{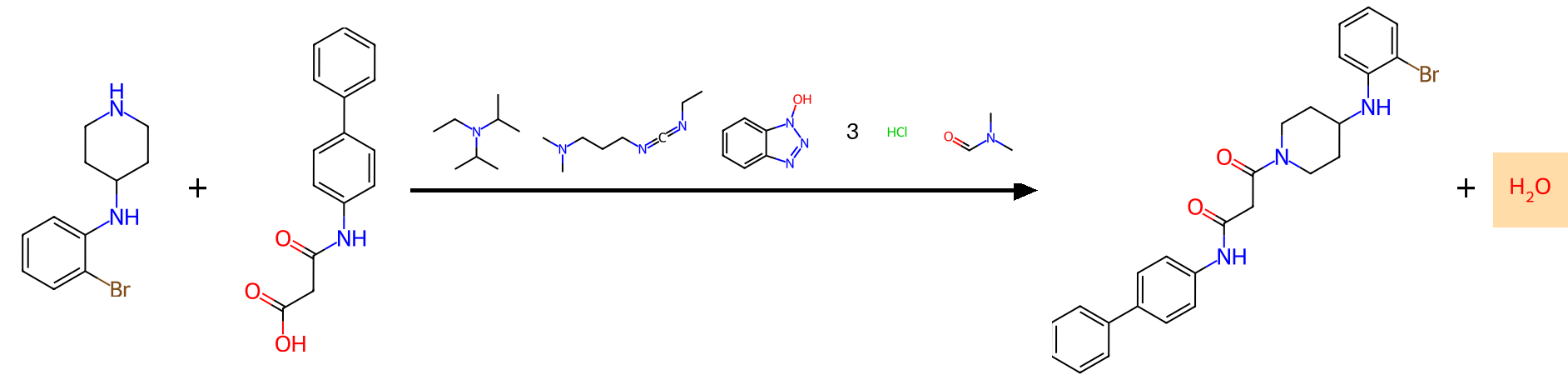}} \\[2mm]
Annotator 2\newline(no match) & \raisebox{-.5\height}{\includegraphics[width=1956\expertauditpx]{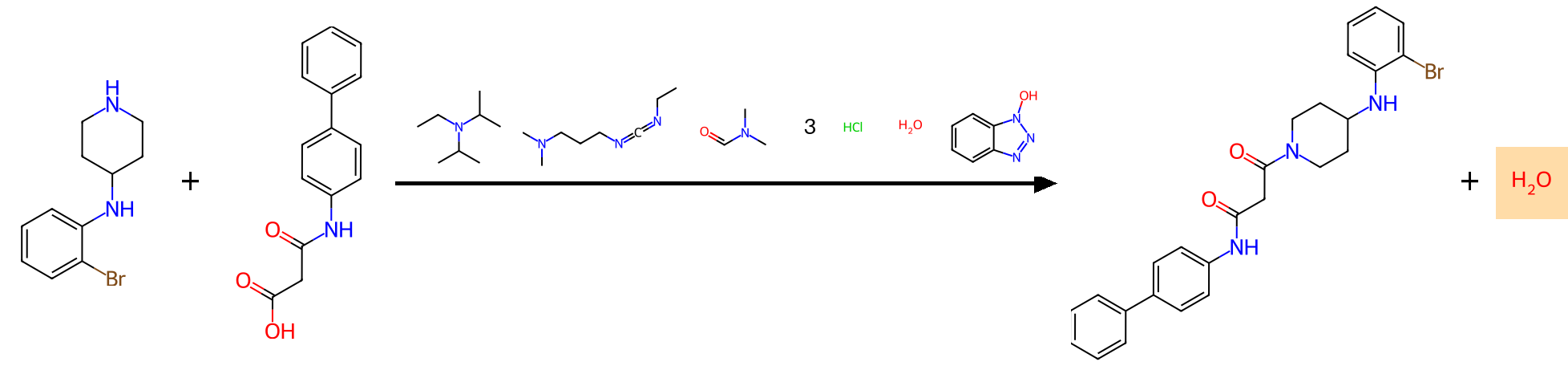}} \\[2mm]
Annotator 3\newline(no match) & \raisebox{-.5\height}{\includegraphics[width=1956\expertauditpx]{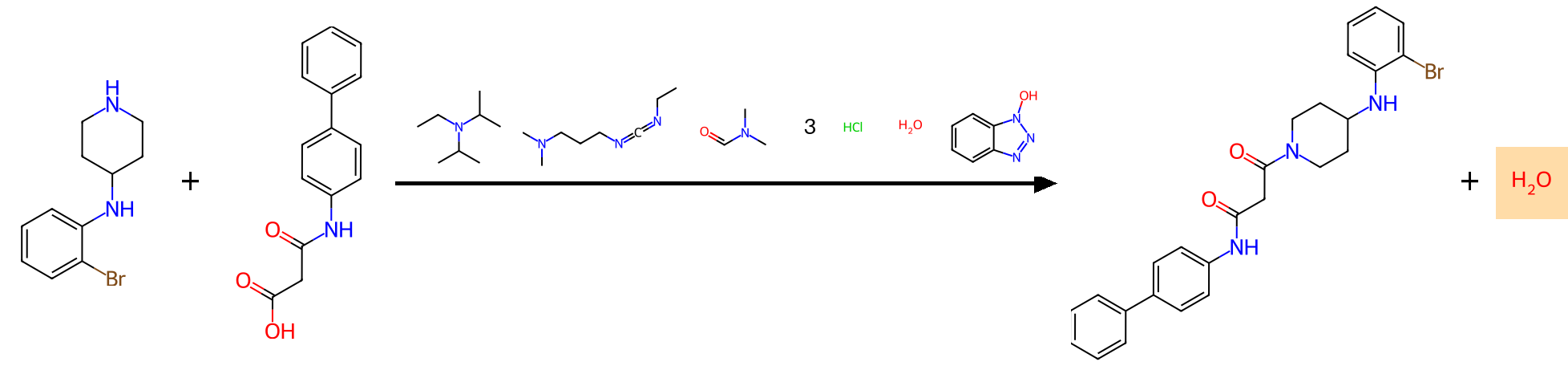}} \\
\end{tabular}
\caption{Target audit, control 35 (\texttt{USPTO\_US08129376B2\_655776}). None of the three independent chemistry-annotator completions match the benchmark target under the equivalence metric. Shaded molecules in the target are missing from the same side of at least one compared completion. Shaded molecules in a completion do not occur on the same side of the target. Agents are drawn above the arrow and are not compared. All 3 controls without a match are shown.}
\label{fig:target_audit_control_35}
\end{figure}

\begin{figure}[htbp]
\centering\small
\begin{tabular}{@{}p{2.4cm}@{\hspace{2mm}}l@{}}
USPTO input & \raisebox{-.5\height}{\includegraphics[width=946\expertauditpx]{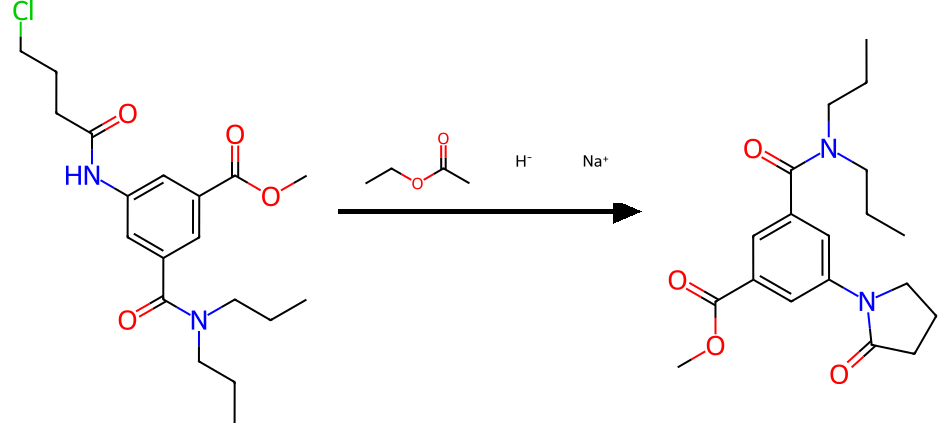}} \\[2mm]
FlowER target & \raisebox{-.5\height}{\includegraphics[width=1389\expertauditpx]{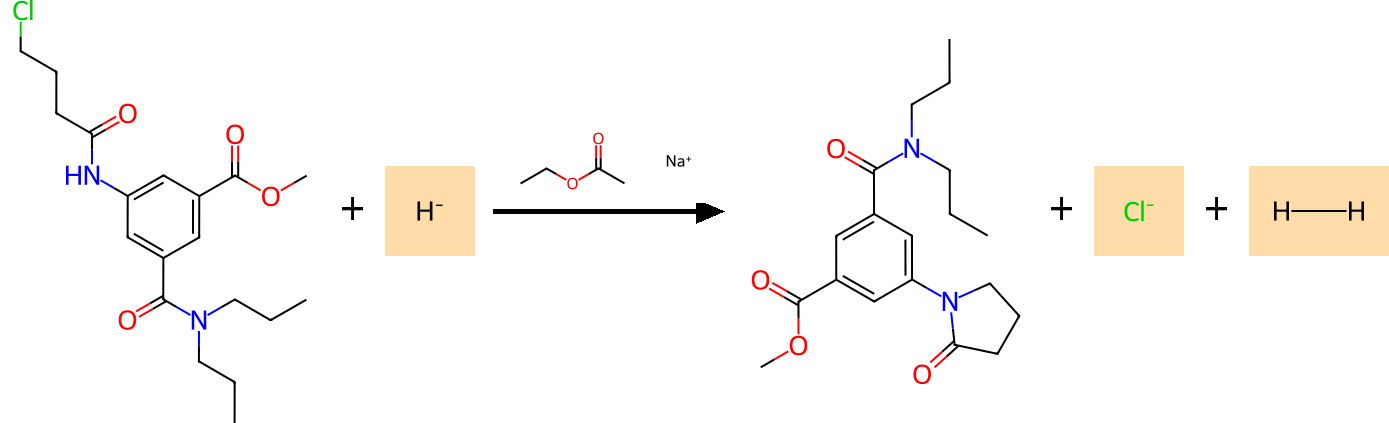}} \\[2mm]
Annotator 1\newline(no match) & \raisebox{-.5\height}{\includegraphics[width=1618\expertauditpx]{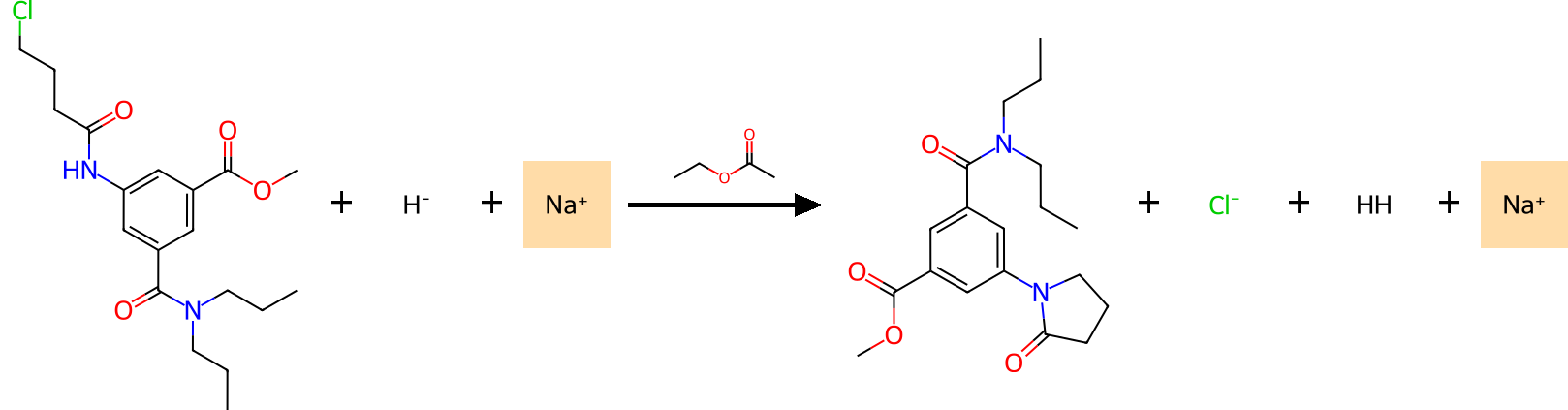}} \\[2mm]
Annotator 4\newline(no match) & \raisebox{-.5\height}{\includegraphics[width=1101\expertauditpx]{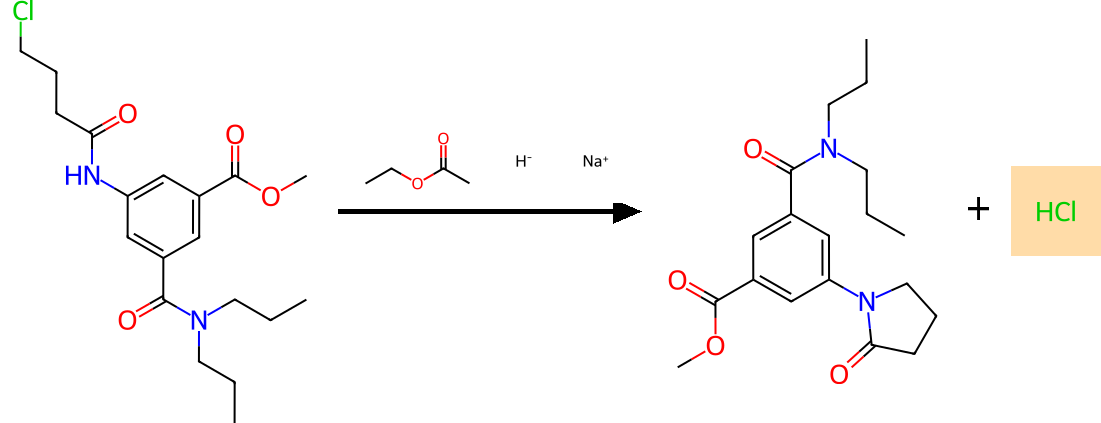}} \\[2mm]
Annotator 5\newline(no match) & \raisebox{-.5\height}{\includegraphics[width=1618\expertauditpx]{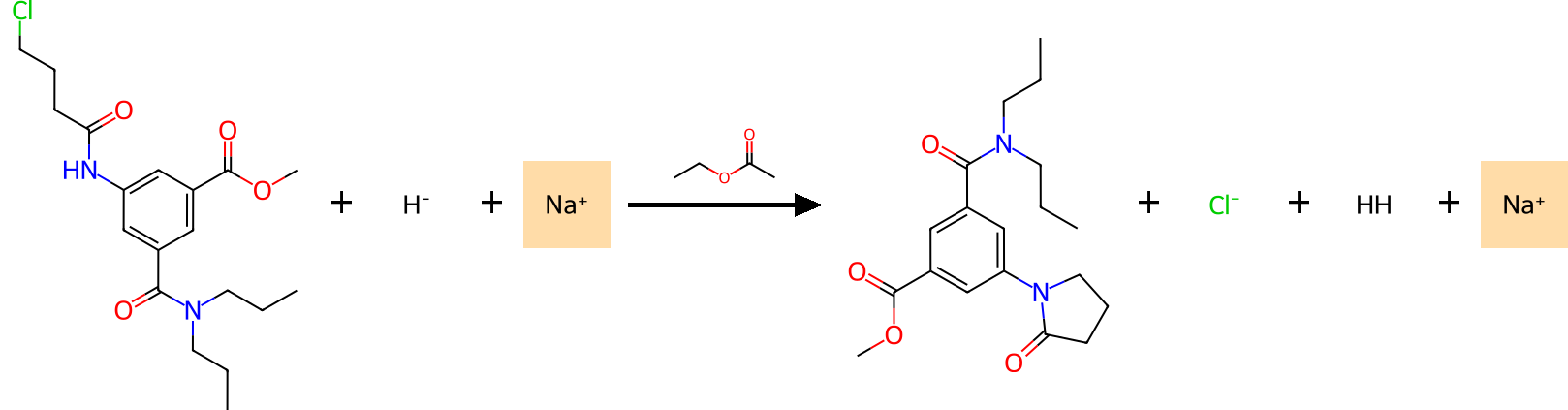}} \\
\end{tabular}
\caption{Target audit, control 83 (\texttt{USPTO\_US07253198B2\_432576}). None of the three independent chemistry-annotator completions match the benchmark target under the equivalence metric. Shaded molecules in the target are missing from the same side of at least one compared completion. Shaded molecules in a completion do not occur on the same side of the target. Agents are drawn above the arrow and are not compared. All 3 controls without a match are shown.}
\label{fig:target_audit_control_83}
\end{figure}

\begin{figure}[htbp]
\centering\small
\begin{tabular}{@{}p{2.4cm}@{\hspace{2mm}}l@{}}
USPTO input & \raisebox{-.5\height}{\includegraphics[width=2207\expertauditpx]{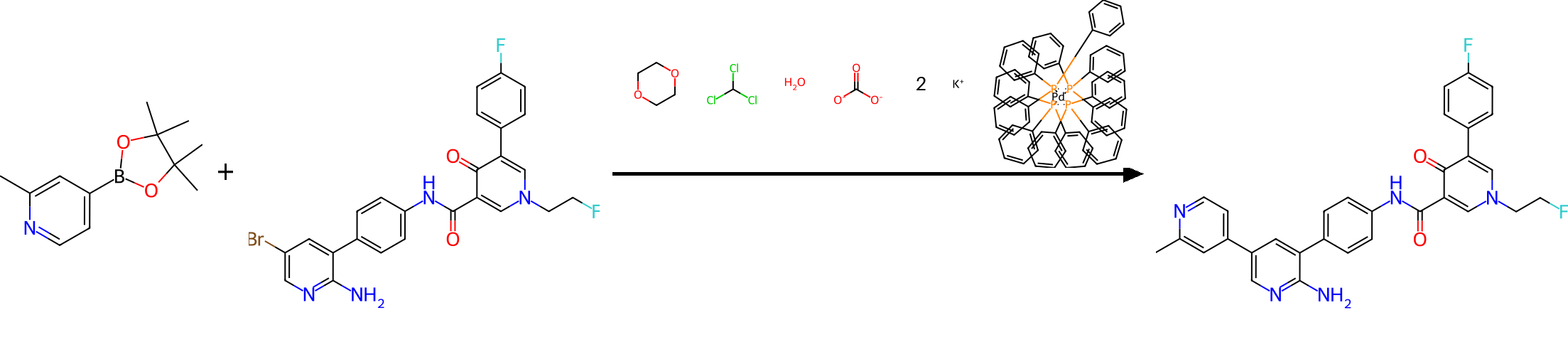}} \\[2mm]
FlowER target & \raisebox{-.5\height}{\includegraphics[width=2654\expertauditpx]{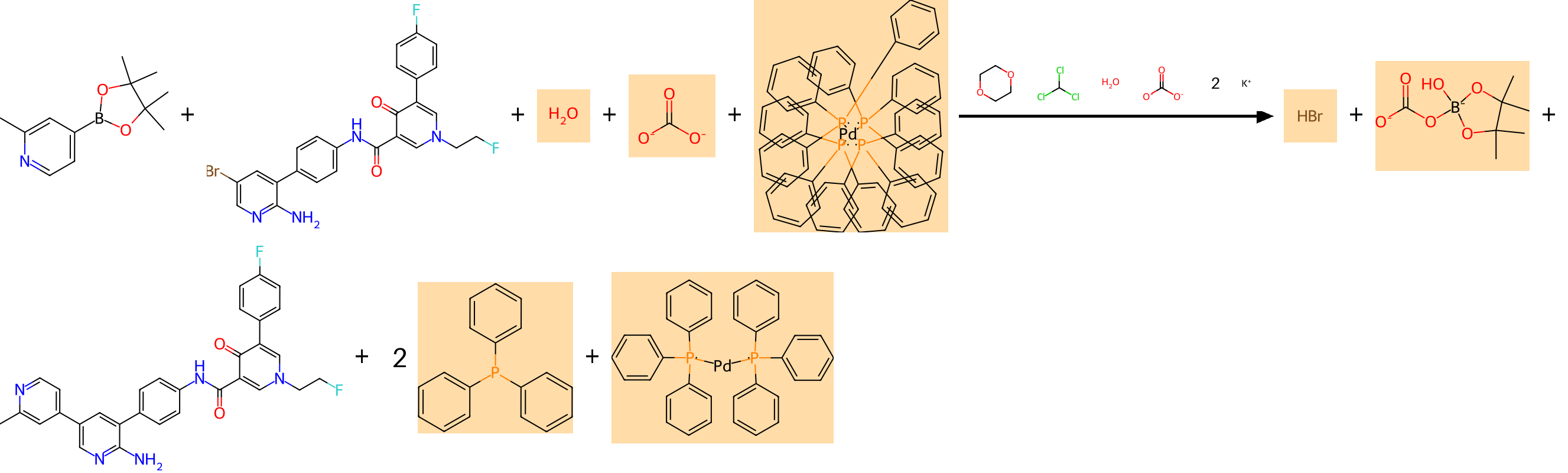}} \\[2mm]
Annotator 1\newline(no match) & \raisebox{-.5\height}{\includegraphics[width=2554\expertauditpx]{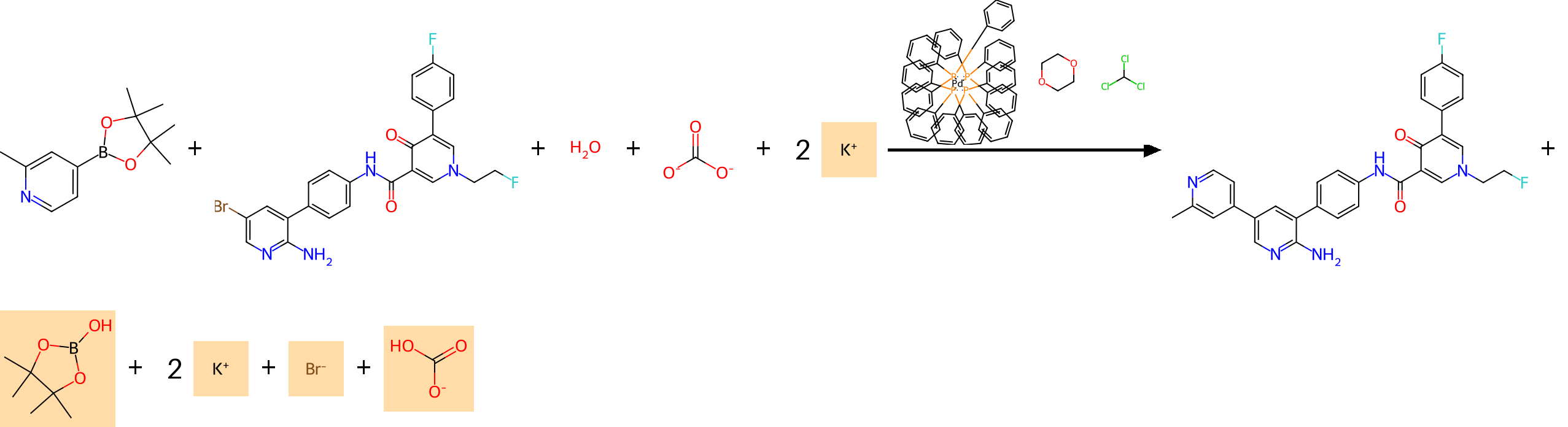}} \\[2mm]
Annotator 4\newline(no match) & \raisebox{-.5\height}{\includegraphics[width=2510\expertauditpx]{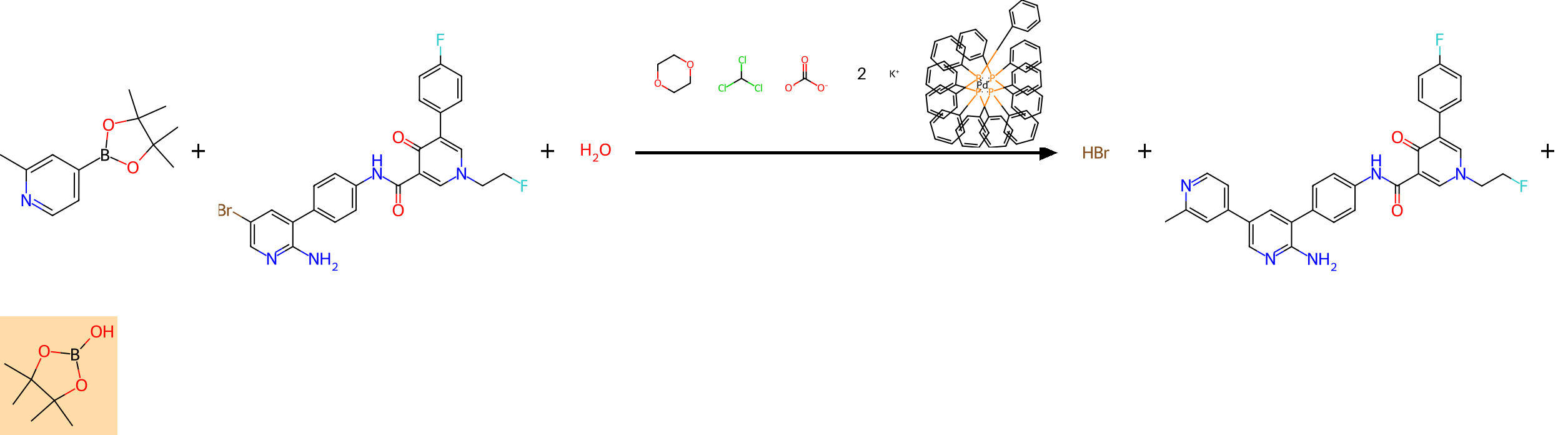}} \\[2mm]
Annotator 6\newline(no match) & \raisebox{-.5\height}{\includegraphics[width=2599\expertauditpx]{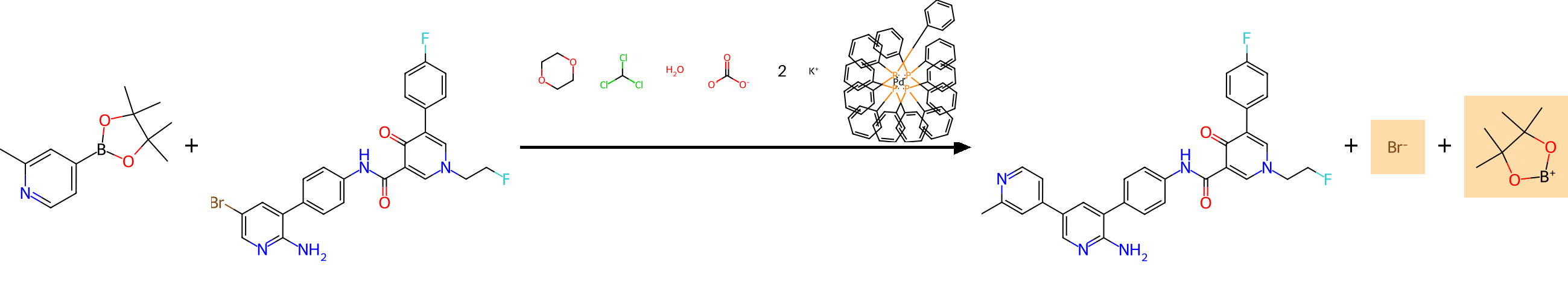}} \\
\end{tabular}
\caption{Target audit, control 92 (\texttt{USPTO\_US08933103B2\_876576}). None of the three independent chemistry-annotator completions match the benchmark target under the equivalence metric. Shaded molecules in the target are missing from the same side of at least one compared completion. Shaded molecules in a completion do not occur on the same side of the target. Agents are drawn above the arrow and are not compared. All 3 controls without a match are shown.}
\label{fig:target_audit_control_92}
\end{figure}

This appendix shows reactions from the record-level audit of benchmark targets (Appendix~\ref{app:target_audit}). Table~\ref{tab:target_audit_agreement} groups the 53 controls by how many of the three independent chemistry-annotator completions match the benchmark target under the equivalence metric. For each level with at least one match, we show the control with the lowest ID (Figures~\ref{fig:target_audit_control_17}--\ref{fig:target_audit_control_177}). Control IDs follow the order of the curation app, in which the controls were embedded at random, so this choice does not depend on the reaction. We show all three controls without a match (Figures~\ref{fig:target_audit_control_35}--\ref{fig:target_audit_control_92}). Each figure shows the USPTO input, the FlowER target and the first completion of each annotator, drawn at the same scale. Shading marks molecules that differ between the target and a completion.

\begin{table}[htbp]
\centering\small
\caption{Number of the 53 target-audit controls by how many of the three independent chemistry-annotator completions match the benchmark target under the equivalence metric, and the figures that show examples.}
\label{tab:target_audit_agreement}
\begin{tabular}{lrl}
\toprule
Matching annotators & Controls & Figures \\
\midrule
3 of 3 & 43 & \ref{fig:target_audit_control_17} \\
2 of 3 & 5 & \ref{fig:target_audit_control_2} \\
1 of 3 & 2 & \ref{fig:target_audit_control_177} \\
0 of 3 & 3 & \ref{fig:target_audit_control_35}, \ref{fig:target_audit_control_83}, \ref{fig:target_audit_control_92} \\
\bottomrule
\end{tabular}
\end{table}

\clearpage

% Generated by analysis/expert_audits/build_appendix.py; do not edit by hand.
\providecommand{\expertauditsection}{\subsection}
\providecommand{\expertauditbarrier}{\clearpage}
\makeatletter\@ifundefined{expertauditpx}{\newlength{\expertauditpx}}{}\makeatother
\setlength{\expertauditpx}{\dimexpr(\textwidth-2.4cm-2mm)/2654\relax}

\expertauditsection{Reactions from the audit of the equivalence metric}
\label{app:metric_audit_cases}

\begin{figure}[htbp]
\centering\small
\begin{tabular}{@{}p{2.4cm}@{\hspace{2mm}}l@{}}
USPTO input & \raisebox{-.5\height}{\includegraphics[width=1967\expertauditpx]{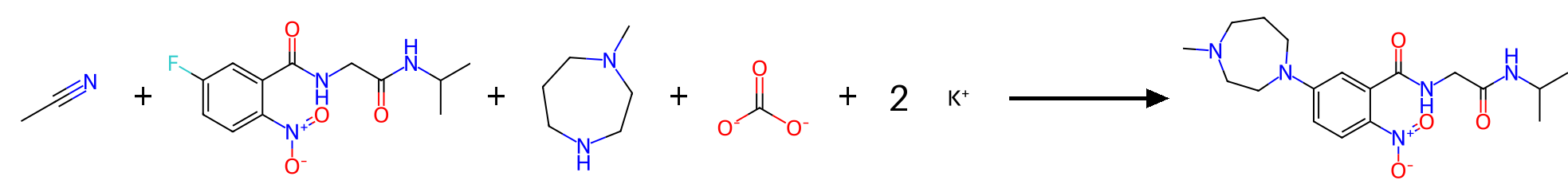}} \\[2mm]
FlowER target & \raisebox{-.5\height}{\includegraphics[width=2518\expertauditpx]{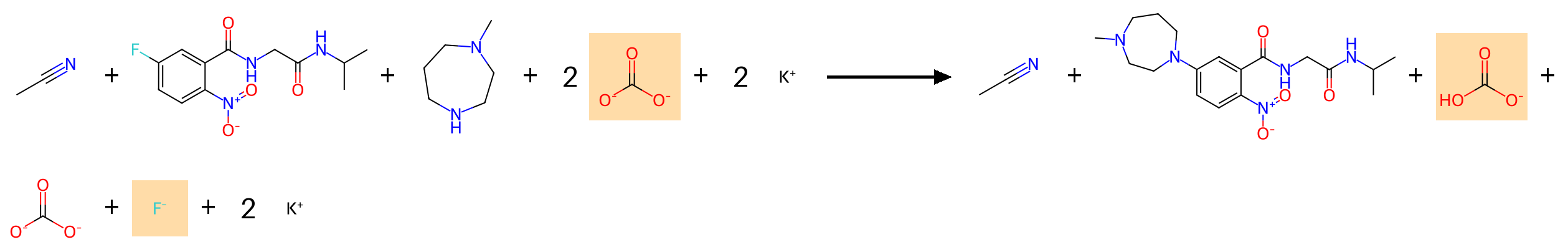}} \\[2mm]
SynRBL completion & \raisebox{-.5\height}{\includegraphics[width=2611\expertauditpx]{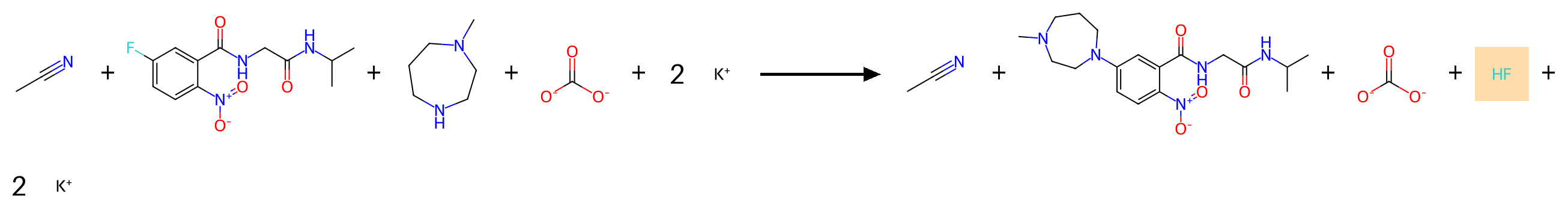}} \\
\end{tabular}
\caption{Equivalence-metric audit, case 4 (\texttt{USPTO\_US07820649B2\_570264}). Annotator judgement: equivalent (confidence: high). Revised equivalence metric: equivalent. Shaded molecules in the target are missing from the same side of at least one compared completion. Shaded molecules in a completion do not occur on the same side of the target. Agents are drawn above the arrow and are not compared. Stereochemistry is ignored, as for all SynRBL comparisons. Selection: of the 4 audited cases with this outcome, the one with the highest annotator confidence (lowest case number on ties).}
\label{fig:metric_audit_case_4}
\end{figure}

\begin{figure}[htbp]
\centering\small
\begin{tabular}{@{}p{2.4cm}@{\hspace{2mm}}l@{}}
USPTO input & \raisebox{-.5\height}{\includegraphics[width=1787\expertauditpx]{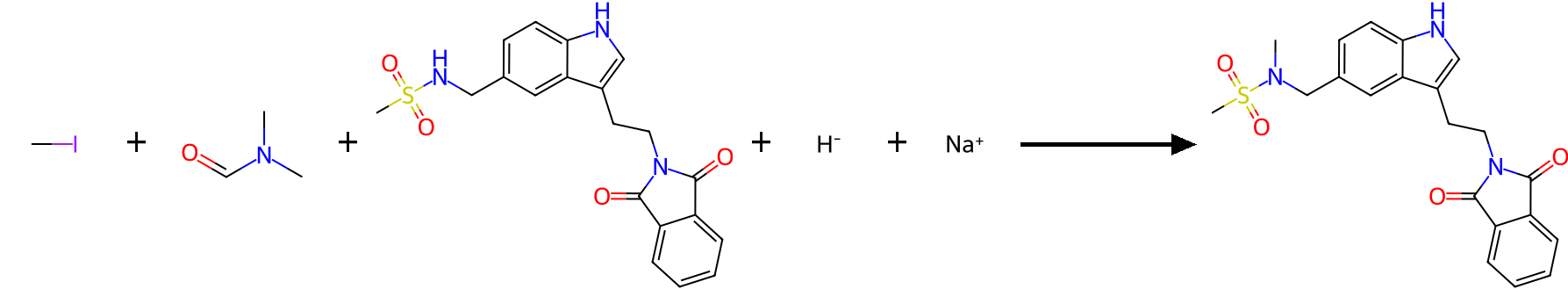}} \\[2mm]
FlowER target & \raisebox{-.5\height}{\includegraphics[width=2560\expertauditpx]{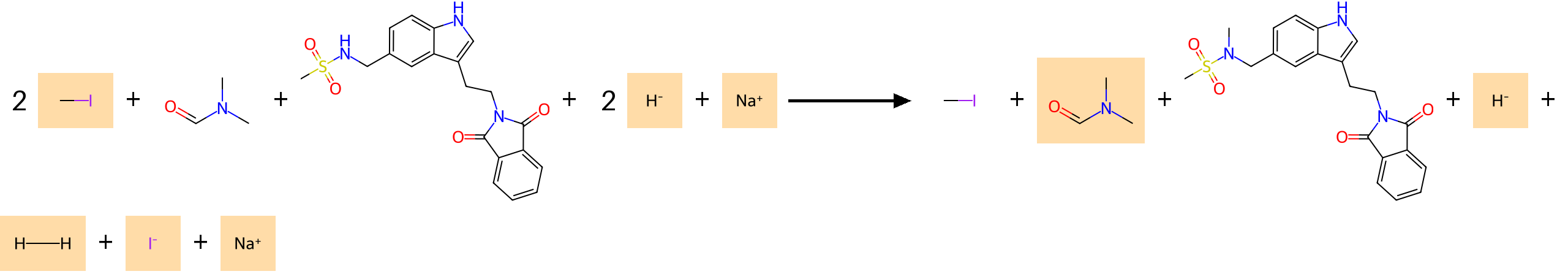}} \\[2mm]
SynRBL completion & \raisebox{-.5\height}{\includegraphics[width=2216\expertauditpx]{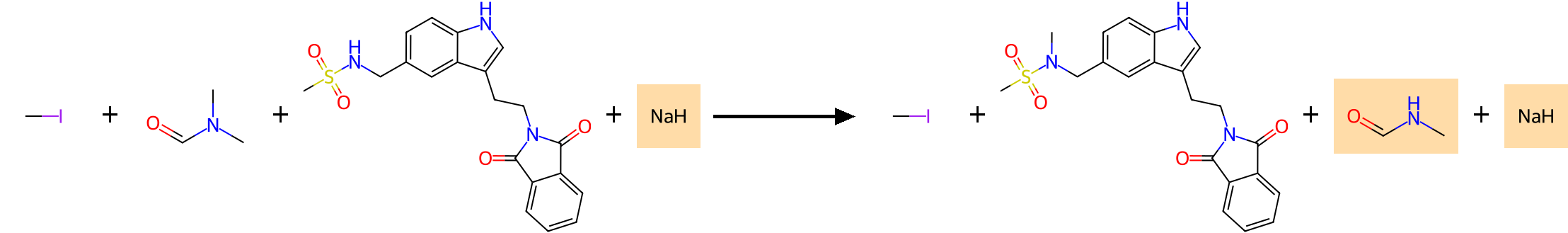}} \\
\end{tabular}
\caption{Equivalence-metric audit, case 2 (\texttt{USPTO\_US04636521\_83739}). Annotator judgement: FlowER target preferred (confidence: high). Revised equivalence metric: not equivalent. Shaded molecules in the target are missing from the same side of at least one compared completion. Shaded molecules in a completion do not occur on the same side of the target. Agents are drawn above the arrow and are not compared. Stereochemistry is ignored, as for all SynRBL comparisons. Selection: of the 8 audited cases with this outcome, the one with the highest annotator confidence (lowest case number on ties).}
\label{fig:metric_audit_case_2}
\end{figure}

\begin{figure}[htbp]
\centering\small
\begin{tabular}{@{}p{2.4cm}@{\hspace{2mm}}l@{}}
USPTO input & \raisebox{-.5\height}{\includegraphics[width=1680\expertauditpx]{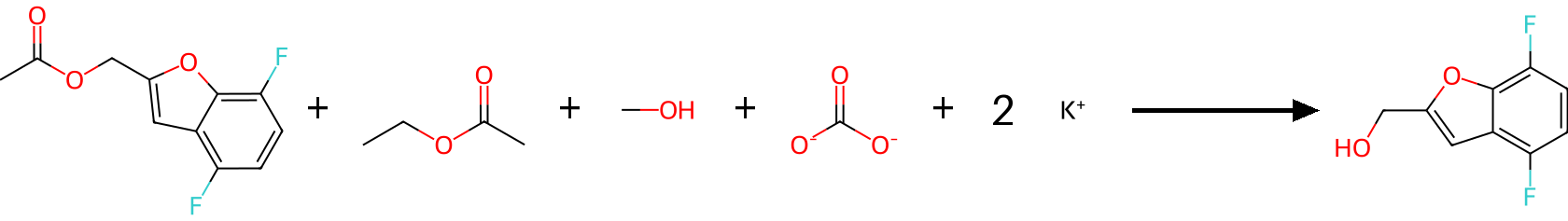}} \\[2mm]
FlowER target & \raisebox{-.5\height}{\includegraphics[width=2473\expertauditpx]{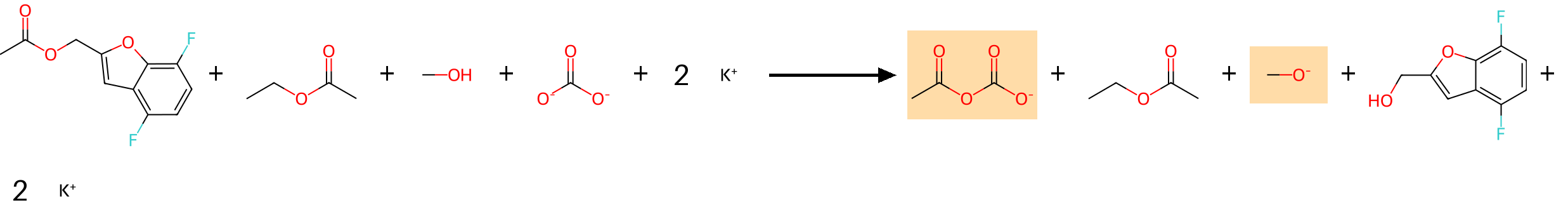}} \\[2mm]
SynRBL completion & \raisebox{-.5\height}{\includegraphics[width=2570\expertauditpx]{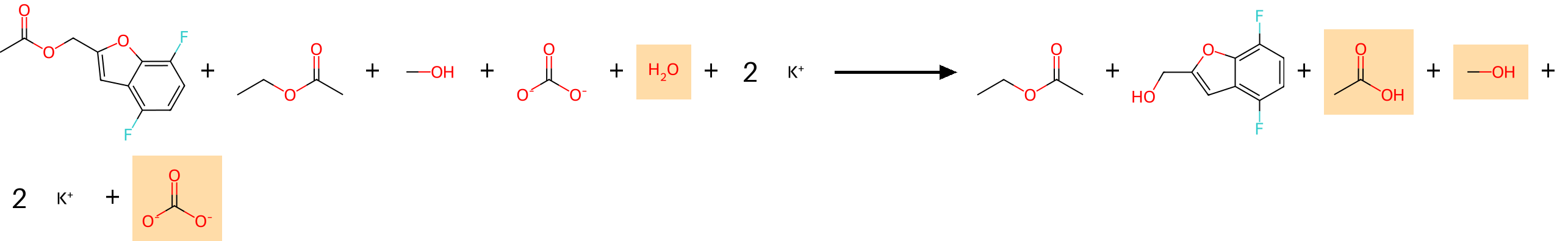}} \\
\end{tabular}
\caption{Equivalence-metric audit, case 1 (\texttt{USPTO\_US06358995B1\_291050}). Annotator judgement: SynRBL completion preferred (confidence: medium). Revised equivalence metric: not equivalent. Shaded molecules in the target are missing from the same side of at least one compared completion. Shaded molecules in a completion do not occur on the same side of the target. Agents are drawn above the arrow and are not compared. Stereochemistry is ignored, as for all SynRBL comparisons. Selection: of the 3 audited cases with this outcome, the one with the highest annotator confidence (lowest case number on ties).}
\label{fig:metric_audit_case_1}
\end{figure}

\begin{figure}[htbp]
\centering\small
\begin{tabular}{@{}p{2.4cm}@{\hspace{2mm}}l@{}}
USPTO input & \raisebox{-.5\height}{\includegraphics[width=1719\expertauditpx]{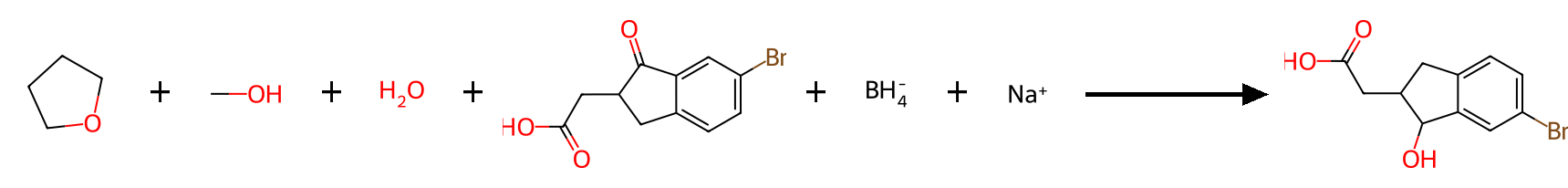}} \\[2mm]
FlowER target & \raisebox{-.5\height}{\includegraphics[width=2580\expertauditpx]{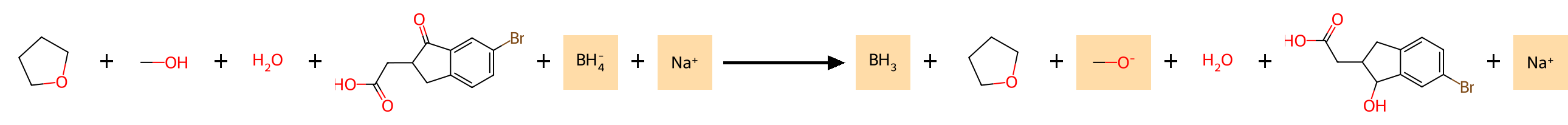}} \\[2mm]
SynRBL completion & \raisebox{-.5\height}{\includegraphics[width=2553\expertauditpx]{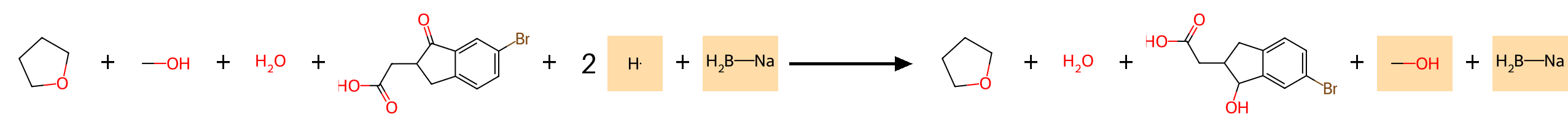}} \\
\end{tabular}
\caption{Equivalence-metric audit, case 8 (\texttt{USPTO\_US05936000\_235803}). Annotator judgement: both plausible (confidence: medium). Revised equivalence metric: not equivalent. Shaded molecules in the target are missing from the same side of at least one compared completion. Shaded molecules in a completion do not occur on the same side of the target. Agents are drawn above the arrow and are not compared. Stereochemistry is ignored, as for all SynRBL comparisons. Selection: of the 3 audited cases with this outcome, the one with the highest annotator confidence (lowest case number on ties).}
\label{fig:metric_audit_case_8}
\end{figure}

This appendix shows reactions from the audit of the equivalence metric by a chemistry annotator (Appendix~\ref{app:equiv_acc_metric}). Table~\ref{tab:metric_audit_cases} lists the annotator judgement for all 20 audited SynRBL completions and the label assigned by the revised metric. For each judgement except uncertain, we show the case with the highest annotator confidence, with ties broken by the lowest case number (Figures~\ref{fig:metric_audit_case_4}--\ref{fig:metric_audit_case_8}). The two cases judged uncertain appear only in the table. Figures of all 20 cases are provided in the code repository. Each figure shows the USPTO input, the FlowER target and the SynRBL completion, drawn at the same scale. Shading marks molecules that differ between the target and the completion.

\begin{table}[htbp]
\centering\small
\caption{Chemistry-annotator judgements for all 20 audited SynRBL completions that the initial equivalence metric classified as incorrect, the label assigned by the revised metric used throughout this paper, and the figures that show examples.}
\label{tab:metric_audit_cases}
\begin{tabular}{rllll}
\toprule
Case & Annotator judgement & Confidence & Revised metric & Figure \\
\midrule
1 & SynRBL completion preferred & Medium & not equivalent & \ref{fig:metric_audit_case_1} \\
2 & FlowER target preferred & High & not equivalent & \ref{fig:metric_audit_case_2} \\
3 & SynRBL completion preferred & Medium & not equivalent &  \\
4 & equivalent & High & equivalent & \ref{fig:metric_audit_case_4} \\
5 & FlowER target preferred & Medium & not equivalent &  \\
6 & FlowER target preferred & Medium & not equivalent &  \\
7 & FlowER target preferred & Medium & not equivalent &  \\
8 & both plausible & Medium & not equivalent & \ref{fig:metric_audit_case_8} \\
9 & SynRBL completion preferred & Low & not equivalent &  \\
10 & equivalent & High & equivalent &  \\
11 & equivalent & High & equivalent &  \\
12 & FlowER target preferred & Medium & not equivalent &  \\
13 & uncertain & Low & not equivalent &  \\
14 & FlowER target preferred & Medium & not equivalent &  \\
15 & FlowER target preferred & High & not equivalent &  \\
16 & FlowER target preferred & Low & not equivalent &  \\
17 & uncertain & Low & not equivalent &  \\
18 & both plausible & Low & not equivalent &  \\
19 & equivalent & High & equivalent &  \\
20 & both plausible & Medium & not equivalent &  \\
\bottomrule
\end{tabular}
\end{table}

\clearpage

%%%%%%%%%%%%%%%%%%%%%%%%%%%%%%%%%%%%%%%%%%%%%%%%%%%%%%%%%%%%

\newpage

\section*{NeurIPS Paper Checklist}

\begin{enumerate}

\item {\bf Claims}
    \item[] Question: Do the main claims made in the abstract and introduction accurately reflect the paper's contributions and scope?
    \item[] Answer: \answerYes{}
    \item[] Justification: We clearly state our contributions in the abstract and introduction, including the dataset, benchmark framework, and analysis. Our experiments support these claims and we explicitly mention generalization limits under distribution shift and real-world noise.


\item {\bf Limitations}
    \item[] Question: Does the paper discuss the limitations of the work performed by the authors?
    \item[] Answer: \answerYes{}
    \item[] Justification: We discuss limitations in Section 8, including template-curation bias, limited coverage of noisy real-world reactions, focus on completion rather than correction, and ambiguity due to multiple valid reaction completions.


\item {\bf Theory assumptions and proofs}
    \item[] Question: For each theoretical result, does the paper provide the full set of assumptions and a complete (and correct) proof?
    \item[] Answer: \answerNA{}
    \item[] Justification: We do not present theoretical results. Our contributions are empirical and dataset-focused.

    \item {\bf Experimental result reproducibility}
    \item[] Question: Does the paper fully disclose all the information needed to reproduce the main experimental results of the paper to the extent that it affects the main claims and/or conclusions of the paper (regardless of whether the code and data are provided or not)?
    \item[] Answer: \answerYes{}
    \item[] Justification: We describe dataset construction, preprocessing, data splits, models, and evaluation metrics in the main text and appendix. We also provide code and data to enable reproducing the results.


\item {\bf Open access to data and code}
    \item[] Question: Does the paper provide open access to the data and code, with sufficient instructions to faithfully reproduce the main experimental results, as described in supplemental material?
    \item[] Answer: \answerYes{}
    \item[] Justification: We release the dataset (aligned reaction pairs, splits and metadata) at \url{https://huggingface.co/datasets/Intelligent-molecular-systems/completeRXN-benchmark}, and the preprocessing pipeline, model training and evaluation code, with instructions for reproducing the experiments, at \url{https://github.com/Intelligent-molecular-systems/CompleteRxn-Benchmarking}. Saved predictions and software containers are at \url{https://huggingface.co/datasets/Intelligent-molecular-systems/completeRXN-reproduction}.


\item {\bf Experimental setting/details}
    \item[] Question: Does the paper specify all the training and test details (e.g., data splits, hyperparameters, how they were chosen, type of optimizer) necessary to understand the results?
    \item[] Answer: \answerYes{}
    \item[] Justification: We describe model architecture, training procedure, hyperparameters (if they vary from previous work), and data splits in Sections 5–6 and the supplement, sufficient to understand and reproduce the results.


\item {\bf Experiment statistical significance}
    \item[] Question: Does the paper report error bars suitably and correctly defined or other appropriate information about the statistical significance of the experiments?
    \item[] Answer: \answerYes{}
    \item[] Justification: We report mean and standard deviation across multiple data splits and folds (e.g., Table 2), capturing variability due to dataset partitioning. We note this Figure and Table captions and in appendix~\ref{app:experiment_details}. 


\item {\bf Experiments compute resources}
    \item[] Question: For each experiment, does the paper provide sufficient information on the computer resources (type of compute workers, memory, time of execution) needed to reproduce the experiments?
    \item[] Answer: \answerYes{} % Replace by \answerYes{}, \answerNo{}, or \answerNA{}.
    \item[] Justification: We estimate the required minimal compute ressources in appendix~\ref{app:experiment_details}.
    
\item {\bf Code of ethics}
    \item[] Question: Does the research conducted in the paper conform, in every respect, with the NeurIPS Code of Ethics \url{https://neurips.cc/public/EthicsGuidelines}?
    \item[] Answer: \answerYes{}
    \item[] Justification: We use publicly available scientific data. Chemistry annotators were hired with hourly pay to label public reactions and no personal or sensitive data were collected. We adhere to ethical standards in data usage and reporting to the best of our knowledge.

    \begin{itemize}
        \item The answer \answerNA{} means that the authors have not reviewed the NeurIPS Code of Ethics.
        \item If the authors answer \answerNo, they should explain the special circumstances that require a deviation from the Code of Ethics.
        \item The authors should make sure to preserve anonymity (e.g., if there is a special consideration due to laws or regulations in their jurisdiction).
    \end{itemize}

\item {\bf Broader impacts}
    \item[] Question: Does the paper discuss both potential positive societal impacts and negative societal impacts of the work performed?
    \item[] Answer: \answerYes{}
    \item[] Justification: We discuss applications in chemical process modeling and sustainability. Potential risks are limited and indirect, but we acknowledge that errors in our data could propagate to downstream analyses.

    
\item {\bf Safeguards}
    \item[] Question: Does the paper describe safeguards that have been put in place for responsible release of data or models that have a high risk for misuse (e.g., pre-trained language models, image generators, or scraped datasets)?
    \item[] Answer: \answerNA{}
    \item[] Justification: The dataset consists of structured chemical reactions derived from public sources and does not pose significant misuse risks requiring special safeguards.


\item {\bf Licenses for existing assets}
    \item[] Question: Are the creators or original owners of assets (e.g., code, data, models), used in the paper, properly credited and are the license and terms of use explicitly mentioned and properly respected?
    \item[] Answer: \answerYes{}
    \item[] Justification: We cite all datasets, models, and tools used (e.g., USPTO, FlowER, SynRBL, Molecular Transformer) and follow their respective terms of use. The license of each source dataset is listed in the dataset card on Hugging Face.


\item {\bf New assets}
    \item[] Question: Are new assets introduced in the paper well documented and is the documentation provided alongside the assets?
    \item[] Answer: \answerYes{}
    \item[] Justification: We document the dataset construction process, mappings, and evaluation protocol in the paper and appendix, and provide code and metadata.


\item {\bf Crowdsourcing and research with human subjects}
    \item[] Question: For crowdsourcing experiments and research with human subjects, does the paper include the full text of instructions given to participants and screenshots, if applicable, as well as details about compensation (if any)? 
    \item[] Answer: \answerYes{}
    \item[] Justification: Seven chemistry students (three graduate, four undergraduate) were hired with hourly pay, funded by an Open Science Community Delft grant, to complete or judge reactions for the audits in Appendices~\ref{app:target_audit} and \ref{app:equiv_acc_metric}. The appendix describes the annotation setup, and the full instructions given to the annotators (a guide to the curation app and the instructions for the metric audit) are provided in the linked code repository.


\item {\bf Institutional review board (IRB) approvals or equivalent for research with human subjects}
    \item[] Question: Does the paper describe potential risks incurred by study participants, whether such risks were disclosed to the subjects, and whether Institutional Review Board (IRB) approvals (or an equivalent approval/review based on the requirements of your country or institution) were obtained?
    \item[] Answer: \answerNA{}
    \item[] Justification: The annotators were hired to label publicly available chemical reactions. No personal data were collected, and the task posed no risks beyond regular work.


\item {\bf Declaration of LLM usage}
    \item[] Question: Does the paper describe the usage of LLMs if it is an important, original, or non-standard component of the core methods in this research? Note that if the LLM is used only for writing, editing, or formatting purposes and does \emph{not} impact the core methodology, scientific rigor, or originality of the research, declaration is not required.
    %this research? 
    \item[] Answer: \answerNA{}
    \item[] Justification: LLMs do not impact the core methodology of this work. 

\end{enumerate}

\end{document}